\documentclass{article}

\usepackage{lipsum}
\usepackage[top=1in, bottom=1in, left=1in, right=1in]{geometry}

\usepackage{amsthm}
\usepackage{amsmath}
\usepackage{amssymb}
\usepackage{algorithm}
\usepackage{algorithmic}
\usepackage[colorlinks = true,
            linkcolor = violet,
            urlcolor  = violet,
            citecolor = violet,
            anchorcolor = violet]{hyperref}
\usepackage{url}
\usepackage[pdftex]{graphicx}
\usepackage{subcaption}
\usepackage{multirow}
\usepackage{mathtools}
\usepackage{xcolor}
\usepackage{authblk}
\usepackage{placeins}
\usepackage{wrapfig}
\usepackage{listings}
\usepackage{tcolorbox}
\usepackage[normalem]{ulem}
\usepackage{booktabs} 
\usepackage[sort,numbers]{natbib}

\definecolor{codeblue}{rgb}{0.1, 0.1, 0.8}
\definecolor{codegreen}{rgb}{0, 0.5, 0}
\definecolor{codegray}{rgb}{0.5, 0.5, 0.7}
\definecolor{darkgray}{rgb}{0.25, 0.25, 0.25} 

\if ab
\newcommand{\maz}[1]{{\color{red}Maz: #1}}
\else 
\newcommand{\maz}[1]{}
\fi
\newcommand{\coticl}{\emph{CoT-ICL Lab}}

\makeatletter
\renewcommand{\paragraph}{%
  \@startsection{paragraph}{4}%
  {\z@}{1.5ex \@plus 1ex \@minus .2ex}{-1em}%
  {\normalfont\normalsize\bfseries}%
}
\makeatother

\newcommand{\norm}[1]{\left\lVert#1\right\rVert}

\def\mA{{\mathbf{A}}}

\def\mE{{\mathbf{E}}}

\def\mI{{\mathbf{I}}}

\def\mS{{\mathbf{S}}}

\def\mU{{\mathbf{U}}}
\def\mV{{\mathbf{V}}}
\def\mW{{\mathbf{W}}}

\def\vzero{{\mathbf{0}}}

\def\ve{{\mathbf{e}}}

\def\vp{{\mathbf{p}}}

\def\vx{{\mathbf{x}}}
\def\vy{{\mathbf{y}}}

\def\gF{{\mathcal{F}}}
\def\gG{{\mathcal{G}}}
\def\gH{{\mathcal{H}}}

\def\gN{{\mathcal{N}}}

\def\gV{{\mathcal{V}}}

\def\sI{{\mathbb{I}}}

\def\sN{{\mathbb{N}}}

\def\sR{{\mathbb{R}}}

\def\tTF{{\texttt{TF}}}
\def\tCE{{\texttt{CE}}}
\def\tacc{{\texttt{accuracy}}}

\newtheorem{theorem}{Theorem}[section]

\newtheorem{definition}[theorem]{Definition}

\title{\coticl : A Synthetic Framework for Studying Chain-of-Thought Learning from In-Context Demonstrations}
\author[]{Vignesh Kothapalli\footnote{Correspondence to: Vignesh Kothapalli, New York University, vk2115@nyu.edu}}
\author[]{Hamed Firooz}
\author[]{Maziar Sanjabi}
\affil[]{}
\date{}
\begin{document}

\maketitle
\begin{abstract}

We introduce \coticl, a framework and methodology to generate synthetic tokenized datasets and systematically study chain-of-thought (CoT) in-context learning (ICL) in language models. \coticl~allows fine grained control over the complexity of in-context examples by decoupling (1) the causal structure involved in chain token generation from (2) the underlying token processing functions. We train decoder-only transformers (up to 700M parameters) on these datasets and show that CoT accelerates the accuracy transition to higher values across model sizes. In particular, we find that model depth is crucial for leveraging CoT with limited in-context examples, while more examples help shallow models match deeper model performance. Additionally, limiting the diversity of token processing functions throughout training improves causal structure learning via ICL. We also interpret these transitions by analyzing transformer embeddings and attention maps. Overall, \coticl~serves as a simple yet powerful testbed for theoretical and empirical insights into ICL and CoT in language models. The code is available at: \href{https://github.com/kvignesh1420/cot-icl-lab}{\texttt{https://github.com/kvignesh1420/cot-icl-lab}}.

\end{abstract}

\section{Introduction}
Transformer-based language models \cite{Vaswani2017AttentionIA} have demonstrated remarkable capabilities in tasks requiring emergent reasoning behaviors, such as few-shot ICL \cite{brown2020language, firooz2025360brew} and CoT prompting \cite{wei2022chain,nye2021show,kojima2022large}. \textit{In-context learning} refers to a phenomenon wherein language models generalize to new tasks by conditioning on a small number of input-output examples without explicit parameter updates \cite{dong2022survey}. Meanwhile, \emph{chain-of-thought prompting} augments the input with explicit intermediate reasoning steps that can guide the model's generative process toward more accurate solutions \cite{wei2022chain}. Despite the substantial performance gains witnessed in various natural language processing tasks~\cite{kim2023the}, the precise mechanisms and architectural factors driving ICL and CoT remain only partially understood.

Recent studies have ventured into controlled synthetic tasks to understand how transformers learn in-context \cite{garg2022can,von2023transformers, bai2023transformers}. These works often rely on real-valued examples consisting of single-input and single-output pairs, and study if the transformers can learn linear or non-linear function classes. While these tasks facilitate theoretical analysis, they leave open the question of whether the findings \textit{readily} extend to more complex or compositional settings, especially pertaining to discrete tokenized sequences. In a parallel line of research, investigations of CoT prompting for NLP tasks often rely on short, human-annotated explanations or heuristics, thereby limiting the variety and control of ``reasoning'' processes considered \cite{wang2022rationale,liu2024dag,yang2024context,prabhakar2024deciphering}. Although such strategies have yielded valuable insights, there does not exist a setup that unifies ICL and CoT and facilitates systematic probing of different aspects of complexity—ranging from vocabulary size and chain length (i.e, the number of tokens involved in the reasoning process) to the shape and sparsity of dependencies between tokens.

In this work, we introduce \coticl, a tokenized synthetic dataset generation framework, that is specifically designed for studying how transformer-based models acquire chain-of-thought reasoning in-context. Our framework differs from prior work in the following ways:
\begin{enumerate}
    \item \textbf{Tokenized setup akin to language.} Unlike many purely numeric toy tasks, we consider inputs and chain tokens in a discrete token space (i.e, a custom vocabulary $\gV$). This setup aligns closely with natural language prompting and facilitates complexity control via vocabulary size.
    
    \item \textbf{Decoupled structure and token processing functions.} We represent the causal structure of the `reasoning' chain via a directed acyclic graph (DAG) and implement token processing via arbitrary MLP transformations of the corresponding `unknown' data embeddings $\mE_{\texttt{data}} \in \sR^{|\gV| \times d}$. This separation grants flexibility in controlling problem difficulty —e.g., by manipulating chain length, number of edges in the DAG, depth and activation functions in MLPs and the dimension of data embeddings $d$.

    \item \textbf{Multi input-output ICL examples.} A majority of the efforts which study ICL in transformer models rely on (real-valued) single input-output examples in-context~\cite{garg2022can,bai2023transformers,li2023dissecting}. Our setup addresses these limitations and allows researchers to use tokenized, multi-input multi-output examples in-context, which is closer to practice. To the best of our knowledge, this is the first work to introduce and analyze transformer models in such controlled settings. 
    
    \item \textbf{Ablation-friendly design.} By varying one component at a time (vocabulary, number of input tokens or chain tokens per example, DAG connectivity, MLP complexity, or the underlying transformer architecture), researchers can precisely identify which facets of the problem most challenge the model's ICL and CoT capabilities. 
\end{enumerate}

\paragraph{Key Results.} In addition to proposing the \coticl~framework, we showcase how it can be used to gain insights into the abilities of decoder-only transformer models in ICL with and without CoT. 
Specifically:
\begin{itemize}
\item Transformer-models undergo phase transitions in accuracy while training on ICL problems. Such transitions are facilitated by model size, availability of more examples in-context and CoT prompting.

\item We empirically show that the phase transition correlates with the alignment between model's token embeddings and the data/language embeddings $\mE_{\texttt{data}}$. Furthermore, when utilizing a finite set of token processing functions to generate the \coticl~data, this reduction in problem complexity facilitates the attention maps of the model to capture the underlying reasoning DAG and excel at ICL.
\item In essence, we highlight an interplay between the problem complexity induced due to diverse token processing functions and the DAG structure. As DAG sparsity reduces and the number of token processing functions increases, we observed that larger models tend to adapt to such diversity in ICL problems and leverage CoT to outperform the smaller models. Thus, showcasing the intricacies involved in scaling the model size for ICL performance.
\end{itemize}

\section{Related Work}
\paragraph{In-Context Learning.}
Initially popularized by GPT-3 \cite{brown2020language}, in-context learning has garnered extensive attention for its surprising ability to generalize with just a few example prompts. Many investigations center on how transformers might implicitly perform gradient descent or implement other adaptation mechanisms in their hidden activations \citep{garg2022can,akyurek2024incontext,von2023transformers, bai2023transformers}. See \cite{dong2022survey, zhou2024mystery} for surveys on the topic. However, these analyses often assume real-valued examples and very simple data distributions, leaving room to explore richer compositional structures that can align with natural language tasks.

\paragraph{Chain-of-Thought.}
CoT prompting \cite{wei2022chain,nye2021show,kojima2022large, chu2024navigate} has emerged as an effective technique for eliciting more interpretable (and sometimes more accurate) intermediate reasoning from large language models. Despite empirical successes, debate persists as to whether models truly learn a generalized reasoning algorithm or simply latch onto superficial features \cite{wang2022rationale}. While some efforts \cite{liu2024dag,prabhakar2024deciphering} systematically study CoT's potential, they often rely on limited or handcrafted tasks that do not fully capture the complexity of multi-step compositional processes.

\paragraph{Synthetic Tasks for Controlled Model Analysis.}
Synthetic tasks provide controlled environments that enable precise interventions, ablation studies, and theoretical insights into the model behavior and training dynamics \cite{garg2022can,von2023transformers, bai2023transformers}. However, existing synthetic settings generally remain numeric and follow overly restrictive Markovian assumptions \cite{edelman2024the} (e.g., a single parent for each token). Our proposed \coticl~extends these efforts by decoupling the causal structure from token-processing functions. We leverage directed acyclic graphs (DAGs) to control the branching factor in the chain generation, and MLPs for varied levels of token transformations. This design grants extensive configurability, encompassing vocabulary size, multi-input example length, chain length, DAG sparsity, MLP depth, activations and more.

\paragraph{Chain-of-Thought \& Compositional Reasoning.}
Recent studies focus on dissecting CoT to assess how compositionality might emerge from ICL. For instance, \cite{li2023dissecting} examines how CoT might be effectively disentangled into filtering and learning the intermediate features of the MLP components in the prompt. Furthermore, our work can be treated as a generalization of the \texttt{MechanisticProbe} approach by  \cite{hou2023towards} where the filtering and reasoning tree construction process is not limited by the availability of natural language datasets.  While these and related efforts \cite{yang2024context,prabhakar2024deciphering} represent significant progress toward understanding emergent reasoning, their experimental setups typically do not offer the level of systematic and fine-grained complexity that \coticl~enables (see also Appendix~\ref{app:sec:comp_related_work}).

\bigskip

\vspace{-5mm}
\section{Preliminaries and Setup}

\begin{figure*}[t!]
    \centering
    \includegraphics[width=0.9\textwidth]{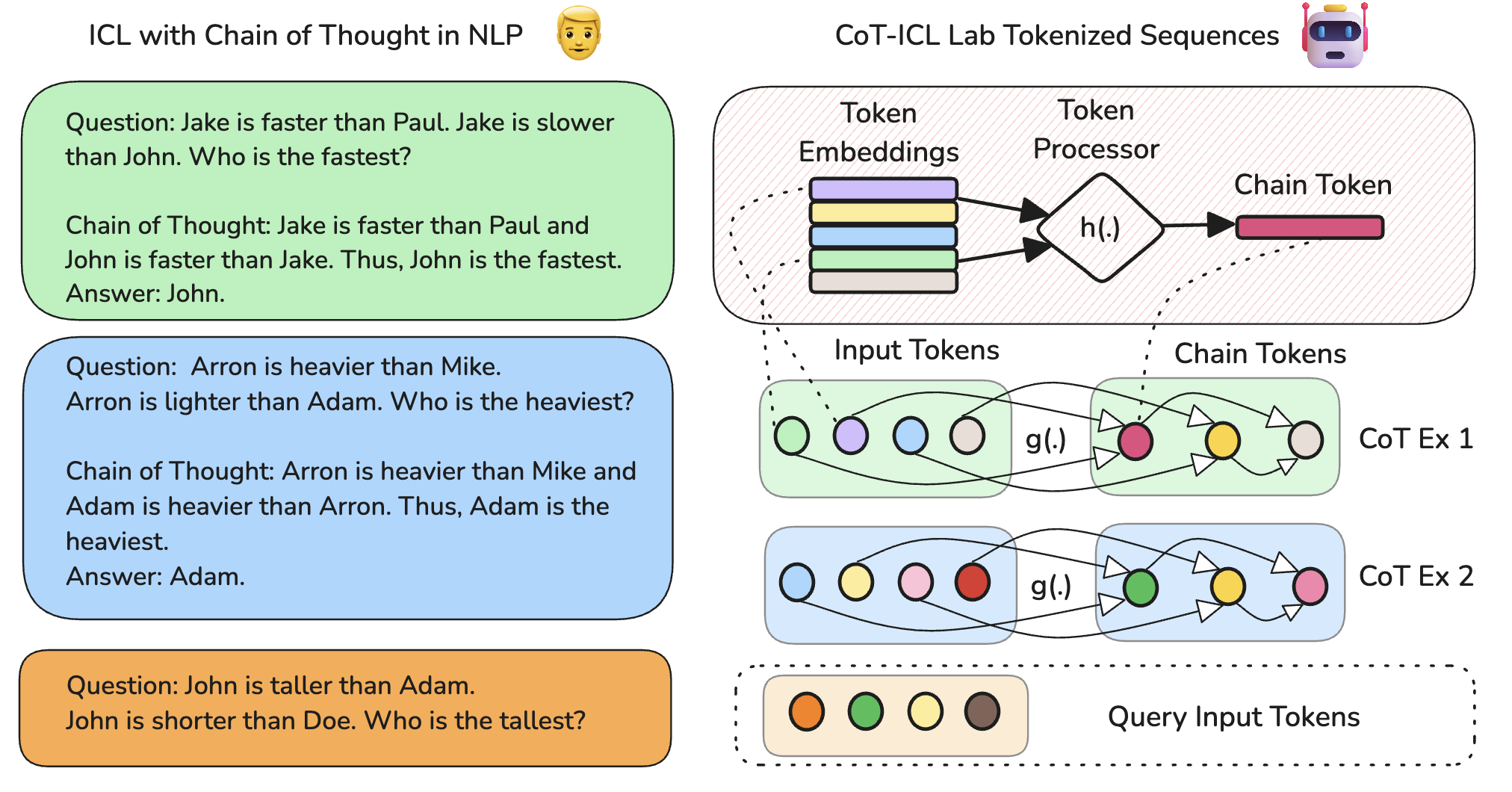}
    \caption{\coticl~overview (right) and comparison to CoT ICL in NLP (left). The figure on the left illustrates a scenario where $2$ CoT examples (colored green and blue) are available in-context along with a question (colored in orange). A corresponding scenario using \coticl~is presented on the right where we model the causal structure via the DAG $g \in \gG$ and process the data embeddings $\mE_{data}$ using the token processor function $h \in \gH$. }
    \label{fig:cot_icl_intro}
    \vspace{-3mm}
\end{figure*}

\paragraph{Notation.} Let $\{1, \cdots, K\} = [K]$. We consider a vocabulary $\gV$ to represent the tokens of a synthetic language. Let $\gF$ denote a class of functions that are compositional in nature. Formally, a function $f \in \gF$ is composed of $C$ sub-functions as: $f = f_C \circ f_{C-1} \cdots \circ f_1$. Given $N$ input tokens $\vx = (x_1, \cdots, x_N) \in \gV^N$, the function $f$ recursively generates $C$ \textit{chain tokens} $\vy = (y_1, \cdots, y_C) \in \gV^C$ as follows:
\begin{align}
\label{eq:y_c_with_f}
    y_c = f_c(x_1, \cdots, x_N, y_1, \cdots, y_{c-1}), \hspace{2pt} \forall c \in [C].
\end{align}
Here $\vy_{:C-1} = (y_1, \cdots, y_{C-1})$ are treated as the \textit{intermediate tokens}  and $y_C$ as the \textit{answer token}. The recursive process involves all the \textit{input} and \textit{intermediate tokens} to generate the \textit{answer token} and presents a generalized setup to study CoT. The full notation list is presented in Table~\ref{tab:notations}.

\subsection{Compositional Nature of $\gF$}

Let $\gG, \gH$ denote the causal structure and token processing function classes respectively. In this work, we consider the sub-functions involved in the composition of $f = f_C \circ f_{C-1} \cdots \circ f_1$ to be formulated as $f_c = h_c \circ g_c$, where $g_c \in \gG, h_c \in \gH$. Given input tokens $(x_1, \cdots, x_N)$, the chain tokens in \eqref{eq:y_c_with_f} are decomposed into:
\begin{align}
\begin{split}
\label{eq:comp_f_dag}
    y_c &= f_c\left(x_1, \cdots, x_n, y_1, \cdots, y_{c-1}\right) \\
    &= h_c\left(g_c(x_1, \cdots, x_n, y_1, \cdots, y_{c-1})\right), \forall c \in [C]
\end{split}
\end{align}

\paragraph{$\bullet$ The causal structure function class $\gG$.} This class represents functions which take an arbitrary number of tokens and filter a fixed number of $M \le N \in \sN$ tokens. These $M$ parent tokens represent the causal dependency of a \textit{chain token}.
\paragraph{$\bullet$ The token processing function class $\gH$.} These functions process the $M$ selected tokens from $\gG$ and output a single chain token. Thus, emulating an arbitrary ground-truth `reasoning process' which the transformer models are trained to approximate.

\subsection{Multi-Input ICL and CoT}

\paragraph{Sequence design.} We consider a generalized ICL problem of learning $f \in \gF$ with (multi) input-output pairs in the token space. An \textit{example} is defined as a vector of $N$ input tokens and the corresponding \textit{answer token}, as per \eqref{eq:y_c_with_f}. A collection of $K \in \sN$ such \textit{examples} results in a sequence $\vp^K(f)$ as follows:
\begin{align}
\label{eq:icl_sequence}
    \vp^K(f) = \left( x_1^{(i)}, \cdots, x_N^{(i)}, y_{C}^{(i)} \right)_{i=1}^K.
\end{align}
By including the \textit{intermediate tokens} in an \textit{example}, we obtain a \textit{CoT example}, which is now a vector of $N$ input tokens, and all the $C$ \textit{chain tokens}. The corresponding sequence $\vp^K_{CoT}(f)$ is given as follows:
\begin{align}
\label{eq:cot_icl_sequence}
    \vp^K_{CoT}(f) = \left( x_1^{(i)}, \cdots, x_N^{(i)}, y_1^{(i)}, \cdots, y_{C}^{(i)} \right)_{i=1}^K.
\end{align}

\section{\coticl: Data Generation}
\label{sec:cot_icl_lab_data_gen}
In this section, we present details about the synthetic data generation using \coticl~and draw parallels with NLP tasks.

\paragraph{Language vocabulary embedding.} 

To create synthetic training and evaluation datasets via the \coticl~framework, we consider a vocabulary $\gV$ of arbitrary size and associate with it a common data embedding matrix $\mE_{\texttt{data}} \in \sR^{|\gV| \times d}$. Here $d$ denotes the data embedding dimension and the entries are sampled i.i.d from $\gN(0,1)$.
In particular, $\mE_{\texttt{data}}$ will be leveraged by $h \in \gH$ to process embeddings of the tokens and return a new token (see Figure~\ref{fig:cot_icl_intro}).

\paragraph{Causal structure via DAGs.} $\gG$ is selected to be a class of topologically sorted DAGs whose (1) edge connectivity represents the causality involved in chain generation and (2) whose sparsity controls the usage of input and intermediate tokens. For notational simplicity, we represent DAGs in our setup as $\gG(M,N,C)$. We sample one DAG per sequence and use it to create all (CoT-) examples within the sequence. For instance, given input tokens $x_1, x_2, x_3, x_4$ and chain tokens $y_1, y_2, y_3$, we illustrate in Figure~\ref{fig:cot_icl_intro} a DAG which maps $y_1 \leftarrow \{x_1, x_2\}$, $y_2 \leftarrow \{x_3, x_4\}$ and $y_3 \leftarrow \{y_1, y_2\}$. The possible structures that can be sampled using a particular choice of $M,N,C$ controls the diversity of causal structures in our dataset of sequences. In the following sections, we experiment with different choices of $M,N,C$ and analyze its impact on model behavior.

\paragraph{Token processing via MLPs.}
The function class $\gH$ is selected to be MLPs whose complexity is controlled by the choice of activations $\phi$ such as \texttt{ReLU, SiLU, LeakyReLU, Identity}, and the depth ranging from $l \in \{1,2,3,4,5\}$. To generate a single chain token, we randomly initialize an MLP based on $l,\phi$ and use it to process the embeddings of the $M$ parent tokens. We take the mean of the $M$ final layer features, apply the activation function again and multiply with $\mE_{\texttt{data}}^\top$ to obtain a chain token via arg-max. For notational simplicity, we represent MLPs of depth $l$ and activation $\phi$ as $\gH(l, \phi)$ (see algorithm in Appendix~\ref{app:alg_H}). Thus, we sample $C$ MLPs per sequence (one for each chain token) and use them to generate the chain tokens of all $K$ examples within the sequence. In essence, these token processing functions are shared across the ICL examples in each sequence but differ across sequences. 
We present comprehensive details about the (1) distribution of tokens, and (2) the flexibility of our setup in terms of simulating the complexity of real world datasets~\cite{kim2023the} in Appendix~\ref{app:cot_icl_datasets}.

\section{Model Training and Evaluation}

\paragraph{Training.} We employ NLP style next-token prediction training of decoder only transformer ($\tTF$) models \cite{radfordimproving} with Cross-Entropy ($\tCE$) loss. We employ the supervised fine-tuning strategy to compute the $\tCE$ loss only on the $K$ answer tokens for $\vp^K$ and on all the $K \times C$ chain tokens for $\vp^K_{CoT}$ \cite{garg2022can, bai2023transformers, li2023dissecting}.

\paragraph{Evaluation via $\tacc$.} To measure the ICL ability of a $\tTF$ model, we measure the $\tacc$ of predicting the answer token of the query (final) example. Formally, we generate the evaluation sequence prefix with $K-1$ in-context examples and append the query input tokens $\tilde{\vx} = (\tilde{x}_1, \cdots, \tilde{x}_N) \in \gV^N$ at the end. For the query input $\tilde{\vx}$, the ground truth chain tokens $\tilde{\vy} = (\tilde{y}_1, \cdots, \tilde{y}_C) \in \gV^C$ are generated by the recursive formulation given in \eqref{eq:y_c_with_f} using a test function $\tilde{f} \in \gF$. Note that there is a difference in model predictions w/ and w/o CoT as follows:
\begin{align}
\begin{split}
    \hat{y}_{pred} &:= \tTF\left( \vp^{K-1}(\tilde{f}), \tilde{\vx} \right) \hspace{10pt} \textit{ w/o CoT,} \\
    \hat{y}_{pred} &:= \tTF^{\circ C}\left( \vp_{CoT}^{K-1}(\tilde{f}), \tilde{\vx} \right) \hspace{10pt} \textit{ w/ CoT}. 
\end{split}
\end{align}
Here $\tTF^{\circ C}(\cdot)$ represents the $C$-step auto-regressive greedy token generation by the model (without any teacher-forcing) as follows \cite{li2023dissecting}:
\begin{align}
    \hat{y}_{C} = \tTF\left( \vp_{CoT}^{K-1}(\tilde{f}), \tilde{\vx}, \underbrace{\hat{y}_1, \cdots, \hat{y}_{C-1}}_{\textit{previous step outputs}} \right).
\end{align}
 Intuitively, when using CoT sequences, we allow the model to generate $C-1$ intermediate tokens, followed by the final token $\hat{y}_{pred} = \hat{y}_C$. Given an evaluation dataset of $\widetilde{T}$ sequences, the $\tacc$ is formulated as $
    \tacc = \frac{1}{\widetilde{T}} \sum_{t=1}^{\widetilde{T}} \sI_{\hat{y}_{pred}=\tilde{y}_C}$.

\paragraph{On intermediate tokens.} Since we have access to all the ground truth chain tokens using $\tilde{f}$, we measure the $\tacc$ of predicting them based on $\hat{y}_c, \forall c \in [C]$. In fact, in Section~\ref{subsec:vary_C}, we show a gradual error propagation phenomenon which results in higher $\tacc$ values on tokens at the beginning of the chain and lower $\tacc$ at the end.

\paragraph{Models.} We create three models \texttt{TF-4, TF-8, TF-12} with varying depth based on the Llama-3 architecture \cite{dubey2024llama} for our experiments (see Table~\ref{tab:model_card}). We ensure that depth is the only varying design factor in the architecture to facilitate a systematic study of the model performance.

\paragraph{Learning the `unknown' embeddings $\mE_{\texttt{data}}.$} Recall that the fixed embedding matrix $\mE_{\texttt{data}} \in \sR^{|\gV| \times d}$ that was used to generate the training sequences is unknown to the \texttt{TF} models. To understand the effect of training on the learnable embeddings $\mE_{\texttt{TF}}$ of the \texttt{TF} models, we measure the subspace similarity between the left singular bases of $\mE_{\texttt{data}}$ and $\mE_{\texttt{TF}}$ \cite{zhu2013angles}. Let $d < |\gV|$ and denote SVD of $\mE_{\texttt{data}}$ and $\mE_{\texttt{TF}}$ as follows to obtain:
\begin{align}
\begin{split}
\mU_{\texttt{data}}\mS_{\texttt{data}}\mV^\top_{\texttt{data}} = \mE_{\texttt{data}} ; \hspace{10pt} \mU_{\texttt{TF}}\mS_{\texttt{TF}}\mV^\top_{\texttt{TF}} = \mE_{\texttt{TF}} \\
    \texttt{sim}(\mE_{\texttt{data}}, \mE_{\texttt{TF}}) := \frac{1}{d}\cdot\norm{\mU_{\texttt{data}}^\top[:d]\mU_{\texttt{TF}}[:d]}_F.
\end{split}
\end{align}
Here $\texttt{sim}(\cdot, \cdot)$ allows us to measure how well the subspaces of $\mE_{\texttt{data}}, \mE_{\texttt{TF}}$ are aligned, and a higher value indicates that the $\tTF$ model is learning the token space of the target language embeddings.

\paragraph{Experimental setup.} To present our empirical findings based on training $\tTF$ models on \coticl, we use the following common parameters to create our datasets. We create a training dataset of size $T=32 \times 10^5$, evaluation dataset of size $\widetilde{T}=10^4$ and use $d=10$ along with $\gH(1, \texttt{LeakyReLU})$. In particular, as also mentioned in Section~\ref{sec:cot_icl_lab_data_gen}, we do not put limitations on the cardinality of $\gG, \gH$. See Appendix~\ref{app:hardware_hyperparams} for details on the hardware, training resources and hyper-parameters used for experiments.

\begin{figure*}[t!]
    \centering
    \begin{subfigure}[b]{0.32\textwidth}
        \centering
        \includegraphics[width=\textwidth]{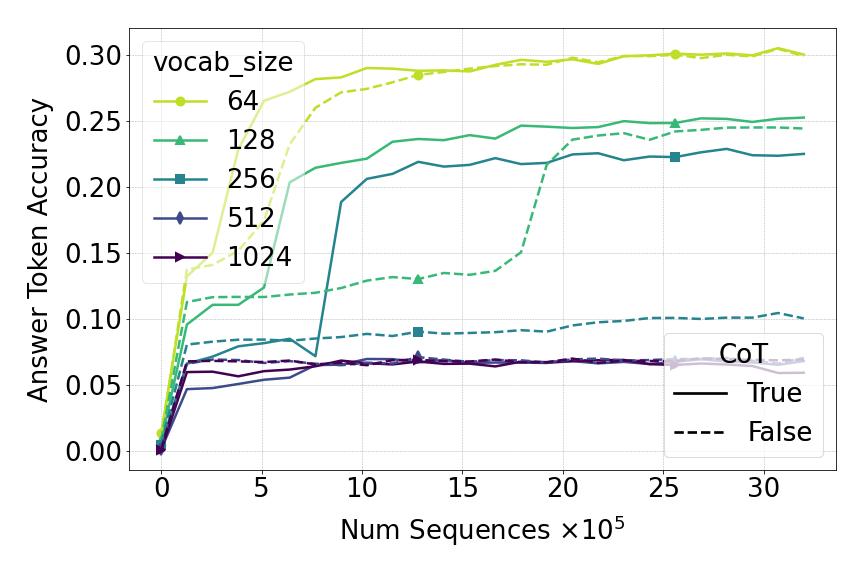}
        \caption{\texttt{TF-4}}
        \label{fig:vary_V_N_4_M_4_C_2_n_ex_30_L_4}
    \end{subfigure}
    \hfill
    \begin{subfigure}[b]{0.32\textwidth}
        \centering
        \includegraphics[width=\textwidth]{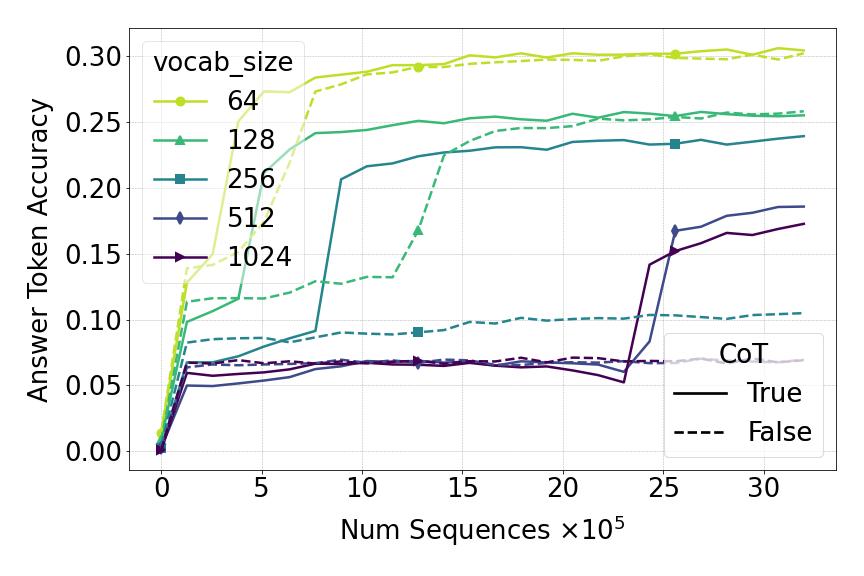}
        \caption{\texttt{TF-8}}
        \label{fig:vary_V_N_4_M_4_C_2_n_ex_30_L_8}
    \end{subfigure}
    \hfill
    \begin{subfigure}[b]{0.32\textwidth}
        \centering
        \includegraphics[width=\textwidth]{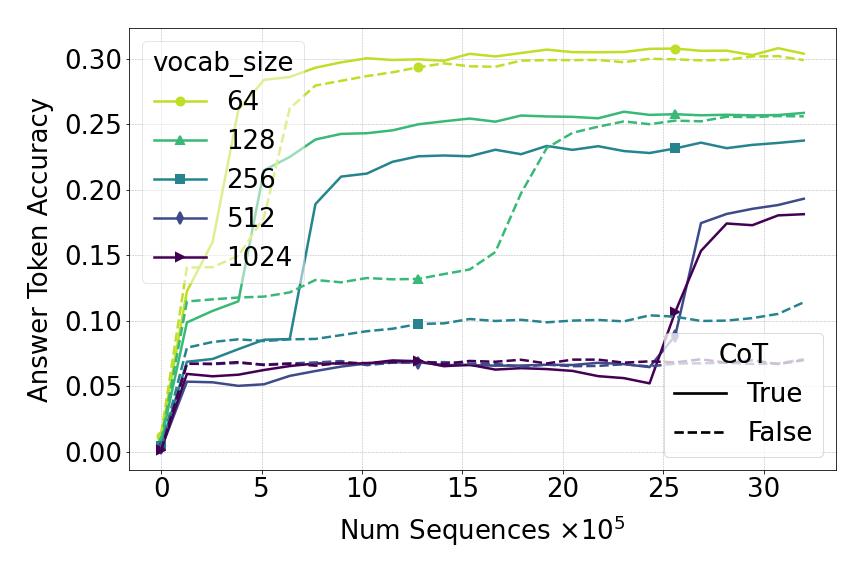}
        \caption{\texttt{TF-12}}
        \label{fig:vary_V_N_4_M_4_C_2_n_ex_30_L_12}
    \end{subfigure}
    \caption{$\tacc$ by varying $\gV$ with $\gG(M=4,N=4,C=2),\gH(1, \texttt{LeakyRelu}), d=10, K=30$. }
    \label{fig:vary_V_N_4_M_4_C_2_n_ex_30}
    \vspace{-3mm}
\end{figure*}

\begin{figure*}[t!]
    \centering
    \begin{subfigure}[b]{0.32\textwidth}
        \centering
        \includegraphics[width=\textwidth]{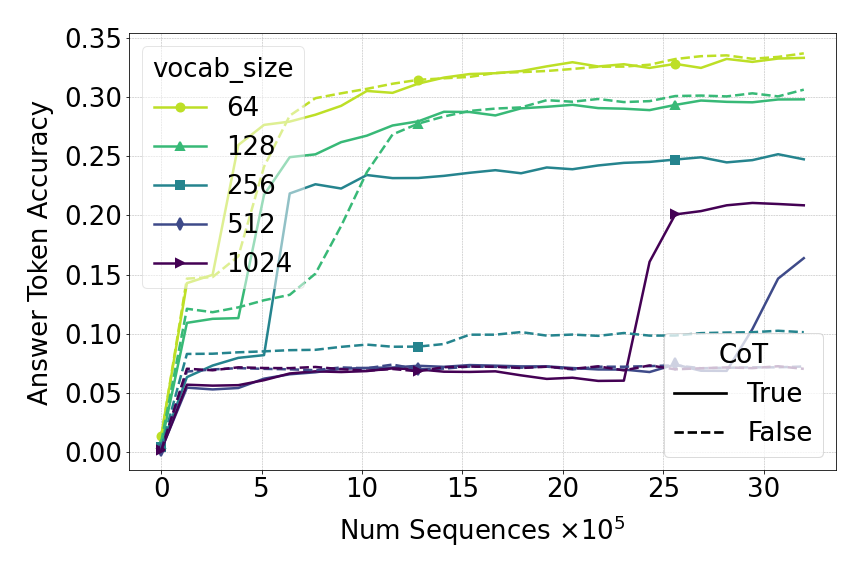}
        \caption{\texttt{TF-4}}
        \label{fig:vary_V_N_4_M_4_C_2_n_ex_40_L_4}
    \end{subfigure}
    \hfill
    \begin{subfigure}[b]{0.32\textwidth}
        \centering
        \includegraphics[width=\textwidth]{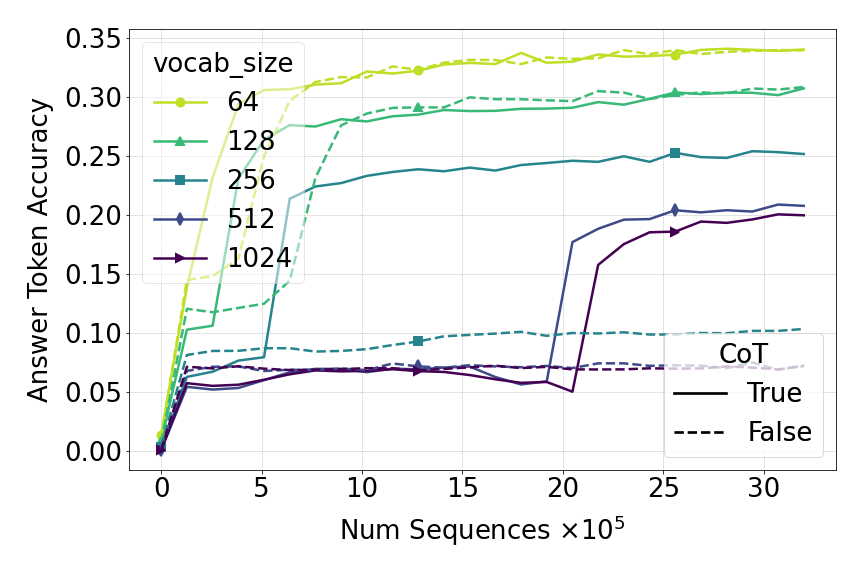}
        \caption{\texttt{TF-8}}
        \label{fig:vary_V_N_4_M_4_C_2_n_ex_40_L_8}
    \end{subfigure}
    \hfill
    \begin{subfigure}[b]{0.32\textwidth}
        \centering
        \includegraphics[width=\textwidth]{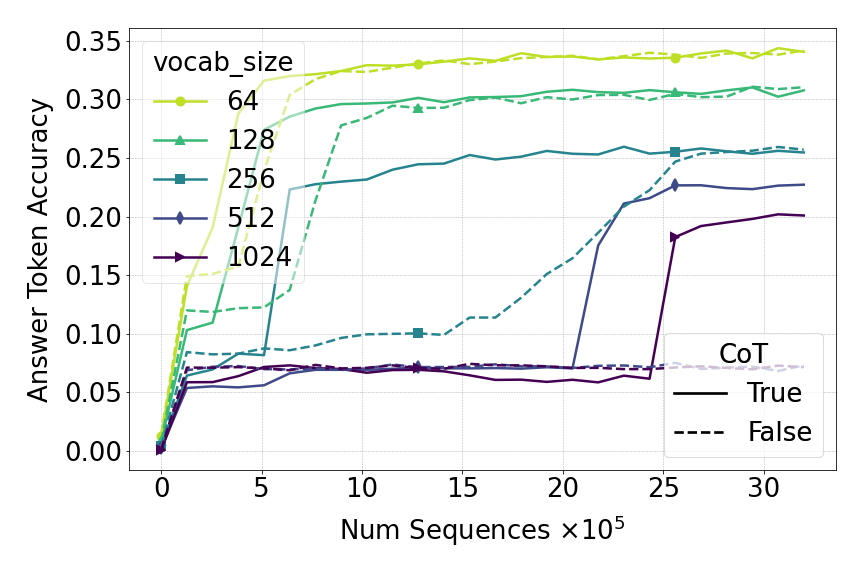}
        \caption{\texttt{TF-12}}
        \label{fig:vary_V_N_4_M_4_C_2_n_ex_40_L_12}
    \end{subfigure}
    \caption{$\tacc$ by varying $\gV$ with $\gG(M=4,N=4,C=2),\gH(1, \texttt{LeakyRelu}), d=10, K=40$.}
    \label{fig:vary_V_N_4_M_4_C_2_n_ex_40}
    \vspace{-3mm}
\end{figure*}

\begin{figure*}[t!]
    \centering
    \begin{subfigure}[b]{0.32\textwidth}
        \centering
        \includegraphics[width=\textwidth]{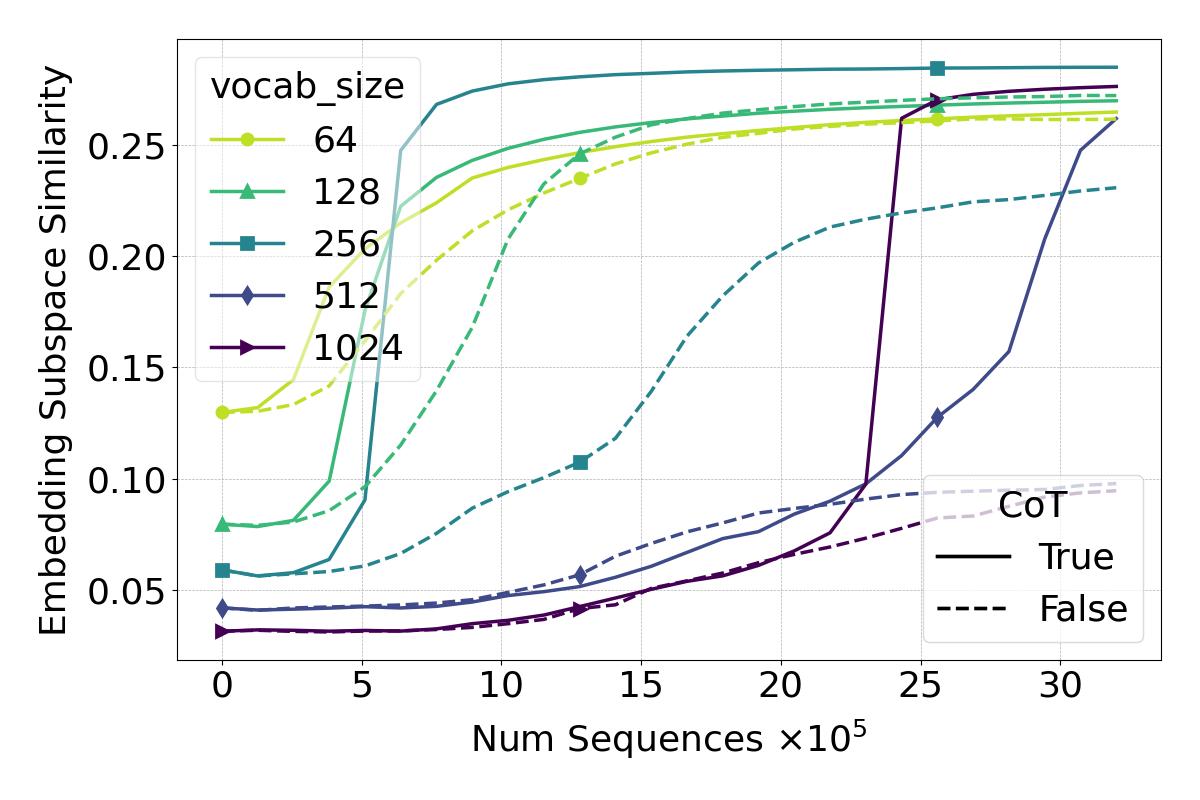}
        \caption{\texttt{TF-4}}
        \label{fig:subspace_sim_vary_V_N_4_M_4_C_2_n_ex_40_L_4}
    \end{subfigure}
    \hfill
    \begin{subfigure}[b]{0.32\textwidth}
        \centering
        \includegraphics[width=\textwidth]{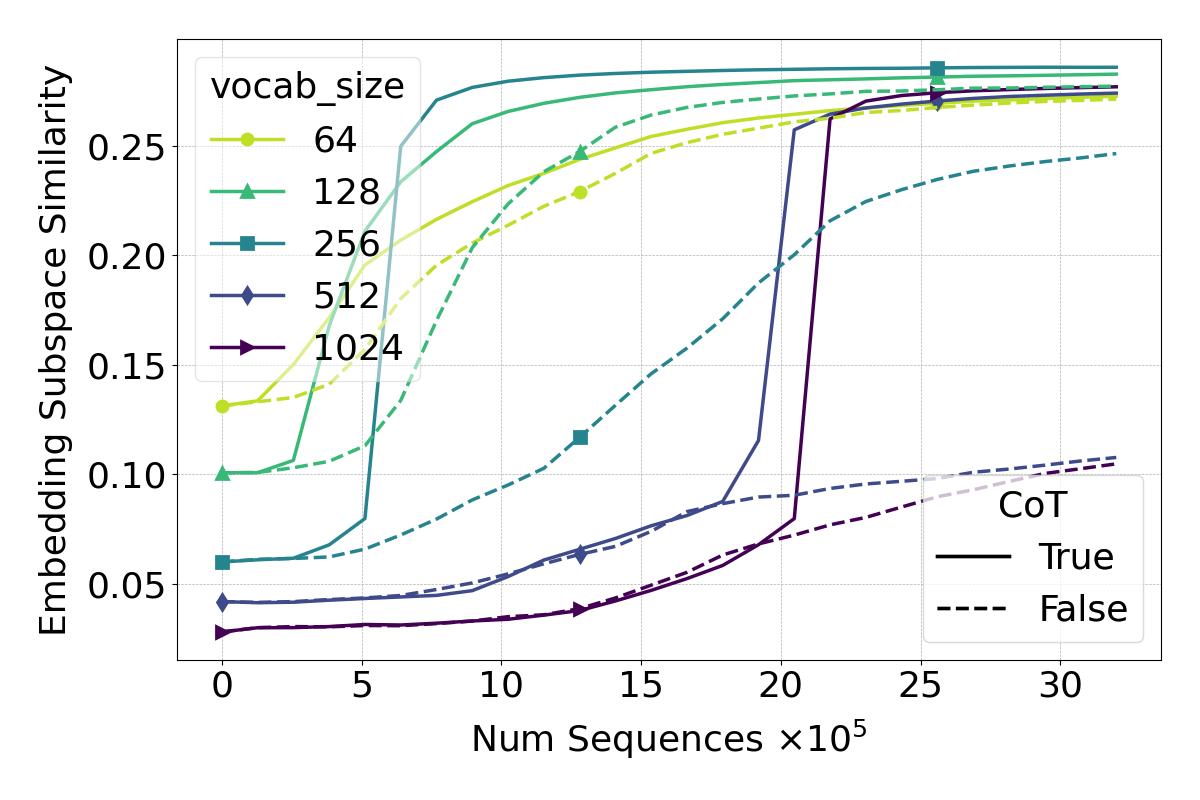}
        \caption{\texttt{TF-8}}
        
    \end{subfigure}
    \hfill
    \begin{subfigure}[b]{0.32\textwidth}
        \centering
        \includegraphics[width=\textwidth]{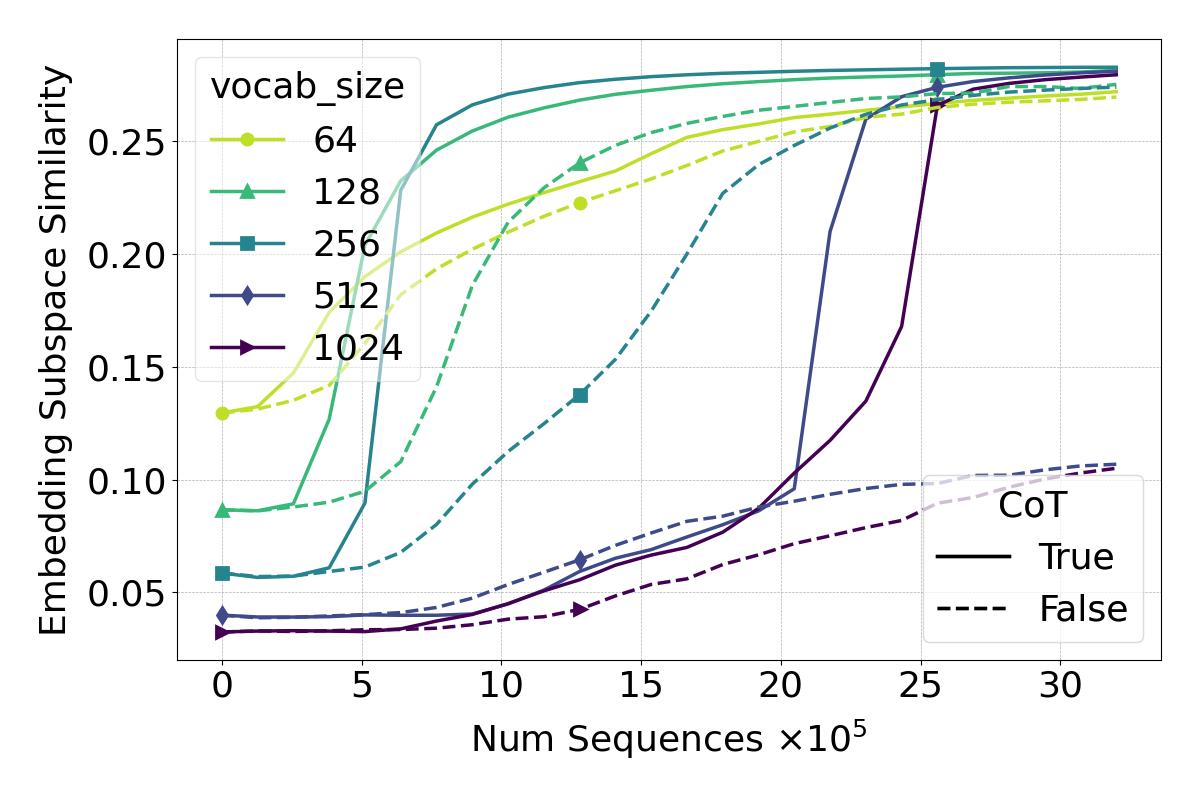}
        \caption{\texttt{TF-12}}
        
    \end{subfigure}
    \caption{$\texttt{sim}(\mE_{\texttt{data}}, \mE_{\texttt{TF}})$ by varying $\gV$ with $\gG(M=4,N=4,C=2),  \gH(1, \texttt{LeakyRelu}), d=10, K=40$.}
    \label{fig:subspace_sim_vary_V_N_4_M_4_C_2_n_ex_40}
    \vspace{-3mm}
\end{figure*}

\section{Results}

\subsection{Effect of Vocabulary Size $|\gV|$}
\label{subsec:vary_V}
As shown in Figure~\ref{fig:cot_icl_intro}, our setup aims to mimic ICL and CoT problems in NLP. However, our synthetic language setup does not have the same priors and rules as tokens in natural language. In this section we test if such a problem is learnable to non-trivial levels of accuracy by $\tTF$ models that only observe patterns in the ICL examples. We vary the size of the vocabulary as per $|\gV|=\{64,128,256,512,1024\}$ along with $N=4,M=4,C=2, K=\{30, 40\}$ and show that the $\tTF$ models achieve non-trivial performance. To the best of our knowledge no prior work has done experiments with synthetic datasets of this vocabulary and model size (see Table~\ref{tab:related_work_comparison}). 

\paragraph{Smaller models fail to leverage CoT for ICL with larger vocabularies.} When $K=30$, observe from Figure~\ref{fig:vary_V_N_4_M_4_C_2_n_ex_30} that towards the end of training, the evaluation $\tacc$ for $|\gV|=\{64, 128\}$, is almost the same for CoT and non-CoT cases across all the three models. However, notice that CoT based ICL results in higher evaluation $\tacc$ after training on relatively less number of sequences than the non-CoT approach, i.e \textbf{CoT results in faster transitions in performance}. More importantly, the benefits of CoT are prominent in $\tTF$-4 when $|\gV|=256$ (see Figure~\ref{fig:vary_V_N_4_M_4_C_2_n_ex_30_L_4}). In this case, CoT based ICL results in a sudden jump in $\tacc$ after training on $\approx 8 \times 10^5$ sequences, but the non-CoT approach fails to exhibit such behavior even towards the end of training. The same observation can be made for $\tTF$-8 (Figure~\ref{fig:vary_V_N_4_M_4_C_2_n_ex_30_L_8}), $\tTF$-12 (Figure~\ref{fig:vary_V_N_4_M_4_C_2_n_ex_30_L_12}). Nonetheless, the benefits of model depth are evident when $|\gV|=\{512, 1024\}$ as $\tTF$-4 fails to leverage CoT with such large vocabularies  whereas $\tTF$-8, $\tTF$-12 clearly show the jumps towards higher $\tacc$ (see Figure~\ref{fig:vary_V_N_4_M_4_C_2_n_ex_30_L_4} vs Figure~\ref{fig:vary_V_N_4_M_4_C_2_n_ex_30_L_8}, Figure~\ref{fig:vary_V_N_4_M_4_C_2_n_ex_30_L_12}). 

\paragraph{More ICL examples facilitate smaller models to leverage CoT.} By increasing $K$ to $40$, we make an interesting observation that even a smaller $\tTF$-4 model can perform on-par with larger $\tTF$-8, $\tTF$-12 models in the difficult setting of $|\gV|=\{512, 1024\}$ (see Figure~\ref{fig:vary_V_N_4_M_4_C_2_n_ex_40}). However, the role of model size comes into play when considering standard non-CoT examples where \texttt{TF-12} gradually improves with $|\gV|=256$ while \texttt{TF-4}, \texttt{TF-8} show a saturated curve (see blue dotted line in Figure~\ref{fig:vary_V_N_4_M_4_C_2_n_ex_40_L_12} vs Figure~\ref{fig:vary_V_N_4_M_4_C_2_n_ex_40_L_4}, Figure~\ref{fig:vary_V_N_4_M_4_C_2_n_ex_40_L_8}). See Appendix~\ref{app:subsec:vary_K} for experiments with smaller $K$ and Appendix~\ref{app:subsec:longer_training} for experiments with training on $3 \times$ larger datasets, which highlight the role of model size for achieving better $\tacc$.

\paragraph{Embedding subspace similarity correlates with transitions in evaluation $\tacc$.} Recall that we employ a common $\mE_{\texttt{data}}$ to prepare our training and evaluation sequences. We noticed that $\texttt{sim}(\mE_{\texttt{data}}, \mE_{\texttt{TF}})$ serves as a useful metric to potentially explain the $\tTF$-4 models earlier transition to a higer evaluation $\tacc$ with $|\gV|=1024$, when compared to $|\gV|=512$ in Figure~\ref{fig:vary_V_N_4_M_4_C_2_n_ex_40_L_4}. From Figure~\ref{fig:subspace_sim_vary_V_N_4_M_4_C_2_n_ex_40_L_4}, notice that $\texttt{sim}(\mE_{\texttt{data}}, \mE_{\texttt{TF}})$ transitions to a higher value after $\approx 23 \times 10^5$ steps, which exactly coincides with the transition point in Figure~\ref{fig:vary_V_N_4_M_4_C_2_n_ex_40_L_4}. Similarly, the delay in alignment with $|\gV|=512$ in Figure~\ref{fig:subspace_sim_vary_V_N_4_M_4_C_2_n_ex_40_L_4} is indicative of a delay in the evaluation $\tacc$ transition in Figure~\ref{fig:vary_V_N_4_M_4_C_2_n_ex_40_L_4}. 
This is an interesting and unique aspect of \coticl~that allows us to interpret and understand how the models learn relationships between the concepts in the synthetic language.

\begin{figure*}[t!]
    \centering
    \begin{subfigure}[b]{0.32\textwidth}
        \centering
        \includegraphics[width=\textwidth]{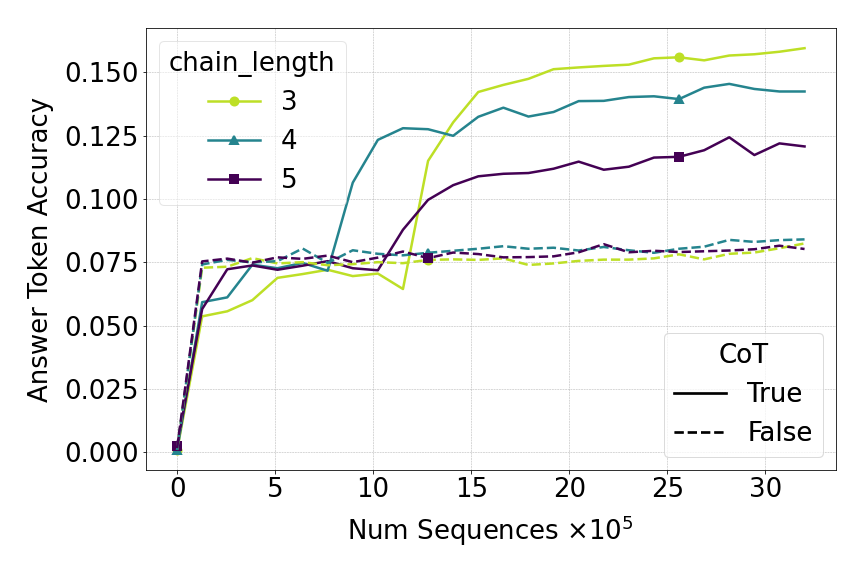}
        \caption{\texttt{TF-4}}\vspace{-10pt}
        \label{fig:vary_C_N_4_M_4_V_1024_d_10_n_ex_40_L_4}
    \end{subfigure}
    \hfill
    \begin{subfigure}[b]{0.32\textwidth}
        \centering
        \includegraphics[width=\textwidth]{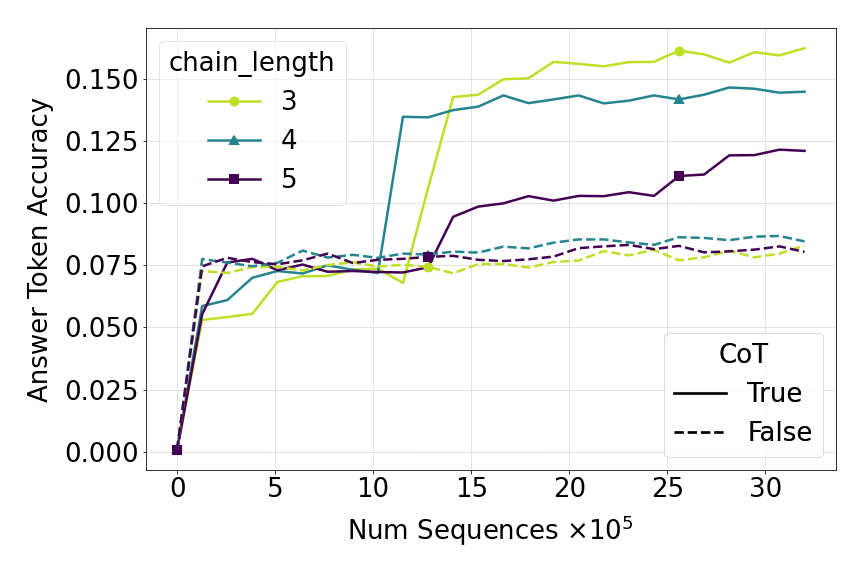}
        \caption{\texttt{TF-8}}\vspace{-10pt}
        \label{fig:vary_C_N_4_M_4_V_1024_d_10_n_ex_40_L_8}
    \end{subfigure}
    \hfill
    \begin{subfigure}[b]{0.32\textwidth}
        \centering
        \includegraphics[width=\textwidth]{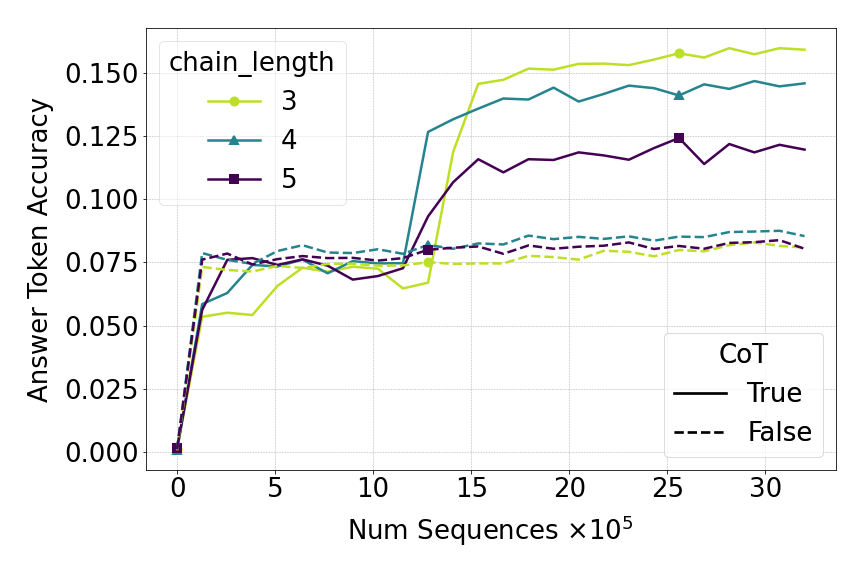}
        \caption{\texttt{TF-12}}\vspace{-10pt}
\label{fig:vary_C_N_4_M_4_V_1024_d_10_n_ex_40_L_12}
    \end{subfigure}
    \caption{$\tacc$ by varying $C$ with $\gG(N=4,M=4),\gH(1, \texttt{LeakyRelu}), d=10,|\gV|=1024,K=40$. }
    \label{fig:vary_C_N_4_M_4_V_1024_d_10_n_ex_40}
    \vspace{-2mm}
\end{figure*}

\begin{figure}[t!]
    \centering
    \begin{subfigure}[b]{0.32\textwidth}
        \centering
        \includegraphics[width=\textwidth]{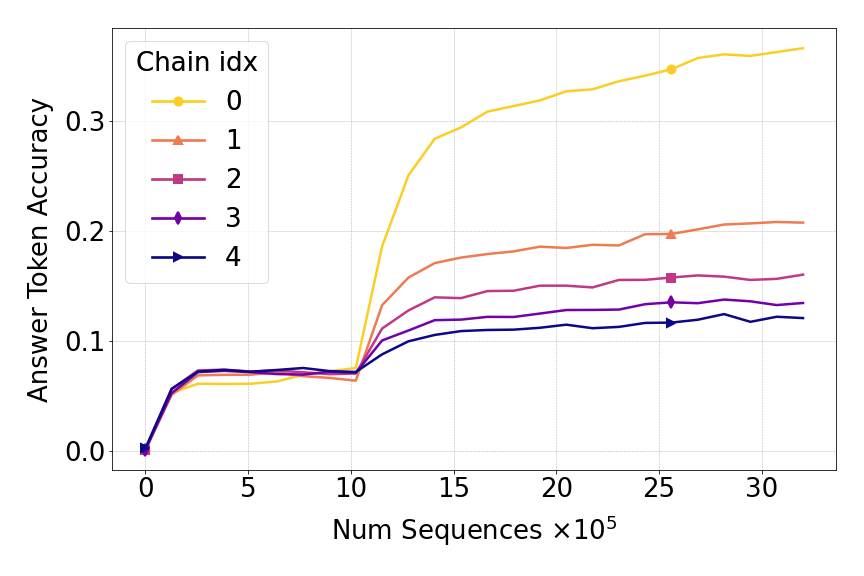}\vspace{-10pt}
        \caption{\texttt{TF-4}}
        
    \end{subfigure}
    \hfill
    \begin{subfigure}[b]{0.32\textwidth}
        \centering
        \includegraphics[width=\textwidth]{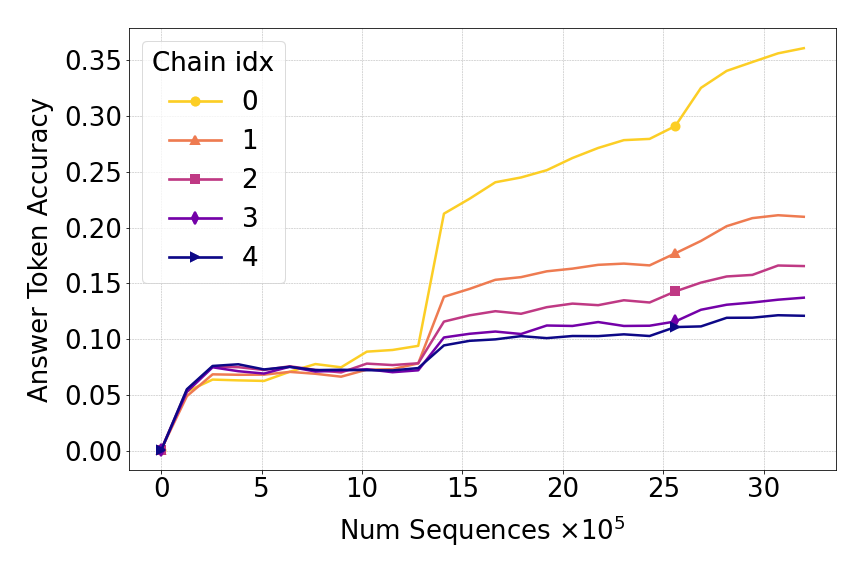}\vspace{-10pt}
        \caption{\texttt{TF-8}}
        
    \end{subfigure}
    \hfill
    \begin{subfigure}[b]{0.32\textwidth}
        \centering
        \includegraphics[width=\textwidth]{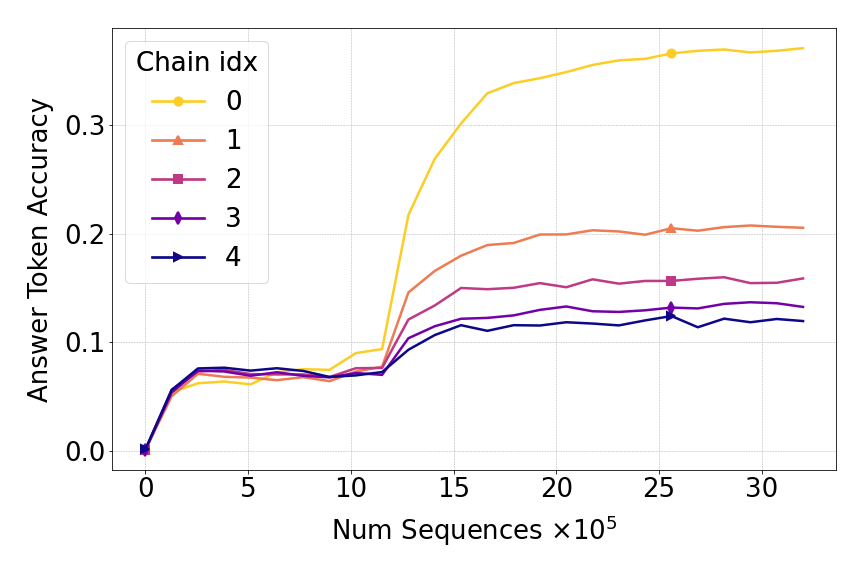}\vspace{-10pt}
        \caption{\texttt{TF-12}}
        
    \end{subfigure}
    \caption{$\tacc$ of predicting all the \textit{chain tokens} (of the query \textit{CoT example} in the evaluation sequences) with $\gG(N=4,M=4, C=5), \gH(1, \texttt{LeakyRelu}), d=10,|\gV|=1024,K=40$. 
    }
\label{fig:error_propagation_C_5_N_4_M_4_V_1024_d_10_n_ex_40}
    \vspace{-2mm}
\end{figure}

\subsection{Effect of Chain Length $(C)$}
\label{subsec:vary_C}
The number of intermediate steps involved in the reasoning process is typically indicative of the complexity of NLP tasks. Its counterpart in our \coticl~is the chain length $C$. By choosing $|\gV|=1024, N=4,M=4$ and varying $C=\{3, 4, 5\}$, we examine and show that longer chains result in harder problems that the models without CoT would not be able to solve effectively (Figure~\ref{fig:vary_C_N_4_M_4_V_1024_d_10_n_ex_40}).

\paragraph{Longer chains result in lower $\tacc$ across all model sizes.} As the chain length increases, we can observe from Figure~\ref{fig:vary_C_N_4_M_4_V_1024_d_10_n_ex_40} that the evaluation $\tacc$ is consistently lower for all models. Since the number of possibilities for the parents of the last chain token increases with $C$, we hypothesize that the models fail to adapt to such increased difficulty. This is indeed what we observe in Figure~\ref{fig:error_propagation_C_5_N_4_M_4_V_1024_d_10_n_ex_40} where the $\tacc$ consistently goes down for tokens towards the end of the chain across all model sizes.

\subsection{Effect of Number of Parent Tokens $(M)$}
\label{subsec:vary_M}

\begin{figure*}[t!]
    \centering
    \begin{subfigure}[b]{0.32\textwidth}
        \centering
        \includegraphics[width=\textwidth]{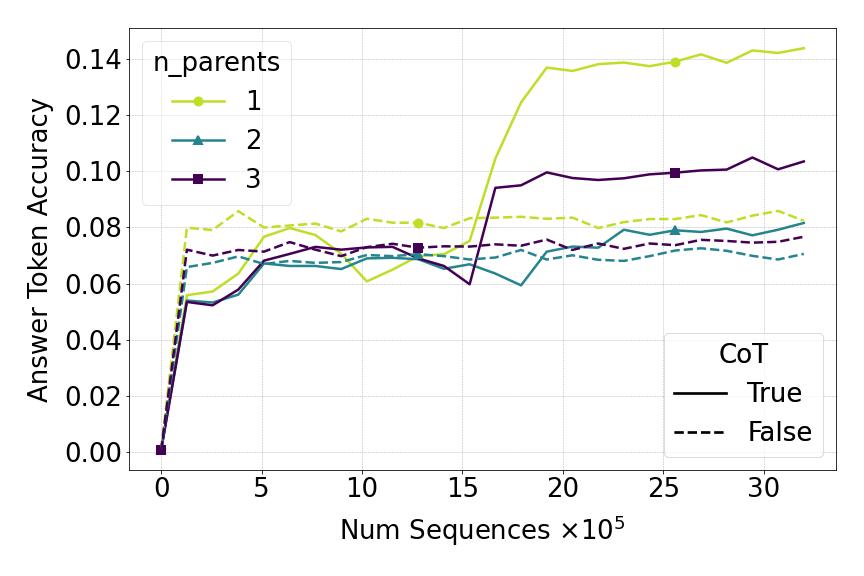}
        \caption{\texttt{TF-4}}
        \label{fig:vary_M_N_4_C_4_V_1024_d_10_n_ex_40_L_4}
    \end{subfigure}
    \hfill
    \begin{subfigure}[b]{0.32\textwidth}
        \centering
        \includegraphics[width=\textwidth]{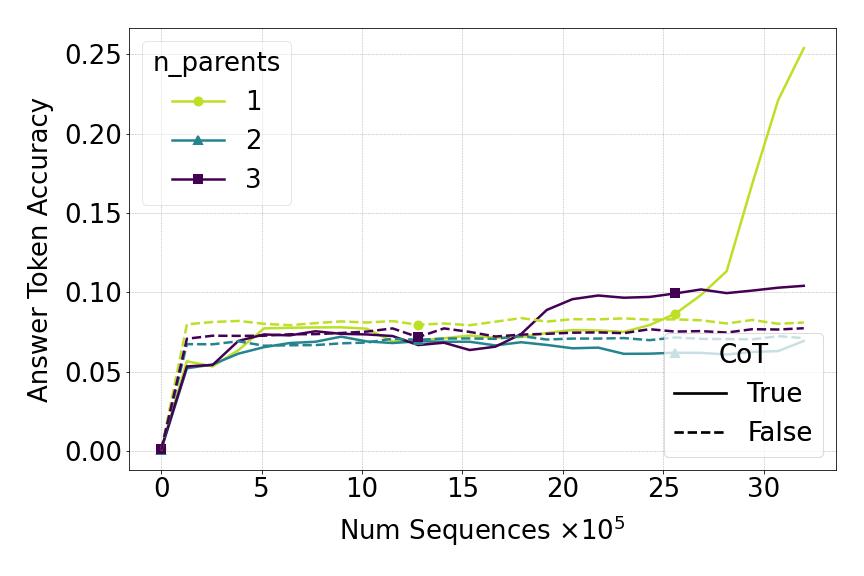}
        \caption{\texttt{TF-8}}
        \label{fig:vary_M_N_4_C_4_V_1024_d_10_n_ex_40_L_8}
    \end{subfigure}
    \hfill
    \begin{subfigure}[b]{0.32\textwidth}
        \centering
        \includegraphics[width=\textwidth]{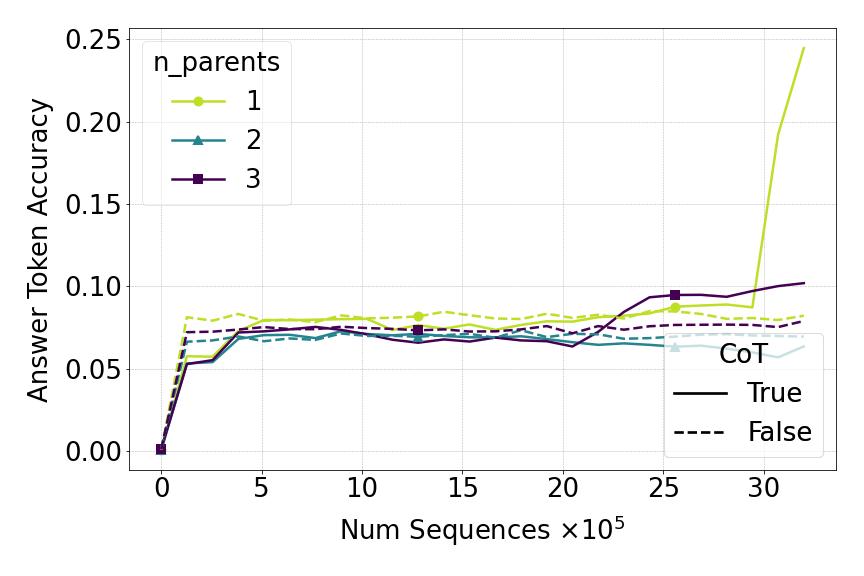}
        \caption{\texttt{TF-12}}
        \label{fig:vary_M_N_4_C_4_V_1024_d_10_n_ex_40_L_12}
    \end{subfigure}
    \caption{$\tacc$ by varying $M$ with $\gG(N=4,C=4),\gH(1, \texttt{LeakyReLU}), d=10,|\gV|=1024,K=40$. 
    }
    \label{fig:vary_M_N_4_C_4_V_1024_d_10_n_ex_40}
    \vspace{-3mm}
\end{figure*}

The dependency between tokens in NLP tasks is a variable that is usually not controllable (see Figure~\ref{fig:cot_icl_intro}). But in \coticl~it can be controlled via $M$.  By choosing $|\gV|=1024, N=4,C=4$ and varying $M=\{1, 2, 3\}$, we observe that larger $M$ (i.e dependency on more prior tokens) can make the problem harder for the models (see Figure~\ref{fig:vary_M_N_4_C_4_V_1024_d_10_n_ex_40}).

\paragraph{Large models outperform small ones when DAGs are sparse.} When $M=1$, we make an interesting observation that $\tTF$-8 (Figure~\ref{fig:vary_M_N_4_C_4_V_1024_d_10_n_ex_40_L_8}), $\tTF$-12 (Figure~\ref{fig:vary_M_N_4_C_4_V_1024_d_10_n_ex_40_L_12}) models exploit CoT towards the end of training to significantly outperform the $\tTF$-4 model (Figure~\ref{fig:vary_M_N_4_C_4_V_1024_d_10_n_ex_40_L_4}). This shows an interesting difference in the training dynamics of deeper models, which could be an interesting research question for future works.

\begin{figure*}[t!]
    \centering
    \begin{subfigure}[b]{0.32\textwidth}
        \centering
        \includegraphics[width=\textwidth]{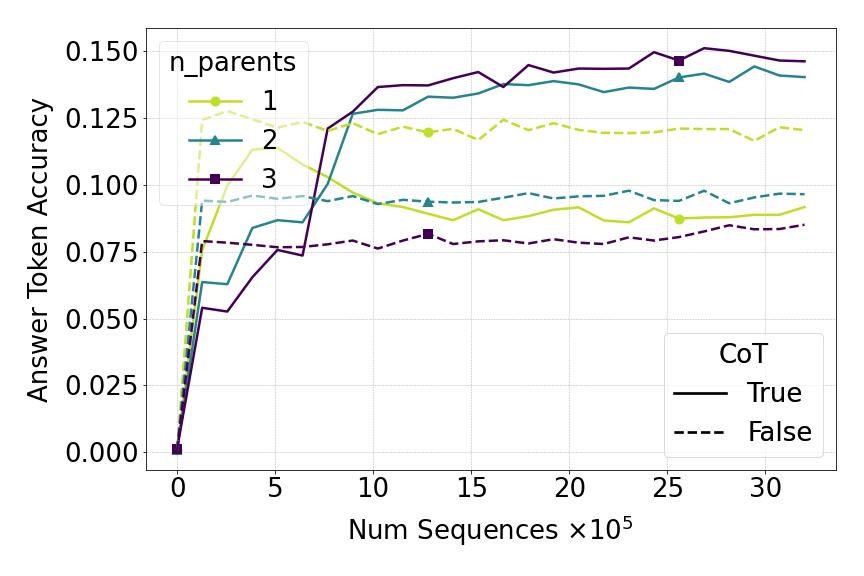}
        \caption{\texttt{TF-4} (Fixed DAG)}
        \label{fig:ablation_H_G_fixed_random_dag_vary_M_N_4_C_4_V_1024_d_10_n_ex_40_L_4}
    \end{subfigure}
    \hfill
    \begin{subfigure}[b]{0.32\textwidth}
        \centering
        \includegraphics[width=\textwidth]{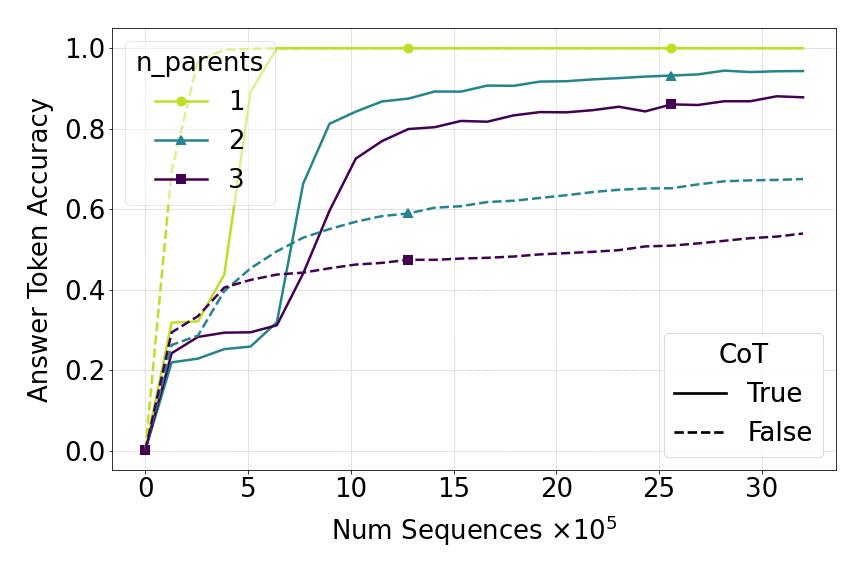}
        \caption{\texttt{TF-4} (Fixed $4$ MLPs)}
        \label{fig:ablation_H_G_fixed_H_vary_M_N_4_C_4_V_1024_d_10_n_ex_40_L_4}
    \end{subfigure}
    \hfill
    \begin{subfigure}[b]{0.32\textwidth}
        \centering
        \includegraphics[width=\textwidth]{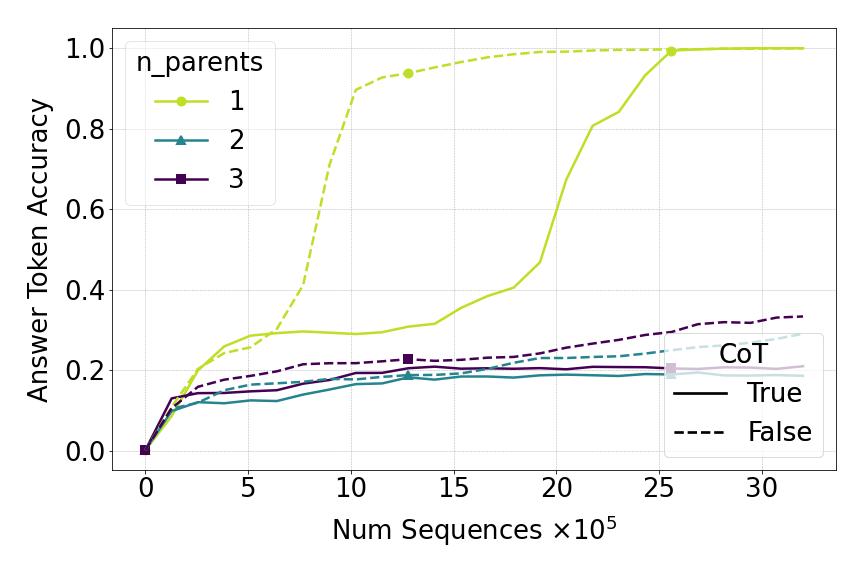}
        \caption{\texttt{TF-4} (Fixed $40$ MLPs)}
\label{fig:ablation_H_G_fixed_H_10_vary_M_N_4_C_4_V_1024_d_10_n_ex_40_L_4}
    \end{subfigure}
    \hfill
    \caption{Ablation experiments with fixed DAG (a) and Fixed MLPs (b, c) for the \texttt{TF-4} model by measuring $\tacc$ with varying $M$ and $\gG(N=4,C=4),\gH(1, \texttt{LeakyRelu}),d=10,|\gV|=1024,K=40$.}
    \label{fig:ablation_H_G_vary_M_N_4_C_4_V_1024_d_10_n_ex_40}
    \vspace{-3mm}
\end{figure*}

\subsection{Ablations with $\gG$ and $\gH$}
\label{subsec:ablations_G_H}
Our \coticl~datasets, in their most general form, i.e. no restrictions on the cardinality of $\gG$ and $\gH$, are quite difficult for the models to solve (based on the low evaluation $\tacc$ observed in prior sections). Owing to the flexibility of our setup, we decouple the effect of $\gG$ and $\gH$ to better understand the source of difficulty for the ICL problems.

\paragraph{$\bullet$ Fixed DAG structure.} We follow the same setup as Section~\ref{subsec:vary_M} and sample different token processing MLPs but choose a fixed random DAG for all the training and validation sequences. 

\paragraph{$\bullet$ Fixed token processors.} Contrary to the above case, we sample random DAGs for different sequences but instead choose $C$ fixed MLPs, one per chain location, as our token processors for all sequences. 

\vspace{2mm}
\noindent By comparing Figure~\ref{fig:ablation_H_G_fixed_random_dag_vary_M_N_4_C_4_V_1024_d_10_n_ex_40_L_4}, for fixed DAG, and Figure~\ref{fig:ablation_H_G_fixed_H_vary_M_N_4_C_4_V_1024_d_10_n_ex_40_L_4} for fixed token processors, we can observe that the {\bf models reach higher accuracies in the fixed token processor setting}. This points to the possibility that the dense transformer models we consider here can identify the causal structure of the chain generation process with ease. We verify this by analyzing the attention map of the models trained in the fixed token processor setting (explained below). Furthermore, this gives us another lever to adjust the difficulty of synthetic datasets in \coticl, i.e adjusting the number of possible token processors (from infinite in the general case to a finite set). See Appendix~\ref{app:subsec:ablation_fixed_G_H} for experiments with \texttt{TF-8} and \texttt{TF-12}.

\subsubsection{Attention Maps reflect the DAG}

\begin{figure*}[t!]
    \centering
    \begin{subfigure}[b]{0.48\textwidth}
        \centering
        \includegraphics[width=\textwidth]{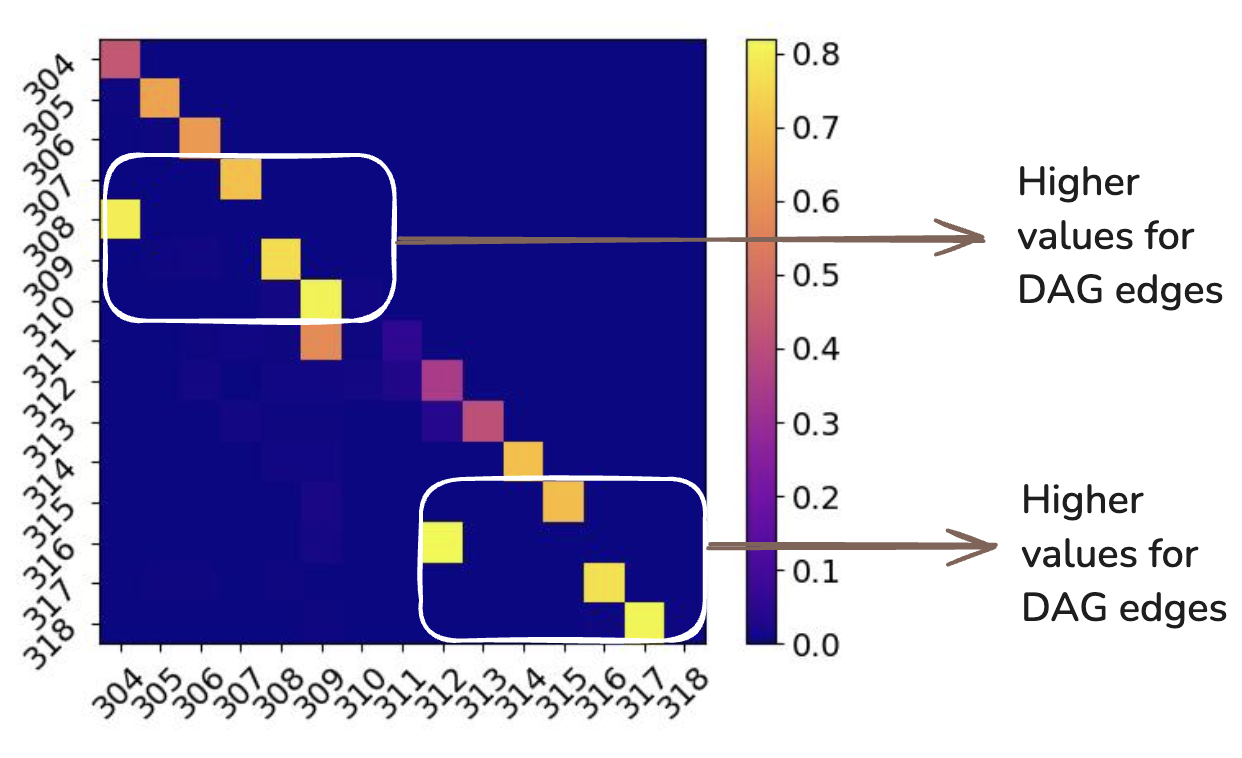}
    \caption{Average attention map $\mA$}
\label{fig:fixed_H_all_head_attn_TF_4_M_1_N_4_C_4_layer_3_zoom}
    \end{subfigure}
    \begin{subfigure}[b]{0.44\textwidth}
        \centering
     \includegraphics[width=\textwidth]{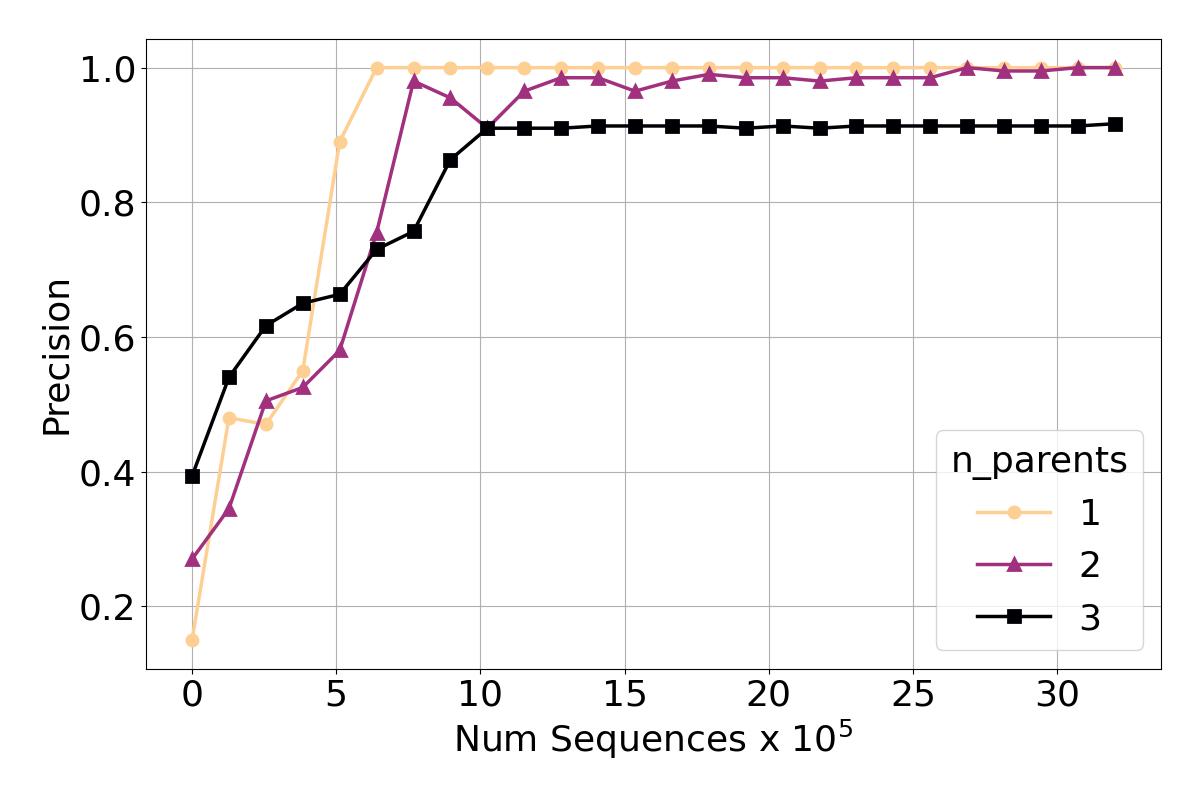}
    \caption{\texttt{Precision} during training }
\label{fig:fixed_H_all_head_attn_TF_4_vary_M_N_4_C_4_precision}
    \end{subfigure}
    \caption{(Left) Mean attention matrix (last $16$ rows and columns) of all heads for the final layer of a trained \texttt{TF-4} model. The highlighted regions show that the model is attending to the correct parent tokens $y_1 \leftarrow \{x_4\}, y_2 \leftarrow \{x_1\},y_3 \leftarrow \{y_1\}, y_4 \leftarrow \{y_2\}$ to generate the chain tokens. (Right) \texttt{Precision} in detecting the parent tokens using average attention map $\mA$ of a trained \texttt{TF-4} model.}
\end{figure*}

We consider models trained in the fixed token processor setting and analyze the attention map $\mA$ averaged across all the heads $\mA_h ,h \in [H]$ of the last layer. Formally $\mA := \frac{1}{H} \sum_{h=1}^{H} \mA_h$. We plot such $\mA$ for the last two examples of a single validation sequence in  Figure~\ref{fig:fixed_H_all_head_attn_TF_4_M_1_N_4_C_4_layer_3_zoom} for $M=1$. Notice that most of the attention scores are almost zero and the ones with large values correspond to the parent tokens of each chain token, i.e. it matches the underlying DAG structure for this sequence. See Appendix~\ref{app:sec:attention_maps} for illustrations with $M=2,3$. 

\paragraph{Quantitative measurement via \texttt{Precision}.} To quantify the attention map analysis to diverse settings, we calculate the \texttt{Precision} of identifying the parent tokens which are needed to generate the final answer token in the evaluation sequences. Let us assume that $G$ is the set of ground truth parent tokens to formulate \texttt{Precision} as follows:
\begin{align}
\begin{split}
    \mA_{\text{query}} &:= \mA[-1, -(N + C - 1):] \\
    \tilde{G} &:= \texttt{argsort}(\mA_{\text{query}})[-M:] \\
    \texttt{Precision} &:= \frac{\sum_{i \in \tilde{G}} \sI(i \in G) }{M}.
\end{split}
\end{align}
Figure~\ref{fig:fixed_H_all_head_attn_TF_4_vary_M_N_4_C_4_precision} shows that as the model accuracy rises in this setting, the \texttt{Precision} of detecting the right DAG structure increases as well. 
It is worth noting that we repeated such an analysis for datasets with infinitely possible token processors but could not find such patterns. We believe that formalizing advanced tools for understanding how/if models detect causal structures is an important research area.       

\paragraph{Patterns beyond the `induction heads'.} Previous works such as \cite{olsson2022context,edelman2024the,nichani2024how} typically relied on `attention-layer' only models to analyze the `induction head' characteristics for improved ICL performance. Our analysis goes beyond such prefix matching and copying mechanisms of induction heads. In particular, our analysis of the attention maps and the \texttt{Precision} indicates that practical transformer models are capable of inferring the causal structure even without any prefix matching mechanisms. Thus, indicating a deeper connection between the underlying reasoning process via chain tokens and the level of diversity in the token processing functions that can facilitate such learning of the causal structures.

\subsubsection{On Finite Token Processors}

So far, we have considered the extreme cases of sampling from infinite token processors for every sequence or sample once and keep them fixed for all the sequences. Next, we show that by controlling the diversity of sequences by using a fixed collection ($40$ in this case, i.e $10$ tuples of $C=4$ MLPs.) of token processors to choose from and generate the sequences, the $\tTF$ models require relatively more training sequences, compared to the fixed token processor setting above, to transition to higher accuracies (see Figure~\ref{fig:ablation_H_G_fixed_H_10_vary_M_N_4_C_4_V_1024_d_10_n_ex_40_L_4}). Intuitively, the final accuracy of the model for this setting is higher than the infinite token processor case (Figure~\ref{fig:vary_M_N_4_C_4_V_1024_d_10_n_ex_40_L_4}) and lower than the fixed token processor case (Figure~\ref{fig:ablation_H_G_fixed_H_vary_M_N_4_C_4_V_1024_d_10_n_ex_40_L_4}).

\section{Connection to NLP}

Tokenized prompts associated with NLP tasks are grounded in real-world knowledge. But due to the unknown causal structure of the underlying data it is hard to do controlled experiments with these datasets. On the other hand, although the synthetic tokenized sequences in~\coticl~are not associated with any grounded semantic information, we have observed some intricate connections between~\coticl~and NLP data by conducting additional experiments on NLP pre-trained models. Specifically, we show that models that are pre-trained on NLP learn \coticl~tasks better and faster. Moreover, when examining the NLP models' attention maps, while performing reasoning on math tasks, we see sparse patterns in attentions between tokens. This motivates the design choice of using sparse causal structures in \coticl.

\subsection{Training NLP Pre-Trained Models on \coticl}

\paragraph{Setup.} We consider the open-source pre-trained \texttt{Llama-3.2-1B-Instruct} model (from HuggingFace) and its random weight counterpart. The \texttt{model.resize\_token\_embeddings()} API is used to restrict the vocabulary size to $1024$ for both models. The models are then trained on our synthetic~\coticl~dataset using the same experimental setup as in Section~\ref{subsec:vary_V} with $\gG(M=4,N=4,C=2),  \gH(1, \texttt{LeakyRelu}), d=10, K=40$ and $|\gV|=1024$. Note that the vocabulary $\gV$ and the data embeddings $\mE_{\texttt{data}}$ used in the~\coticl~does not have any connection to the initial embeddings of the pre-trained model. 

\begin{wrapfigure}{r}{0.5\textwidth}  
    \centering
    
         \includegraphics[width=0.5\textwidth]{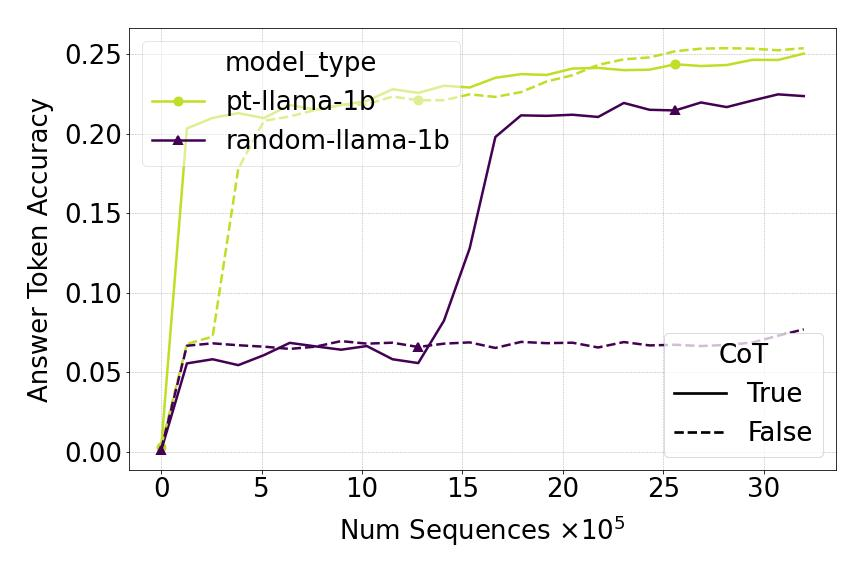}
    
    \caption{$\tacc$ of pre-trained and randomly initialized \texttt{Llama-3.2-1B-Instruct} models on \coticl~dataset with $|\gV|=1024$, $\gG(M=4,N=4,C=2),\gH(1, \texttt{LeakyRelu}), d=10, K=40$.}
    \label{fig:pt_vs_random_llama_1B}
\end{wrapfigure}

\paragraph{NLP pre-trained models transition faster than random counterparts.} As shown in Figure~\ref{fig:pt_vs_random_llama_1B} the pre-trained model transitions to a higher \tacc~in the early phases of training with CoT sequences. On the contrary the random model exhibits such a transition after being trained on $\approx 13 \times 10^5$ sequences with CoT. Towards the end of training, the pre-trained model has an overall \tacc~of $0.25$, compared to $0.22$ for the randomly initialized model.

In the non-CoT training case, the pre-trained model has a final \tacc~of $0.25$ and clearly outperforms the random counterpart (which has an \tacc~of $0.08$). Note that the randomly initialized model does not exhibit a transition throughout training with non-CoT sequences. These results highlight that the pre-trained model has both higher accuracy w/ and w/o CoT and exhibits faster transition/improvement during training on~\coticl~data (with the difference being significant when CoT is disabled). Thus, hinting at a much deeper and interesting connection/similarity between the patterns that the Llama model learnt from natural language and the ones it is being trained and evaluated on in the synthetic setup. 

\subsection{On the Sparsity of Attention Maps in Reasoning LLMs}

A key aspect of \coticl~is the freedom it provides to control the sparsity of DAGs while creating the tokenized sequences. Drawing parallels with NLP, we aim to gain insights by analyzing the mean attention maps of the last layer of reasoning LLMs when prompted with math questions from the MMLU dataset \cite{hendrycks2021measuring}.

\paragraph{Setup.} We prompt the \texttt{DeepSeek-R1-Distill-Llama-8B} reasoning model to answer a high school mathematics question from the MMLU dataset. We employ the \texttt{GenerationConfig} API in the \texttt{transformers} library to generate the response with a token limit of $1024$ and various temperature values. For visualization purposes, we apply thresholding on the mean attention map using a value of $10^{-3}$. The attention scores greater than the threshold are assigned a value of $1$ and the rest are set to $0$.

\begin{figure*}[t!]
    \centering
    \begin{subfigure}[b]{0.48\textwidth}
        \centering
        \includegraphics[width=\textwidth]{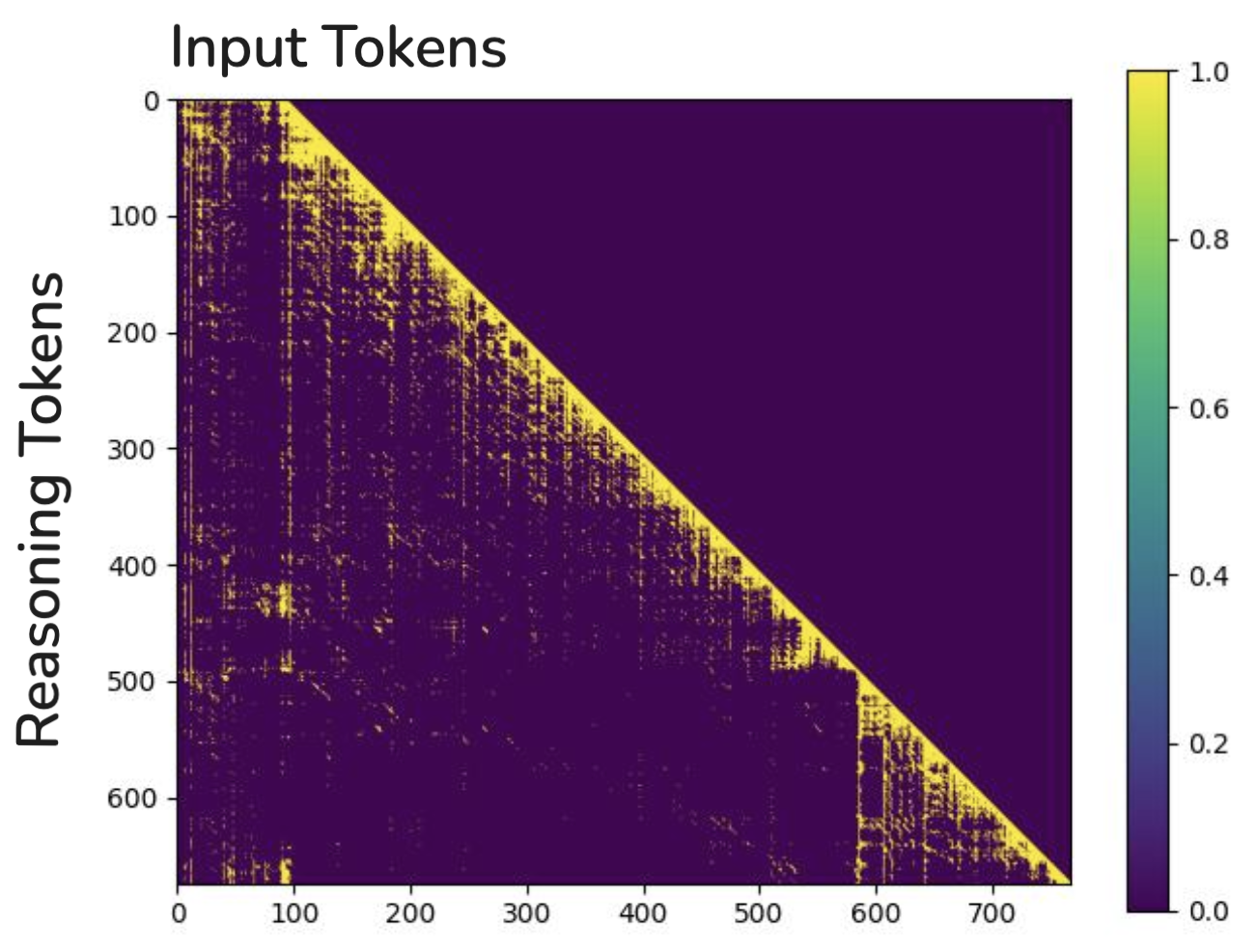}
    \caption{\texttt{temperature=0.6}}
    \end{subfigure}
    \begin{subfigure}[b]{0.47\textwidth}
        \centering
     \includegraphics[width=\textwidth]{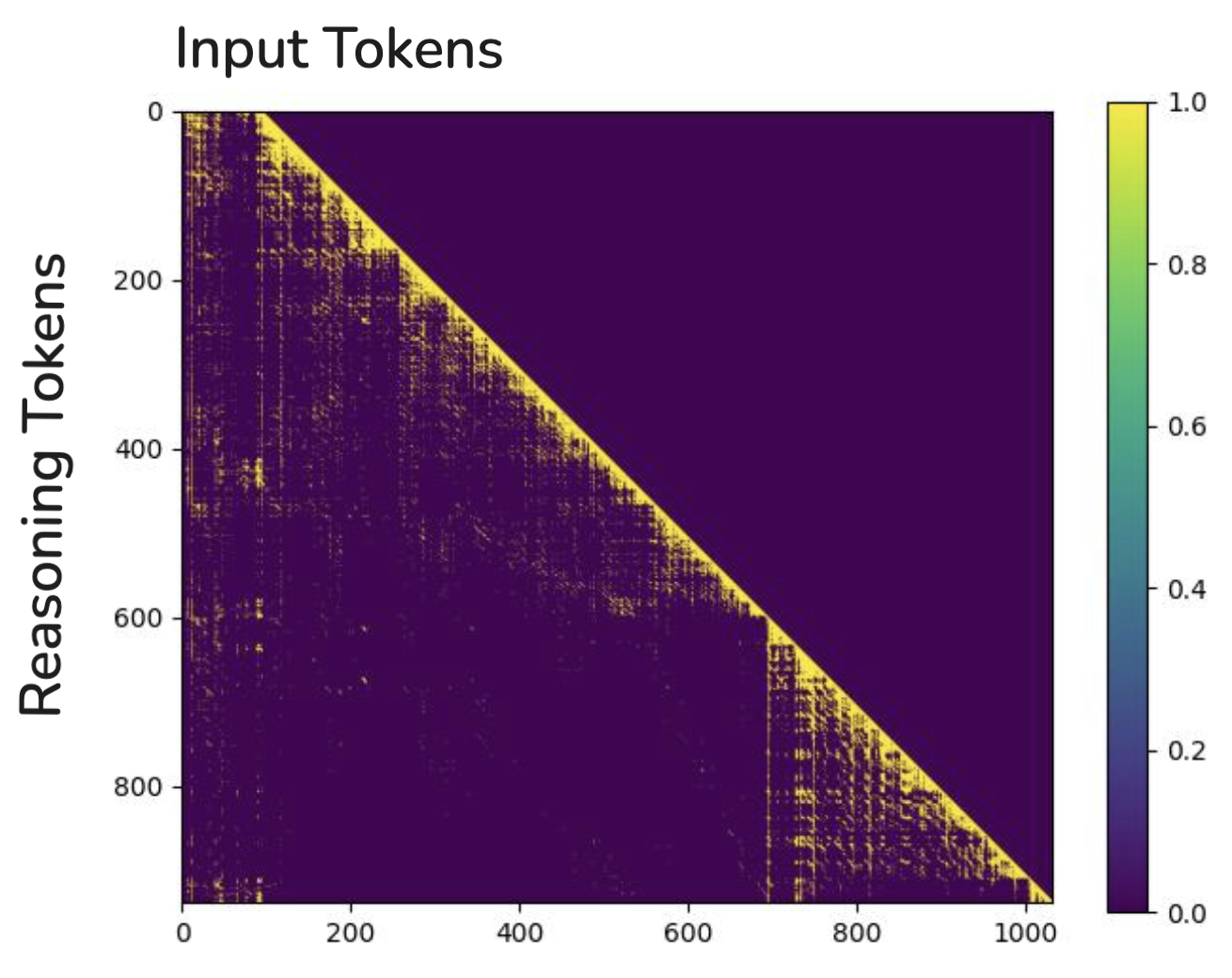}
    \caption{\texttt{temperature=0.9} }
    \end{subfigure}
    \caption{Mean attention matrix of all heads for the final layer of the \texttt{DeepSeek-R1-Distill-Llama-8B} model when generating the reasoning output for a high school mathematics question from MMLU.}
\label{fig:deepseek_llama_attn_map}
\end{figure*}

\paragraph{The binarized attention maps of \texttt{DeepSeek-R1-Distill-Llama-8B} can be extremely sparse.} We plot the mean attention maps corresponding to the reasoning (output) tokens in Figure~\ref{fig:deepseek_llama_attn_map}. Notice that the \texttt{temperature} not only affects the reasoning output lengths, but the attention patterns as well. Interestingly, in both the scenarios, notice that the attention scores tend to concentrate around the immediately previous tokens in the reasoning chain. The input prompt and reasoning outputs are presented in Appendix~\ref{app:reasoning_outputs_deepseek_llama}.

\section{Conclusion}

In this paper, we introduced \coticl, a framework to generate synthetic Chain-of-Thought sequences and systematically study the role of CoT for In-Context Learning tasks. We used the flexibility and controllability of this framework to (1) generate synthetic multi input-output ICL datasets, (2) design controlled experiments to gain better insights into the role of CoT for ICL and (3) interpret the model behavior via embeddings and attention map analysis. We believe these insights, and many more that could be extracted by experimenting with \coticl, would play a crucial role in better understanding CoT for ICL in NLP, which is of utmost importance for the success of large language models.

\section{Limitations}

While \coticl~is designed to closely mirror the chain-of-thought process in in-context learning for NLP, we acknowledge that its synthetic nature does not fully capture the linguistic properties of natural language. Specifically, \coticl~tokens are not grounded in real-world concepts and therefore do not inherently align with the priors that govern natural language tokens. Consequently, researchers utilizing \coticl~for experimentation should carefully weigh its advantages—such as flexibility and controllability—against its limitations, particularly its synthetic nature, and consider the potential impact on their results.

\bibliographystyle{unsrtnat}
\bibliography{references} 

\newpage
\appendix

\section{Real World CoT Datasets and \coticl}
\label{app:cot_icl_datasets}

In this section, we provide additional comprehensive details about the \coticl~setup and draw parallels with real world NLP CoT datasets.

\subsection{Algorithm for Token Processing}
\label{app:alg_H}

Algorithm~\ref{alg:tok_proc_H} formalizes the token processing via $\gH(l, \phi)$ by utilizing the corresponding data embedding matrix $\mathbf{E}_{\texttt{data}}$ and generating a single \textit{chain token}. Formally, given a token embedding $\ve \in \sR^{d}$, the output of the MLP $h \in \gH(l, \phi)$ is formulated as:
\begin{align}
    h(\ve) = \mW_L\left( \phi\left(\mW_{L-1}(\cdots \phi\left(\mW_{1}\left( \ve \right) \right))\right) \right).
\end{align}
Here $\mW_l \in \sR^{d \times d} , \forall l \in [L]$ denote the linear layers whose width is kept constant ($d$) across layers, and whose entries are sampled from $\gN(0,1)$.
The usefulness of \coticl~lies in its flexibility to modify Algorithm~\ref{alg:tok_proc_H}. For instance:
\begin{enumerate}
    \item Future efforts can explore non-random $\mE_{\texttt{data}}$ and also modify Step 3 to employ more complex function classes beyond just MLPs.
    \item One can also explore feature aggregation techniques (Step 5-6) when scaling to ($|\gV| > 1024$).
\end{enumerate}
\begin{algorithm}
\caption{Generate a single \textit{chain token} $y_c$}
\begin{algorithmic}[1]
\label{alg:tok_proc_H}
\REQUIRE $M$ parent token embeddings $\{ \mathbf{e}^i_{\texttt{data}}\}_{i=1}^{M}$ from the data embedding matrix $\mE_{\texttt{data}} \in \sR^{|\gV| \times d}$, choices of depth $l$ and activation functions $\phi$, 

\STATE Initialize MLP $h_c \in \mathcal{H}(l, \phi)$
\FOR{$i = 1$ to $M$}
    \STATE $\mathbf{h}^i \leftarrow h_c(\mathbf{e}^i_{\texttt{data}})$ \quad 
    \hspace{10pt}\COMMENT{Process each parent embedding}
\ENDFOR
\STATE $\mathbf{h}_{\text{mean}} \leftarrow \frac{1}{M} \sum_{i=1}^{M} \mathbf{h}^i$ \quad 
\hspace{15pt}\COMMENT{Mean of final layer features}
\STATE $\mathbf{h}_{\text{act}} \leftarrow \phi(\mathbf{h}_{\text{mean}})$ \quad 
\hspace{30pt}\COMMENT{Apply activation again}
\STATE $y_c \leftarrow \texttt{argmax}(\mathbf{E}_{\texttt{data}} \mathbf{h}_{\text{act}})$ \quad \
\COMMENT{Compute chain token}
\RETURN $y_c$
\end{algorithmic}
\end{algorithm}

\subsection{Synthetic Datasets and Token Distributions}
\label{subsec:token_dist}

To create synthetic training and evaluation sequence datasets via the \coticl~framework, we consider a vocabulary $\gV$ of arbitrary size and the data embedding matrix $\mE_{\texttt{data}} \in \sR^{|\gV| \times d}$. To create a single sequence, we randomly sample a DAG from $\gG(M,N,C)$ and sample $C$ MLPs from $\gH(l, \phi)$. The $N$ input tokens per example are sampled uniformly from $\gV$ and are then used to generate the $C$ chain tokens using \eqref{eq:comp_f_dag}. Note that $g_c(.)$ corresponds to the $M$ edges of the DAG that map the parent tokens to the chain token (i.e, the filtering function) and $h_c(.)$ corresponds to the token processing function via a MLP (as per Algorithm~\ref{alg:tok_proc_H}). Creating $K$ such examples/CoT examples gives us a sequence and creating $T$ such sequences gives us the desired synthetic dataset.

\begin{figure}[ht!]
    \centering
    \begin{subfigure}[b]{0.23\textwidth}
        \centering
        \includegraphics[width=\textwidth]{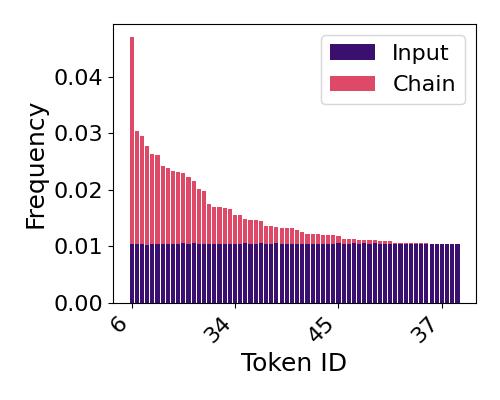}
        \caption{$d=10$}
    \end{subfigure}
    \hfill
    \begin{subfigure}[b]{0.23\textwidth}
        \centering
        \includegraphics[width=\textwidth]{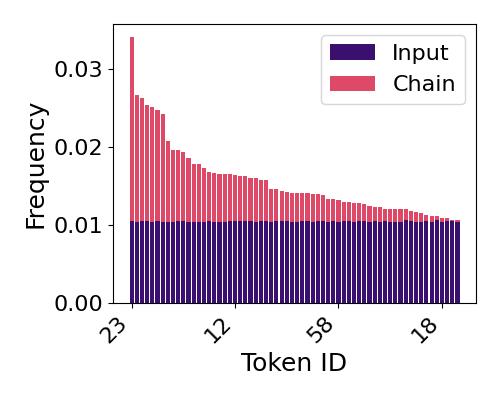}
        \caption{$d=20$}
    \end{subfigure}
    \hfill
    \begin{subfigure}[b]{0.23\textwidth}
        \centering
        \includegraphics[width=\textwidth]{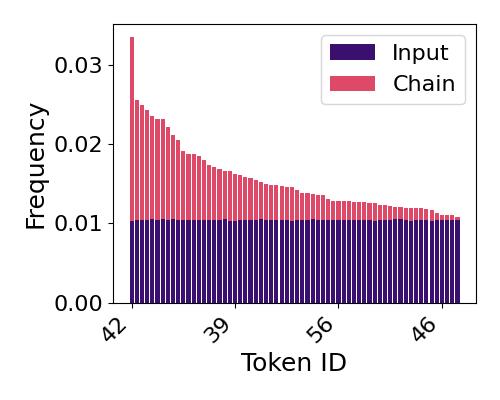}
        \caption{$d=30$}
    \end{subfigure}
    \hfill
    \begin{subfigure}[b]{0.23\textwidth}
        \centering
        \includegraphics[width=\textwidth]{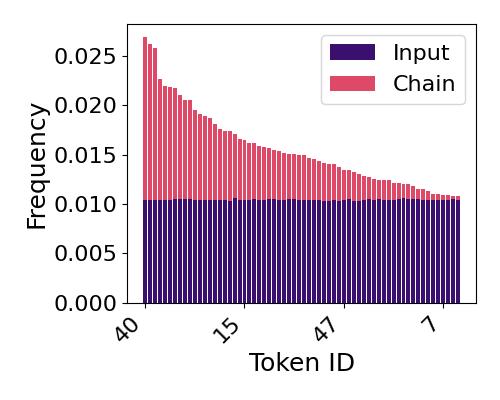}
        \caption{$d=40$}
    \end{subfigure}
    \caption{Token distribution from $10,000$ CoT sequences with $|\gV|=64, K=40, \gH(1, \texttt{LeakyReLU})$ and $\gG(M=4, N=4, C=2)$.}
\label{fig:token_dist_randn_V_64_act_leaky_relu_cot_True}
\end{figure}

\begin{figure}[ht!]
    \centering
    \begin{subfigure}[b]{0.23\textwidth}
        \centering
        \includegraphics[width=\textwidth]{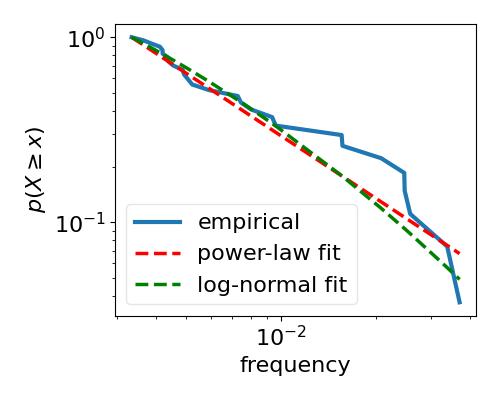}
        \caption{ReLU}
    \end{subfigure}
    \hfill
    \begin{subfigure}[b]{0.23\textwidth}
        \centering
        \includegraphics[width=\textwidth]{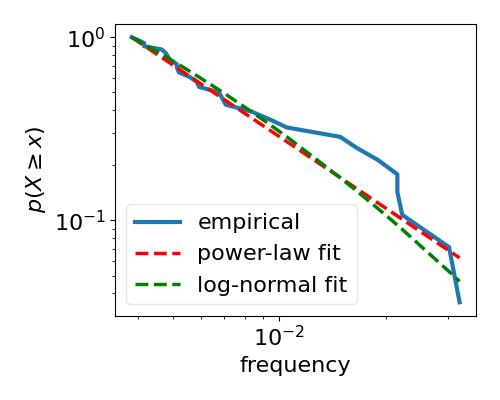}
        \caption{SiLU}
        
    \end{subfigure}
    \hfill
    \begin{subfigure}[b]{0.23\textwidth}
        \centering
        \includegraphics[width=\textwidth]{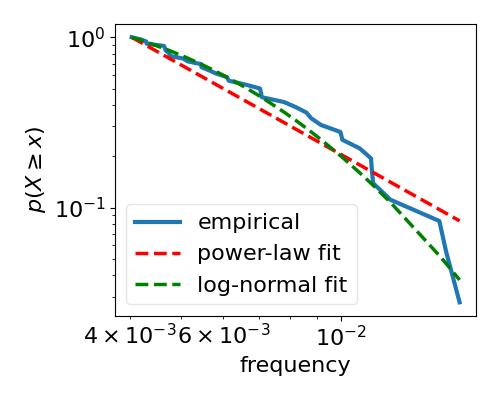}
        \caption{LeakyReLU}
        
    \end{subfigure}
    \hfill
    \begin{subfigure}[b]{0.23\textwidth}
        \centering
        \includegraphics[width=\textwidth]{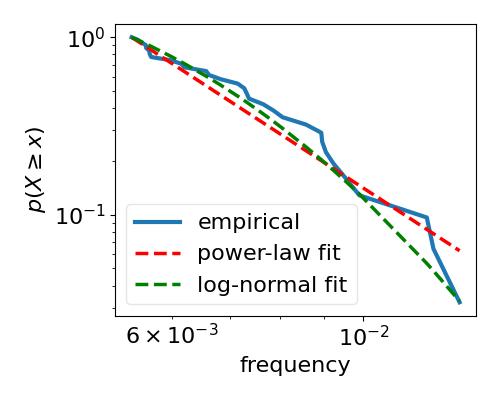}
        \caption{Identity}
        
    \end{subfigure}
    \caption{Complementary cumulative distribution plots of the \textit{chain token} distribution, it's corresponding power-law and log-normal fits. We sample $10,000$ CoT sequences with $|\gV|=64, d=40, \gH(1, \phi), \gG(M=4, N=4, C=2), K=40$.}
    \label{fig:token_dist_randn_V_64_act_identity_cot_True_pl}
\end{figure}

\subsection{Token Distribution Fits}
To understand the distribution of tokens in our synthetic datasets, we sample $10,000$ CoT sequences with $|\gV|=64, d=\{10, 20, 30, 40\}, \gH(1, \texttt{LeakyReLU}), N=4, M=4, C=2, K=40$ and plot the distribution of input tokens and chain tokens in Figure~\ref{fig:token_dist_randn_V_64_act_leaky_relu_cot_True}. Observe that the distribution of chain tokens in the sequences exhibit a decay that depends on $d$. By utilizing the \texttt{powerlaw} python package \cite{alstott2014powerlaw} to fit power-law and log-normal distributions to the \emph{chain tokens} frequencies, we quantify that the distribution is more likely to be log-normal. See Figure~\ref{fig:token_dist_randn_V_64_act_identity_cot_True_pl} for the $d=40$ case with varying $\phi$.

To quantify the impact of varying factors (such as $\gV, d, l, \phi$), we measure the \texttt{TokenCoverage} over all chain tokens in the dataset as follows:
\begin{definition}
    The $\texttt{TokenCoverage} \in [0, 1]$ of a dataset represents the ratio of number of unique chain tokens to the number of unique tokens present in the entire dataset.
\end{definition}

\paragraph{Remark.} We emphasize that \texttt{TokenCoverage} acts only as a first-order explanation of uniqueness in the entire collection of chain tokens and does not account for the unique CoT examples in the dataset.

\subsection{\texttt{TokenCoverage} by MLP depth $l$}

\begin{figure}[ht!]
    \centering
    \begin{subfigure}[b]{0.23\textwidth}
        \centering
        \includegraphics[width=\textwidth]{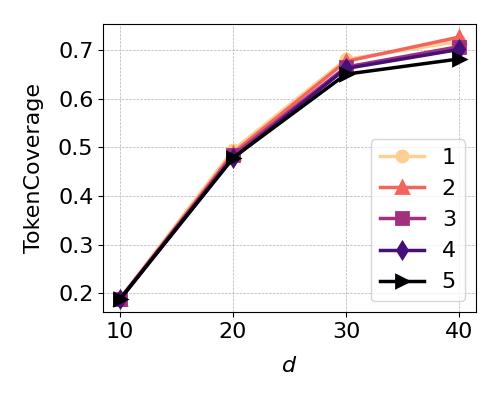}
        \caption{ReLU}
        
    \end{subfigure}
    \begin{subfigure}[b]{0.23\textwidth}
        \centering
        \includegraphics[width=\textwidth]{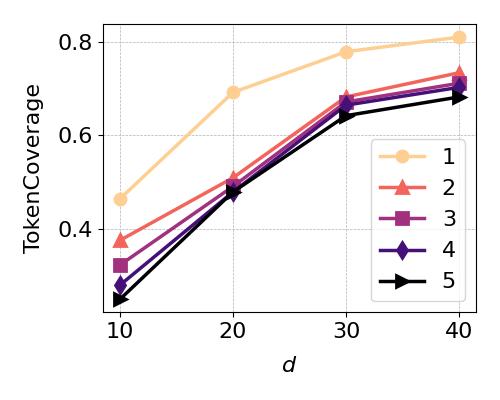}
        \caption{SiLU}
        
    \end{subfigure}
    \begin{subfigure}[b]{0.23\textwidth}
        \centering
        \includegraphics[width=\textwidth]{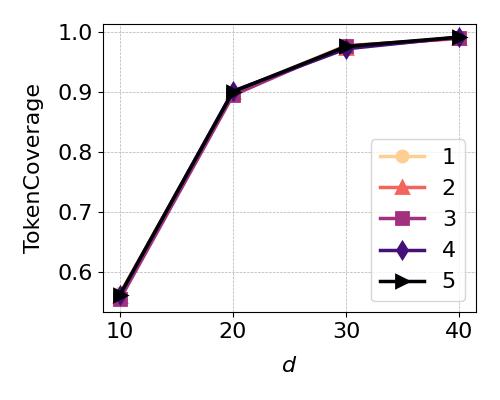}
        \caption{LeakyReLU}
        
    \end{subfigure}
    \begin{subfigure}[b]{0.23\textwidth}
        \centering
        \includegraphics[width=\textwidth]{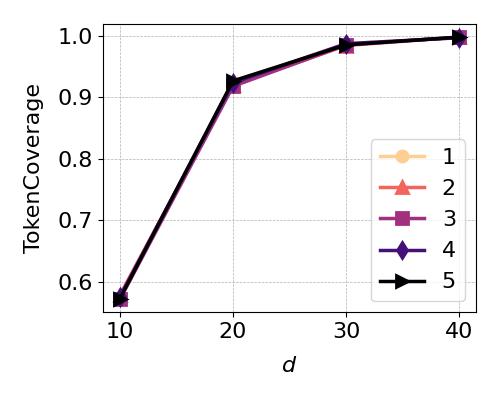}
        \caption{Identity}
        
    \end{subfigure}
    \caption{Measuring \texttt{TokenCoverage} of $T=10,000$ CoT sequences with $\gG(M=4,N=4,C=2),K=40, |\gV|=1024, d=\{10,20,30,40\}$ and $\gH(l, \phi)$. We choose $l=\{1,2,3,4,5\}$ and $\phi= \{\texttt{ReLU, SiLU, LeakyReLU, Identity}\}$.}
    \label{fig:token_cov_V_1024_vary_H_layers}
\end{figure}

To illustrate the role of depth $l$ of the MLPs in generating the chain tokens, we sample $T=10,000$ CoT sequences with $|\gV|=1024, d=\{10,20,30,40\}$ and $\gH(l, \phi)$. We choose $l=\{1,2,3,4,5\}$ and $\phi= \{\texttt{ReLU, SiLU, LeakyReLU, Identity}\}$. Each CoT sequence uses $N=4,M=4,C=2,K=40$. Figure~\ref{fig:token_cov_V_1024_vary_H_layers} illustrates that \texttt{ReLU, LeakyReLU, Identity} functions have approximately the same token coverage for varying $l$. However, \texttt{SiLU} tends to exhibit lower \texttt{TokenCoverage} for larger $l$ values.

\subsection{\texttt{TokenCoverage} by MLP Activation $\phi$.}
\begin{figure}[t!]
    \centering
    \begin{subfigure}[b]{0.32\textwidth}
        \centering
        \includegraphics[width=\textwidth]{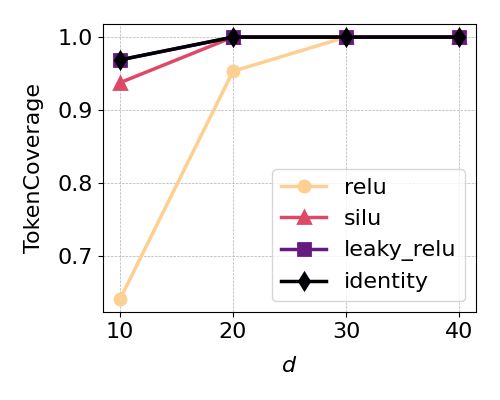}
        \caption{$|\gV|=64$}
        \label{fig:token_cov_vary_act_V_64}
    \end{subfigure}
    \begin{subfigure}[b]{0.32\textwidth}
        \centering
        \includegraphics[width=\textwidth]{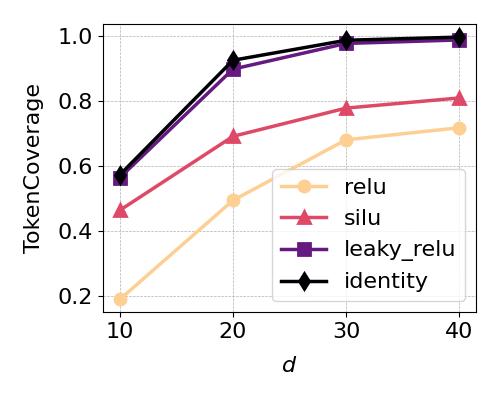}
        \caption{$|\gV|=1024$}
        
    \end{subfigure}
    \caption{\texttt{TokenCoverage} of $10,000$ CoT sequences with $K=40, \gH(1, \phi)$ and $\gG(M=4, N=4, C=2)$.}
    \label{fig:token_cov_vary_act}
    \vspace{-3mm}
\end{figure}

We vary $\phi=\{\texttt{ReLU, SiLU, LeakyReLU, Identity}\}$ and plot the \texttt{TokenCoverage} for a similar dataset with $T=10,000$ CoT sequences in Figure~\ref{fig:token_cov_vary_act}.  A key takeaway is that \texttt{ReLU} and \texttt{SiLU} lead to highly skewed chain token distributions with relatively smaller \texttt{TokenCoverage} for any given dimension $d$. Nonetheless, all $\phi$ exhibit an increasing trend for larger values of $d$.

\subsection{\texttt{TokenCoverage} in the CoT-Collection NLP Dataset}

\begin{figure}[ht!]
    \centering
        \includegraphics[width=0.42\textwidth]{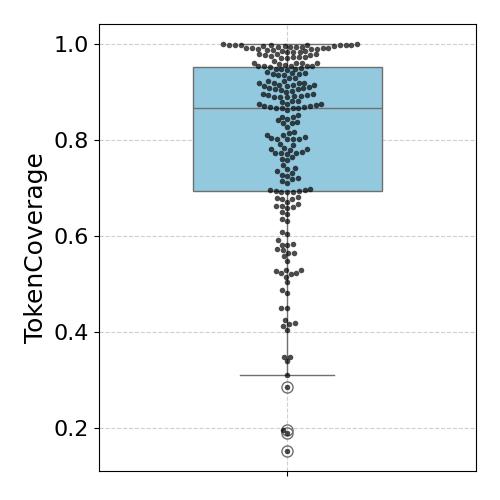}
    \caption{Measuring \texttt{TokenCoverage} of all the $216$ task datasets from the CoT-Collection corpus using the Llama-3.1 Tokenizer with a vocab size of $128256$. }
    \label{fig:token_cov_cot_collection}
\end{figure}

To further show that our setup is flexible enough to resemble realistic NLP scenarios in token distribution, we analyze CoT sequences from the CoT-Collection dataset \cite{kim2023the} which collates $\approx 1.8$ million NLP prompts from a diverse pool of $216$ tasks. We analyze the \texttt{TokenCoverage} by tokenizing the reasoning text of the NLP prompts pertaining to each task with the Llama-3.1 tokenizer ($|\gV|=128256$).  From Figure~\ref{fig:token_cov_cot_collection}, notice that the tasks span a wide range of the \texttt{TokenCoverage} values with the lowest being $<0.2$ and the highest being $1$.  
From the above analysis of simulating chains via different configurations of $\gH, d$, we can notice that our setup is flexible enough to replicate the complexity of real-world datasets (in terms of input token lengths and chain lengths) and offer flexibility of simulating even complex datasets which might not be easy to curate.

\section{Hardware and Hyper-Parameters for Training and Evaluation}
\label{app:hardware_hyperparams}
We use the \texttt{DistributedDataParallel} APIs in PyTorch \cite{paszke2019pytorch} to run each training job on 4 H100 NVIDIA GPUs. Furthermore, since our modeling code leverages HuggingFace APIs, we also employ Liger-Kernels \cite{hsu2024liger} for faster training. We created 3 different models based on the Llama-3 architecture whose details are presented in Table~\ref{tab:model_card}. The smallest model $\tTF$-4 has $\approx 240$M parameters without the embedding layer and the largest model $\tTF$-12 has $\approx 730$M parameters. For all the training runs, we use a batch size of $64$ per rank and the \texttt{AdamW} optimizer with a learning rate $5\times 10^{-5}$. We employ the \texttt{GenerationConfig} API in the \texttt{transformers} library to greedily generate the model predictions without teacher forcing. This API is used for evaluations on checkpoints during the training runs. For larger scale and on-demand evaluations, we provide code examples to leverage vLLM \cite{kwon2023efficient} and SGLang \cite{zheng2024sglang} based inference.

\begin{table*}[ht!]
\centering
\begin{tabular}{|c|c|c|c|c|c|}
\hline
\textbf{Model} & \textbf{Layers} & \textbf{Attention Heads} & \textbf{Hidden Size} & \textbf{Intermediate Size} & \textbf{Params} \\ \hline
\texttt{TF-4}      &  $4$   &  $32$  &  $2048$   & $8192$ & $243,288,064$\\ \hline
\texttt{TF-8}      &  $8$   &  $32$  &  $2048$   & $8192$ & $486,574,080$ \\ \hline
\texttt{TF-12}      &  $12$   &  $32$  &  $2048$   & $8192$ & $729,860,096$ \\ \hline
\end{tabular}
\caption{Model Card: A naming convention for the \texttt{TF} models along with their number of layers, attention heads, hidden size, and intermediate size. The parameter count excludes the Embedding layer weights.}
\label{tab:model_card}
\vspace{-2mm}
\end{table*}

\section{Additional Experiments}
\label{app:sec:add_exp}

\subsection{Varying Activation Functions ($\phi$)}

\paragraph{Setup.} Considering a vocabulary $|\gV|=64, d=10$, we employ $\gH(1, \phi)$, $N=4, M=4, C=2,K=40$ with varying $\phi=\{\texttt{ReLU, SiLU, LeakyReLU} \}$. All models are trained on a dataset with $T=32 \times 10^5$ sequences and evaluated on $\widetilde{T} = 10,000$ sequences.

\begin{figure*}[ht!]
    \centering
    \begin{subfigure}[b]{0.32\textwidth}
        \centering
        \includegraphics[width=\textwidth]{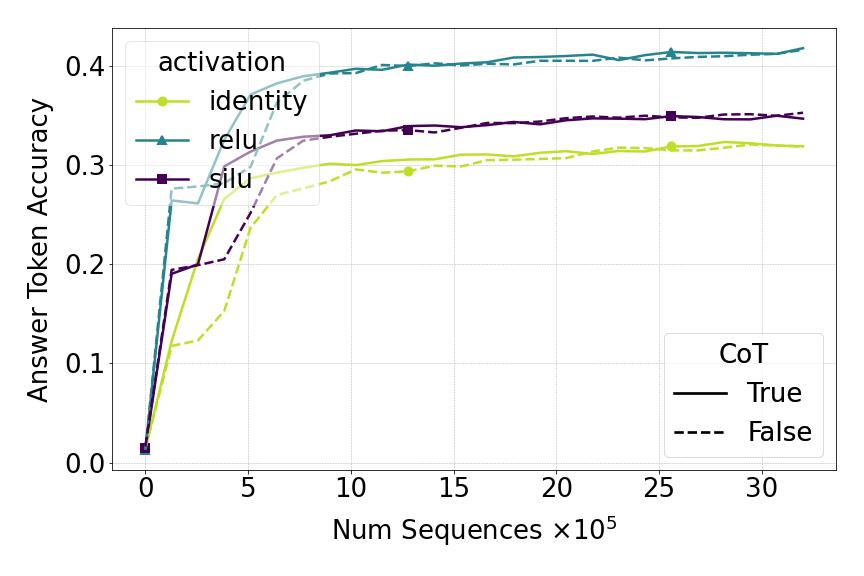}
        \caption{\texttt{TF-4}}
        
    \end{subfigure}
    \hfill
    \begin{subfigure}[b]{0.32\textwidth}
        \centering
        \includegraphics[width=\textwidth]{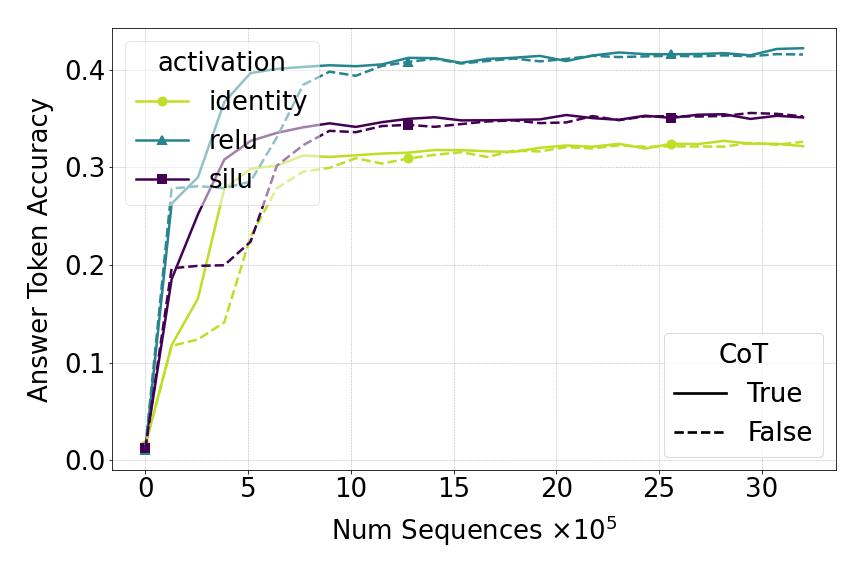}
        \caption{\texttt{TF-8}}
        
    \end{subfigure}
    \hfill
    \begin{subfigure}[b]{0.32\textwidth}
        \centering
        \includegraphics[width=\textwidth]{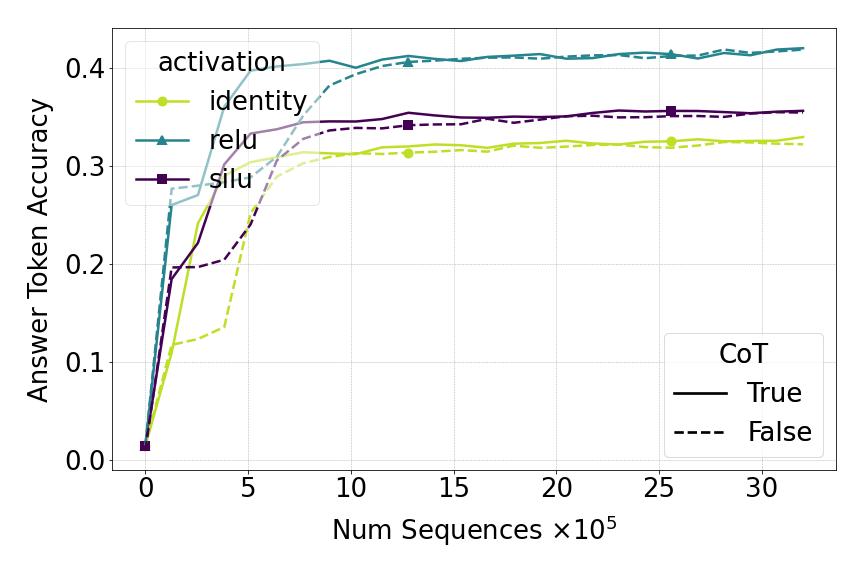}
        \caption{\texttt{TF-12}}
        
    \end{subfigure}
    \caption{$\tacc$ by varying $\phi$ in $\gH(1, \phi)$ with $\gG(M=4, N=4,C=2),d=10,|\gV|=64,K=40$.}
    \label{fig:vary_act_M_4_N_4_C_2_V_64_d_10_n_ex_40}
    \vspace{-2mm}
\end{figure*}

\paragraph{Lower \texttt{TokenCoverage} leads to higher $\tacc$.} We previously observed from Figure~\ref{fig:token_cov_vary_act} that the activation function $(\phi)$ corresponding to $\gH$ plays a key-role in determining the \texttt{TokenCoverage} of chain tokens in the sequences. For instance, when $|\gV|=64$, the \texttt{TokenCoverage} with \texttt{Identity} and \texttt{SiLU} is relatively larger than \texttt{ReLU}. This implies that the number of unique tokens that a model would have to correctly predict is relatively lower in the \texttt{ReLU} case. We can observe from Figure~\ref{fig:vary_act_M_4_N_4_C_2_V_64_d_10_n_ex_40} that such smaller coverage can indeed result in higher evaluation $\tacc$ across all model sizes.

\vspace{-2mm}
\subsection{Vary Data Embedding Dimension ($d$)}

\begin{figure*}[ht!]
    \centering
    \begin{subfigure}[b]{0.32\textwidth}
        \centering
        \includegraphics[width=\textwidth]{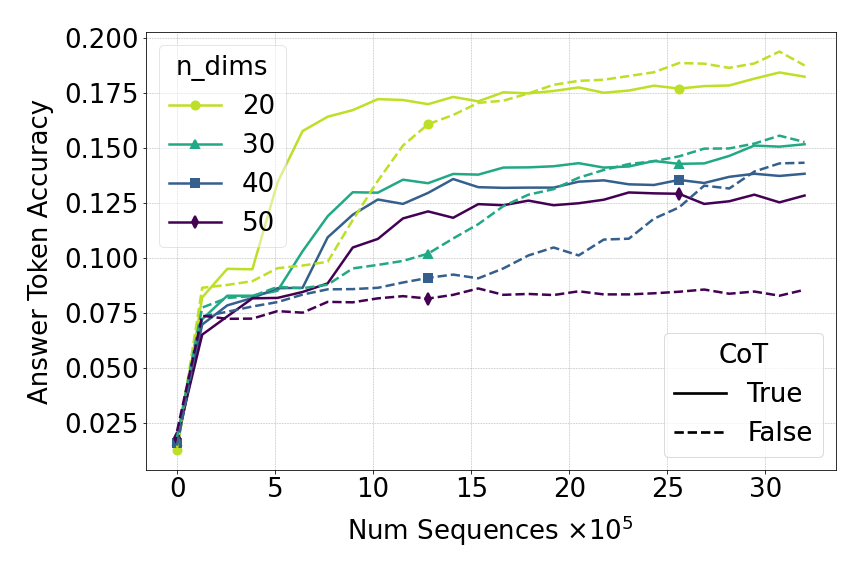}
        \caption{\texttt{TF-4}}
        
    \end{subfigure}
    \hfill
    \begin{subfigure}[b]{0.32\textwidth}
        \centering
        \includegraphics[width=\textwidth]{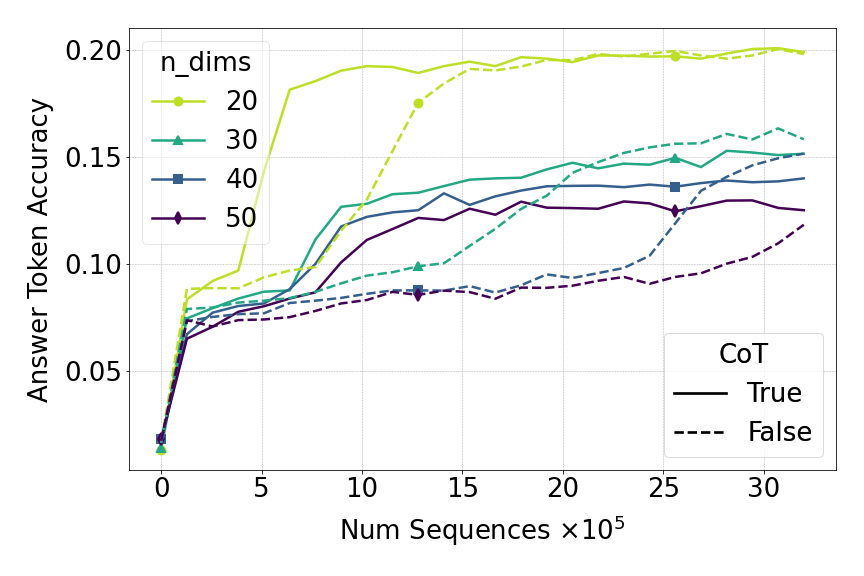}
        \caption{\texttt{TF-8}}
        
    \end{subfigure}
    \hfill
    \begin{subfigure}[b]{0.32\textwidth}
        \centering
        \includegraphics[width=\textwidth]{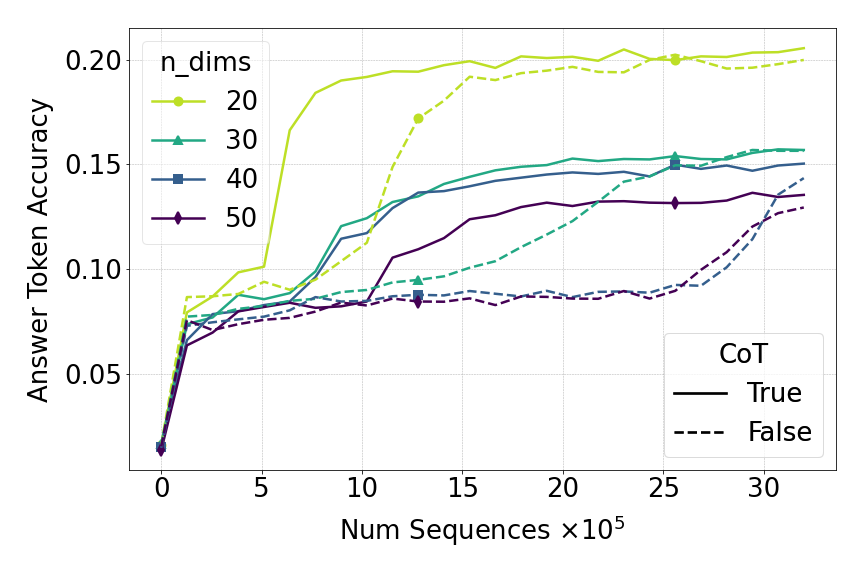}
        \caption{\texttt{TF-12}}
        
    \end{subfigure}
    \caption{$\tacc$ by varying $d$ with $\gG(M=4,N=4,C=2),\gH(1, \texttt{LeakyRelu}),|\gV|=64,K=40$. }
    \label{fig:vary_n_dims_N_4_M_4_C_2_V_64}
    \vspace{-2mm}
\end{figure*}

Based on the token distribution analysis, we have noticed that the \texttt{TokenCoverage} increases monotonically with $d$ for various choices of $\gH(l,\phi)$ in Figure~\ref{fig:token_cov_V_1024_vary_H_layers}. As a result, Figure~\ref{fig:vary_n_dims_N_4_M_4_C_2_V_64} shows that larger coverage makes it difficult for the $\tTF$ models to attain high evaluation $\tacc$ as they have to now correctly predict a larger fraction of tokens in $\gV$.

\vspace{-2mm}
\subsection{Vary Number of (CoT-) examples $K$}
\label{app:subsec:vary_K}
\begin{figure*}[ht!]
    \centering
    \begin{subfigure}[b]{0.32\textwidth}
        \centering
        \includegraphics[width=\textwidth]{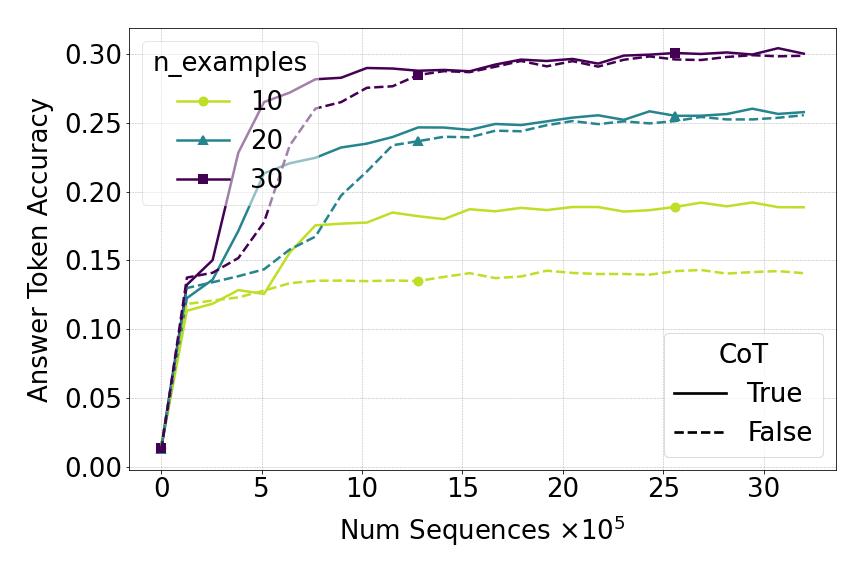}
        \caption{\texttt{TF-4}}
        
    \end{subfigure}
    \hfill
    \begin{subfigure}[b]{0.32\textwidth}
        \centering
        \includegraphics[width=\textwidth]{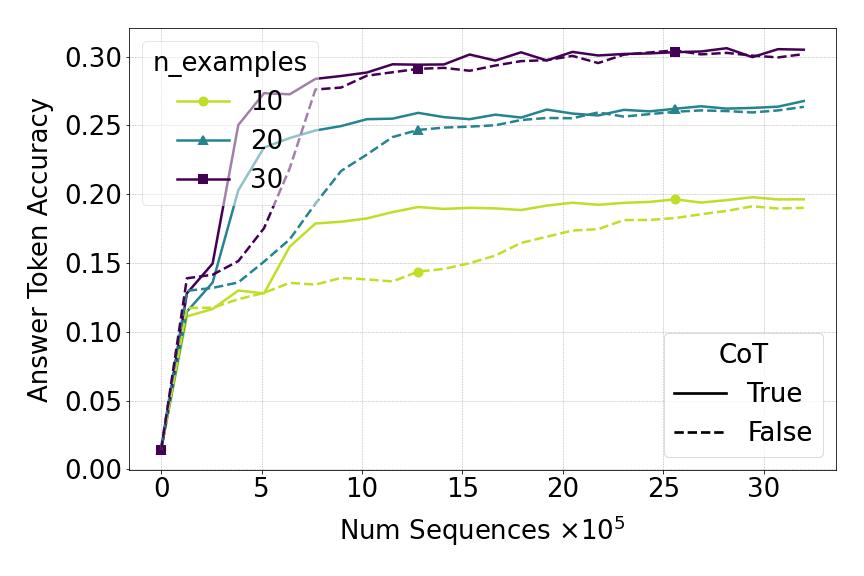}
        \caption{\texttt{TF-8}}
        
    \end{subfigure}
    \hfill
    \begin{subfigure}[b]{0.32\textwidth}
        \centering
        \includegraphics[width=\textwidth]{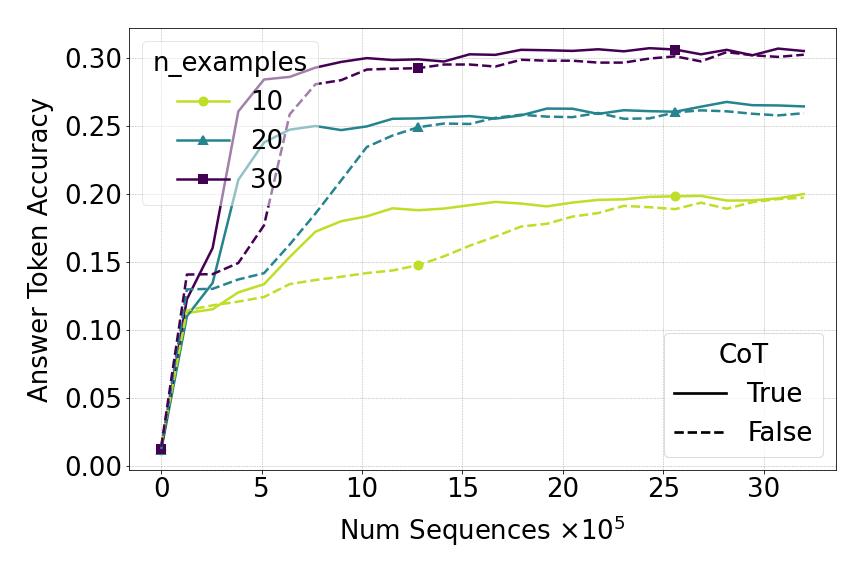}
        \caption{\texttt{TF-12}}
        
    \end{subfigure}
    \caption{$\tacc$ by varying $K$ with $\gG(M=4, N=4,C=2),\gH(1, \texttt{LeakyRelu}),d=10,|\gV|=64$.}
    \label{fig:vary_n_examples_M_4_N_4_C_2_V_64_d_10}
\end{figure*}

\begin{figure*}[ht!]
    \centering
    \begin{subfigure}[b]{0.32\textwidth}
        \centering
        \includegraphics[width=\textwidth]{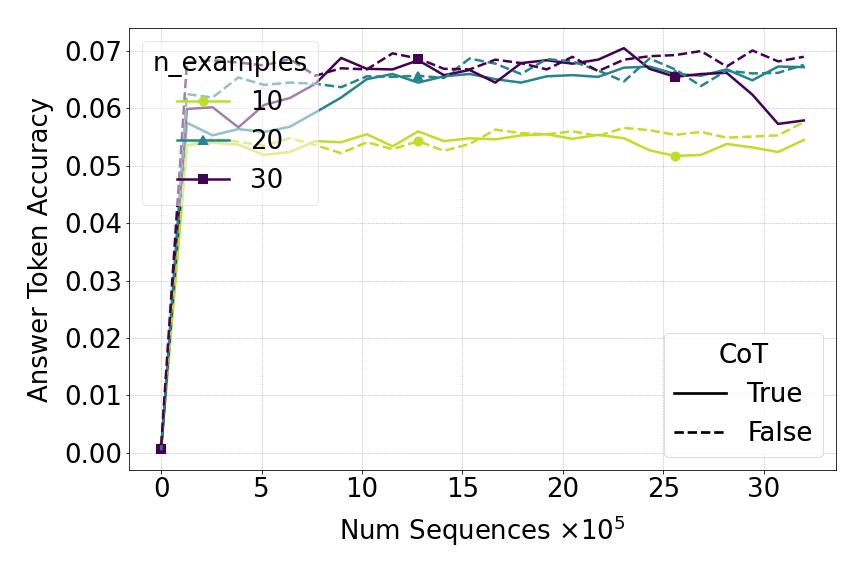}
        \caption{\texttt{TF-4}}
        
    \end{subfigure}
    \hfill
    \begin{subfigure}[b]{0.32\textwidth}
        \centering
        \includegraphics[width=\textwidth]{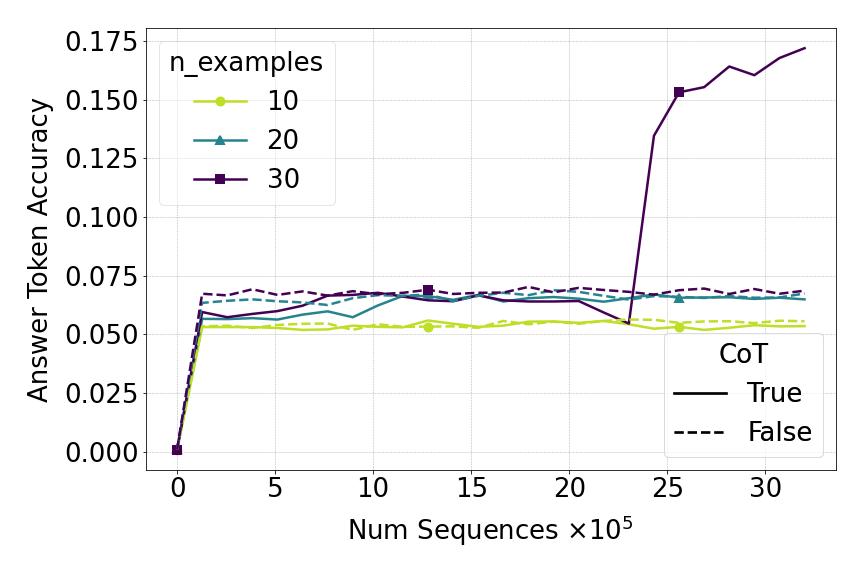}
        \caption{\texttt{TF-8}}
        
    \end{subfigure}
    \hfill
    \begin{subfigure}[b]{0.32\textwidth}
        \centering
        \includegraphics[width=\textwidth]{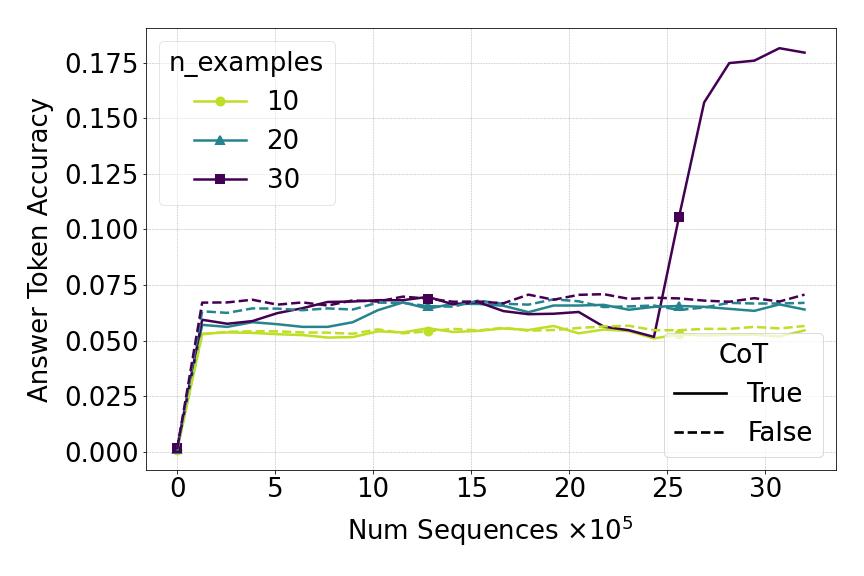}
        \caption{\texttt{TF-12}}
        
    \end{subfigure}
    \caption{$\tacc$ by varying $K$ with $\gG(M=4, N=4,C=2),\gH(1, \texttt{LeakyRelu}),d=10,|\gV|=1024,$.}
    \label{fig:vary_n_examples_M_4_N_4_C_2_V_1024_d_10}
\end{figure*}

In Section~\ref{subsec:vary_V}, we observed that more examples in-context can help smaller models to perform on-par with bigger models. To this end, we plot in Figure~\ref{fig:vary_n_examples_M_4_N_4_C_2_V_64_d_10} the $\tacc$ by varying $K=\{10, 20, 30\}$. For simplicity we choose a smaller vocab size of $|\gV|=64$ and show that smaller $K$ can hurt performance even when CoT is employed across all model sizes. Additionally, larger models tend to outperform the smaller one $\tTF$-4 in the extreme case of $K=10$ without any CoT. Next, by increasing the vocabulary size to $|\gV|=1024$, we observe from Figure~\ref{fig:vary_n_examples_M_4_N_4_C_2_V_1024_d_10} that $\tTF$-8, $\tTF$-12 leverage their depth and outperform $\tTF$-4 by utilizing CoT when $K=30$. For $K=10, 20$, the problem turns out to be harder even for these bigger models.

\subsection{Longer Training}
\label{app:subsec:longer_training}

\begin{figure*}[ht!]
    \centering
    \begin{subfigure}[b]{0.32\textwidth}
        \centering
        \includegraphics[width=\textwidth]{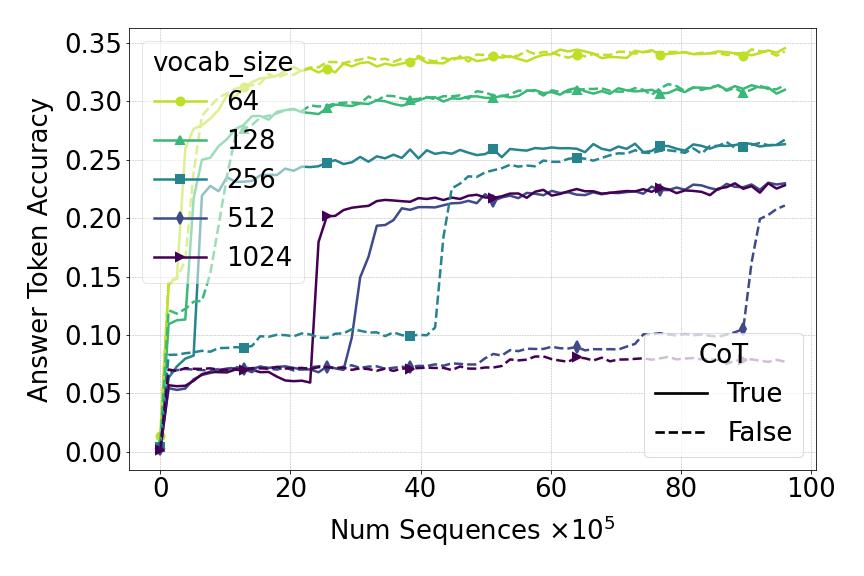}
        \caption{\texttt{TF-4}}
        \label{fig:vary_V_N_4_M_4_C_2_n_steps_150k_L_4}
    \end{subfigure}
    \hfill
    \begin{subfigure}[b]{0.32\textwidth}
        \centering
        \includegraphics[width=\textwidth]{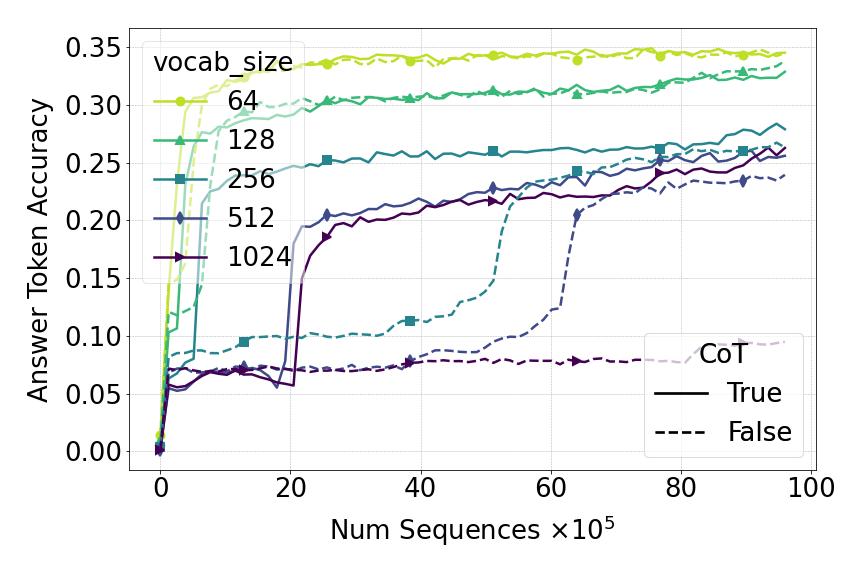}
        \caption{\texttt{TF-8}}
        \label{fig:vary_V_N_4_M_4_C_2_n_steps_150k_L_8}
    \end{subfigure}
    \hfill
    \begin{subfigure}[b]{0.32\textwidth}
        \centering
        \includegraphics[width=\textwidth]{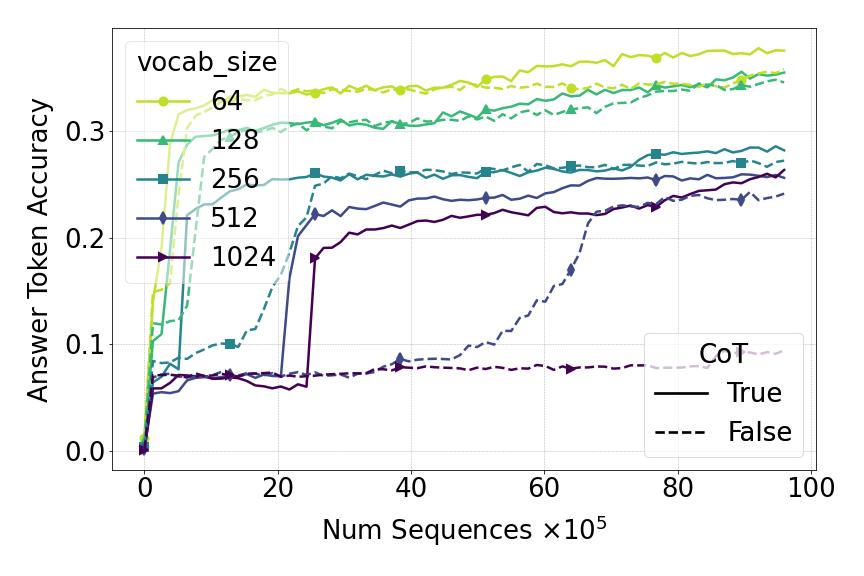}
        \caption{\texttt{TF-12}}
        \label{fig:vary_V_N_4_M_4_C_2_n_steps_150k_L_12}
    \end{subfigure}
    \caption{$\tacc$ by varying $\gV$ with $\gG(M=4,N=4,C=2),\gH(1, \texttt{LeakyReLU}), d=10,K=40$ and training for $3\times$ steps (i.e $96 \times 10^5$ (CoT-) sequences). }
    \label{fig:vary_V_N_4_M_4_C_2_n_steps_150k}
\end{figure*}

We follow the same setup of Section~\ref{subsec:vary_V} and explore the impact of longer training on the $\tacc$ of the $\tTF$ models. In particular, we train on $3\times$ more steps, which results in $96 \times 10^5$ (CoT-) sequences per dataset. Observe from Figure~\ref{fig:vary_V_N_4_M_4_C_2_n_steps_150k_L_4} that by training beyond $32 \times 10^5$ sequences, even the $\tTF$-4 model exhibits a transition in non-CoT evaluation $\tacc$ for $|\gV|=\{256, 512\}$. Similar behavior can be observed with $\tTF$-8 (Figure~\ref{fig:vary_V_N_4_M_4_C_2_n_steps_150k_L_8}) and $\tTF$-12 (Figure~\ref{fig:vary_V_N_4_M_4_C_2_n_steps_150k_L_12}) where the transitions tend to occur with relatively less number of training sequences. Additionally, the bigger models tend to reach a final $\tacc$ with CoT which is higher than that of the smaller $\tTF$-4 model.

\subsection{Ablations with fixed DAGs and Token Processors}
\label{app:subsec:ablation_fixed_G_H}

\begin{figure*}[ht!]
    \centering
    \begin{subfigure}[b]{0.32\textwidth}
        \centering
        \includegraphics[width=\textwidth]{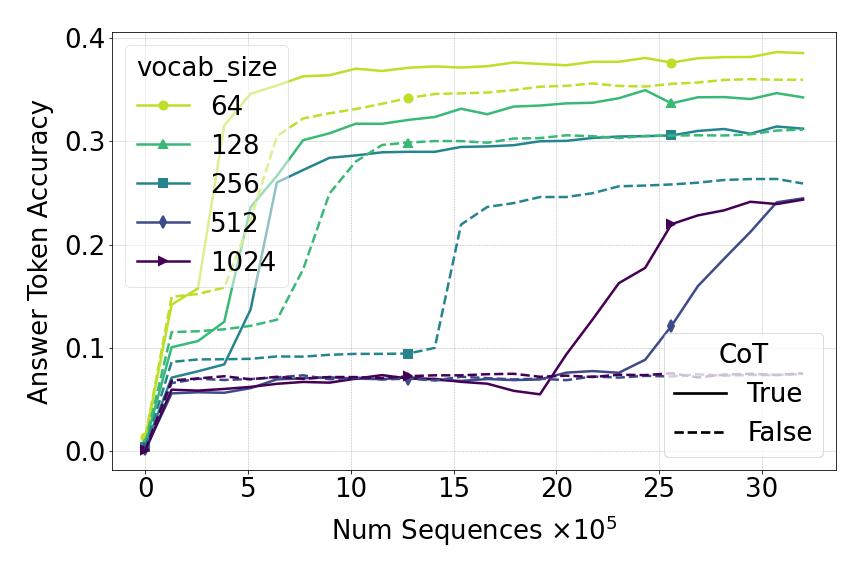}
        \caption{\texttt{TF-4}}
        \label{fig:ablation_fixed_random_dag_vary_V_N_4_M_4_C_2_n_ex_40_L_4}
    \end{subfigure}
    \hfill
    \begin{subfigure}[b]{0.32\textwidth}
        \centering
        \includegraphics[width=\textwidth]{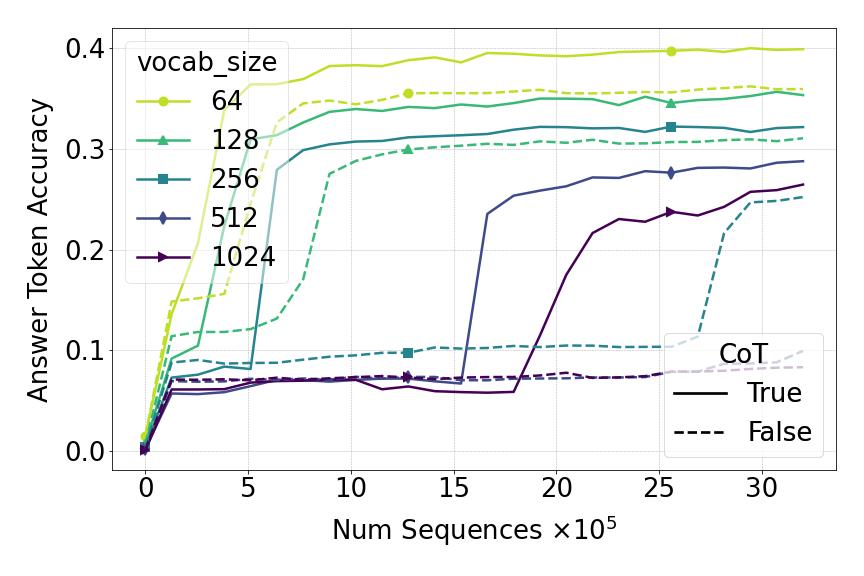}
        \caption{\texttt{TF-8}}
        \label{fig:ablation_fixed_random_dag_vary_V_N_4_M_4_C_2_n_ex_40_L_8}
    \end{subfigure}
    \hfill
    \begin{subfigure}[b]{0.32\textwidth}
        \centering
        \includegraphics[width=\textwidth]{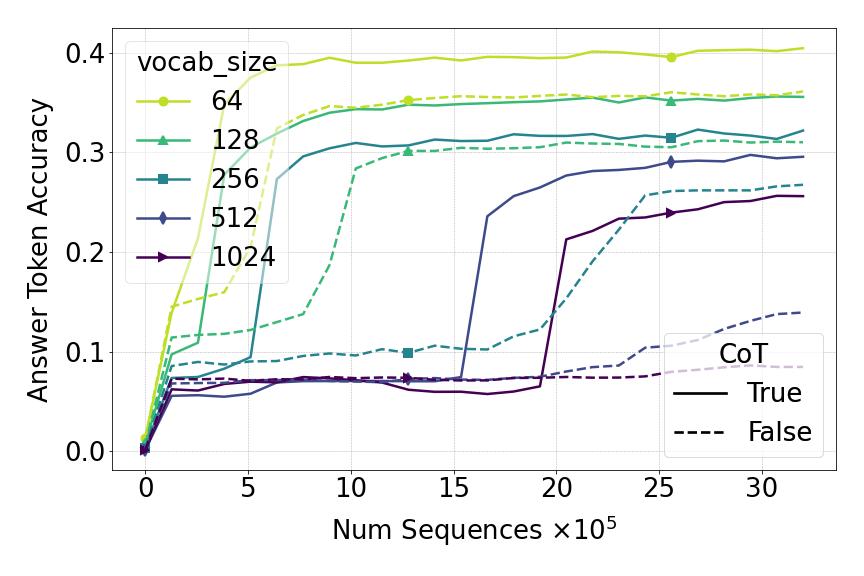}
        \caption{\texttt{TF-12}}
        \label{fig:ablation_fixed_random_dag_vary_V_N_4_M_4_C_2_n_ex_40_L_12}
    \end{subfigure}
    \caption{$\tacc$ by varying $\gV$ with $\gH(1, \texttt{LeakyRelu}), d=10, K=40$ and a \textbf{fixed DAG} sampled from $\gG(M=4,N=4,C=2)$.}
    \label{fig:ablation_fixed_random_dag_vary_V_N_4_M_4_C_2_n_ex_40}
    \vspace{-2mm}
\end{figure*}

\begin{figure*}[ht!]
    \centering
    \begin{subfigure}[b]{0.32\textwidth}
        \centering
        \includegraphics[width=\textwidth]{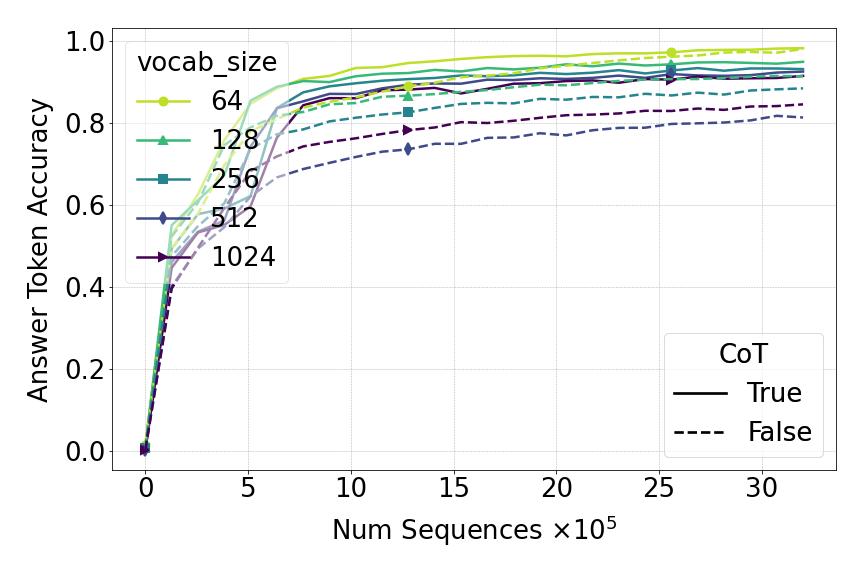}
        \caption{\texttt{TF-4}}
        \label{fig:ablation_fixed_H_vary_V_N_4_M_4_C_2_n_ex_40_L_4}
    \end{subfigure}
    \hfill
    \begin{subfigure}[b]{0.32\textwidth}
        \centering
        \includegraphics[width=\textwidth]{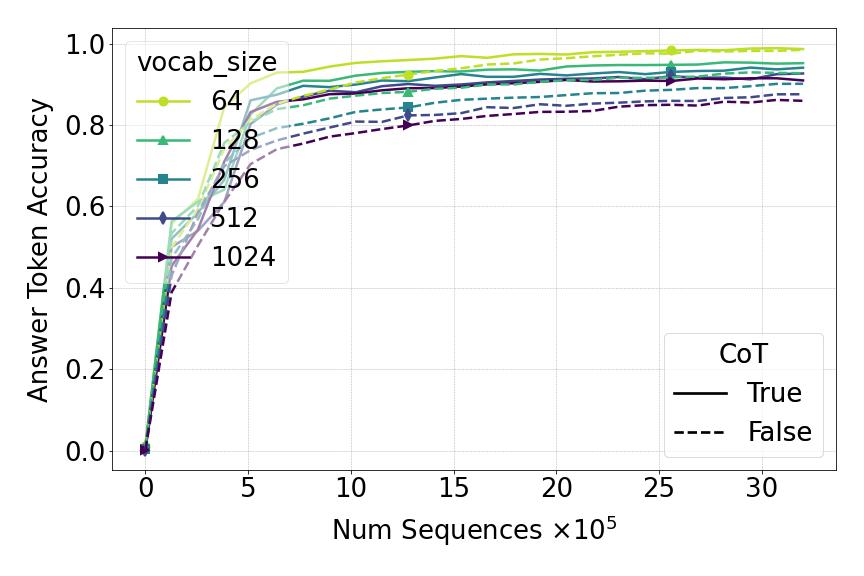}
        \caption{\texttt{TF-8}}
        \label{fig:ablation_fixed_H_vary_V_N_4_M_4_C_2_n_ex_40_L_8}
    \end{subfigure}
    \hfill
    \begin{subfigure}[b]{0.32\textwidth}
        \centering
        \includegraphics[width=\textwidth]{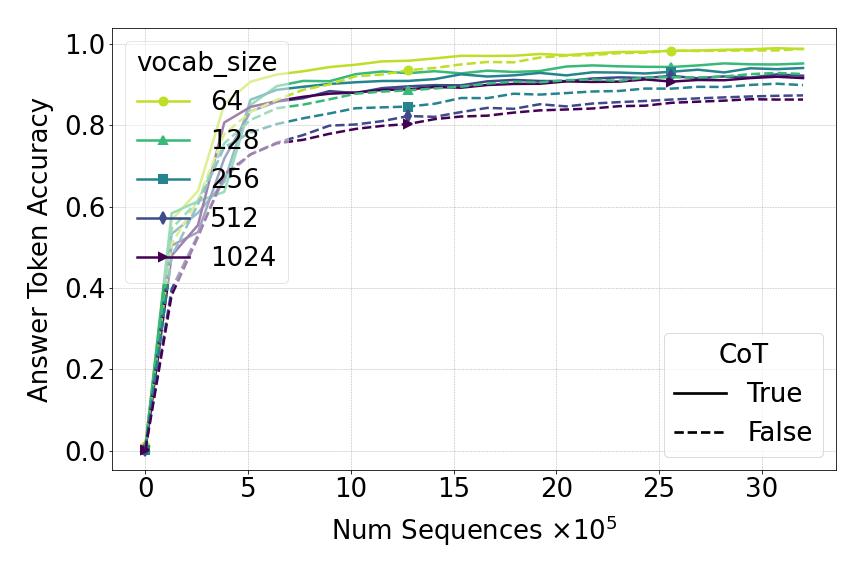}
        \caption{\texttt{TF-12}}
        \label{fig:ablation_fixed_H_vary_V_N_4_M_4_C_2_n_ex_40_L_12}
    \end{subfigure}
    \caption{$\tacc$ by varying $\gV$ with $\gG(M=4,N=4,C=2),d=10, K=40$ and \textbf{fixed token processors} sampled from $\gH(1, \texttt{LeakyRelu})$.}
    \label{fig:ablation_fixed_H_vary_V_N_4_M_4_C_2_n_ex_40}
    \vspace{-2mm}
\end{figure*}

\paragraph{Vary $\gV$.} Following the same setup as Section~\ref{subsec:vary_V} with varying vocabularies, we consider the ablations with a fixed DAG and $C$ MLPs. We observe from Figure~\ref{fig:ablation_fixed_random_dag_vary_V_N_4_M_4_C_2_n_ex_40} that when the DAG is fixed across all sequences, there is a slight increase in the evaluation $\tacc$ when compared to the random DAG per sequence case in Figure~\ref{fig:vary_V_N_4_M_4_C_2_n_ex_40}. However, notice from Figure~\ref{fig:ablation_fixed_H_vary_V_N_4_M_4_C_2_n_ex_40} that fixing the $C$ MLPs for all sequences facilitates the models to learn the causal structure with ease and results in very high $\tacc$ when CoT is enabled. Furthermore, notice that for non-CoT datasets, \texttt{TF-8} (Figure~\ref{fig:ablation_fixed_H_vary_V_N_4_M_4_C_2_n_ex_40_L_8}) and \texttt{TF-12} (Figure~\ref{fig:ablation_fixed_H_vary_V_N_4_M_4_C_2_n_ex_40_L_12}) reach higher a $\tacc$ than the smaller \texttt{TF-4} model (Figure~\ref{fig:ablation_fixed_H_vary_V_N_4_M_4_C_2_n_ex_40_L_4}) for $|\gV|=\{512, 1024\}$.

\begin{figure*}[ht!]
    \centering
    \begin{subfigure}[b]{0.32\textwidth}
        \centering
        \includegraphics[width=\textwidth]{images/ablations/fixed_random_dag/vary_M/llama_L_4_V_1024_H_32_M_2_N_4_C_4_n_ex_40_n_dims_10_hs_2048_bs_64_act_leaky_relu_data_std_1_cot_True.jpg}
        \caption{\texttt{TF-4}}
        \label{fig:ablation_fixed_random_dag_vary_M_N_4_C_4_V_1024_d_10_n_ex_40_L_4}
    \end{subfigure}
    \hfill
    \begin{subfigure}[b]{0.32\textwidth}
        \centering
        \includegraphics[width=\textwidth]{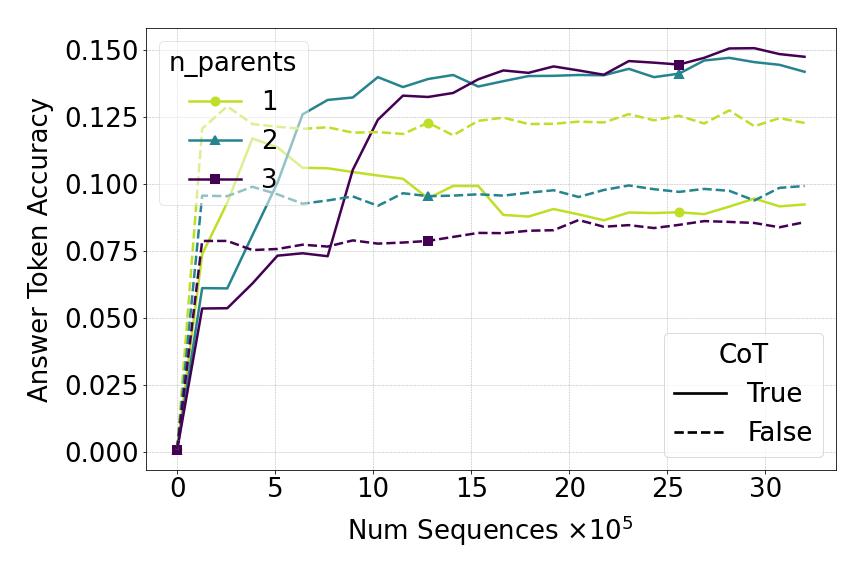}
        \caption{\texttt{TF-8}}
        \label{fig:ablation_fixed_random_dag_vary_M_N_4_C_4_V_1024_d_10_n_ex_40_L_8}
    \end{subfigure}
    \hfill
    \begin{subfigure}[b]{0.32\textwidth}
        \centering
        \includegraphics[width=\textwidth]{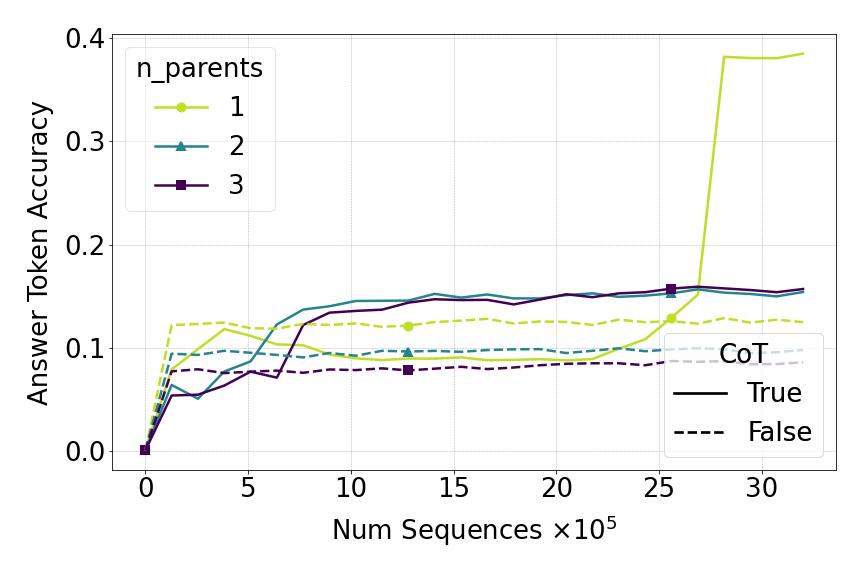}
        \caption{\texttt{TF-12}}
        \label{fig:ablation_fixed_random_dag_vary_M_N_4_C_4_V_1024_d_10_n_ex_40_L_12}
    \end{subfigure}
    \caption{$\tacc$ by varying $M$ with $\gH(1, \texttt{LeakyRelu})$, $d=10,|\gV|=1024,K=40$ and a \textbf{fixed DAG} sampled from $\gG(N=4,C=4)$.}
    \label{fig:ablation_fixed_random_dag_vary_M_N_4_C_4_V_1024_d_10_n_ex_40}
    \vspace{-2mm}
\end{figure*}

\begin{figure*}[ht!]
    \centering
    \begin{subfigure}[b]{0.32\textwidth}
        \centering
        \includegraphics[width=\textwidth]{images/ablations/fixed_H/vary_M/llama_L_4_V_1024_H_32_M_1_N_4_C_4_n_ex_40_n_dims_10_hs_2048_bs_64_act_leaky_relu_data_std_1_cot_False.jpg}
        \caption{\texttt{TF-4}}
        \label{fig:ablation_fixed_H_vary_M_N_4_C_4_V_1024_d_10_n_ex_40_L_4}
    \end{subfigure}
    \hfill
    \begin{subfigure}[b]{0.32\textwidth}
        \centering
        \includegraphics[width=\textwidth]{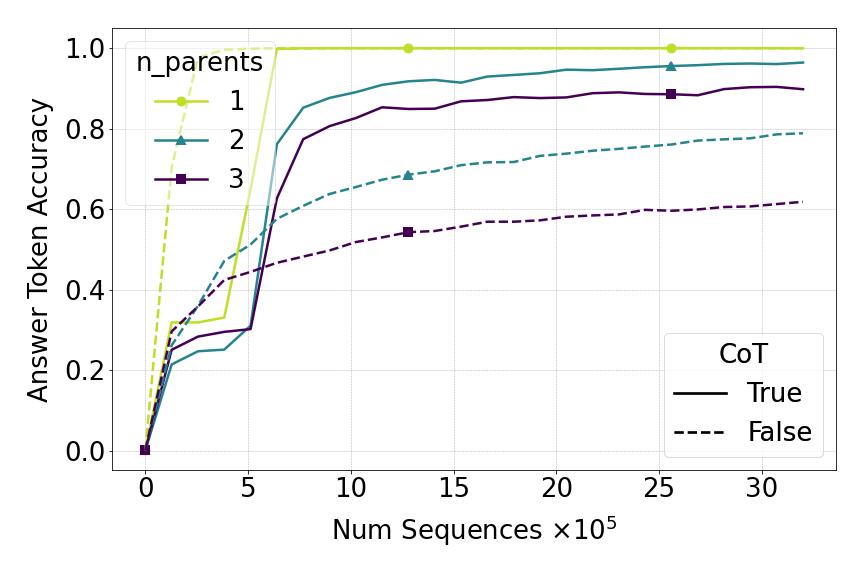}
        \caption{\texttt{TF-8}}
        \label{fig:ablation_fixed_H_vary_M_N_4_C_4_V_1024_d_10_n_ex_40_L_8}
    \end{subfigure}
    \hfill
    \begin{subfigure}[b]{0.32\textwidth}
        \centering
        \includegraphics[width=\textwidth]{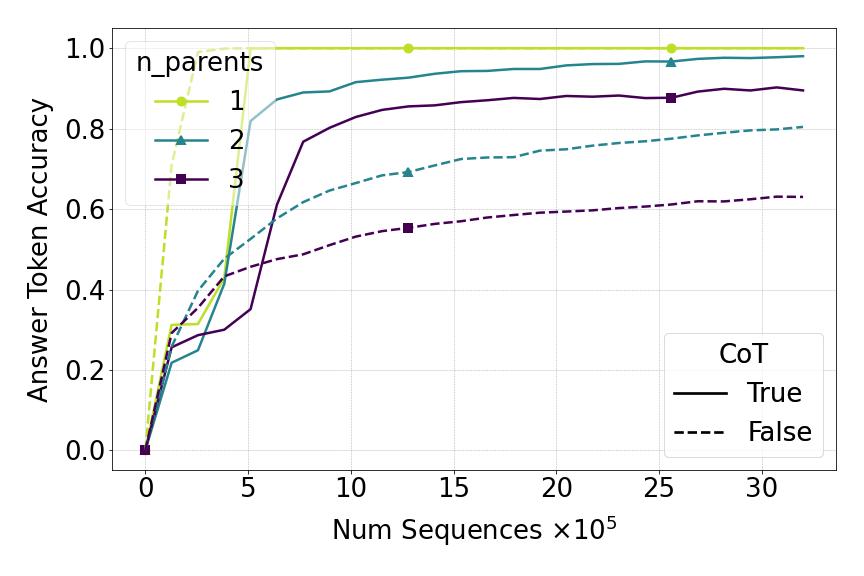}
        \caption{\texttt{TF-12}}
        \label{fig:ablation_fixed_H_vary_M_N_4_C_4_V_1024_d_10_n_ex_40_L_12}
    \end{subfigure}
    \caption{$\tacc$ by varying $M$ with $\gG(N=4,C=4),d=10,|\gV|=1024,K=40$ and \textbf{fixed token processors} sampled from $\gH(1, \texttt{LeakyRelu})$.}
    \label{fig:ablation_fixed_H_vary_M_N_4_C_4_V_1024_d_10_n_ex_40}
    \vspace{-2mm}
\end{figure*}

\paragraph{Vary $M$.} As a follow up of Section~\ref{subsec:ablations_G_H}, we show in Figure~\ref{fig:ablation_fixed_random_dag_vary_M_N_4_C_4_V_1024_d_10_n_ex_40} that when the DAG is fixed across all sequences, the \texttt{TF-12} model exhibits a transition to higher $\tacc$ (with $M=1$) towards the end of training (Figure~\ref{fig:ablation_fixed_random_dag_vary_M_N_4_C_4_V_1024_d_10_n_ex_40_L_12}), which is not observed in the case of \texttt{TF-4} (Figure~\ref{fig:ablation_fixed_random_dag_vary_M_N_4_C_4_V_1024_d_10_n_ex_40_L_4}) and \texttt{TF-8} (Figure~\ref{fig:ablation_fixed_random_dag_vary_M_N_4_C_4_V_1024_d_10_n_ex_40_L_8}). On the other hand, when we fix the $C$ token processors, Figure~\ref{fig:ablation_fixed_H_vary_M_N_4_C_4_V_1024_d_10_n_ex_40} shows that all models can achieve a perfect $\tacc$ of $1$ with $M=1$ and CoT. Interestingly, similar to the previous case of varying $\gV$, we observe that the \texttt{TF-8} (Figure~\ref{fig:ablation_fixed_H_vary_M_N_4_C_4_V_1024_d_10_n_ex_40_L_8}) and \texttt{TF-12} (Figure~\ref{fig:ablation_fixed_H_vary_M_N_4_C_4_V_1024_d_10_n_ex_40_L_12}) models reach a higher $\tacc$ on the non-CoT datasets pertaining to $M=\{2,3\}$, when compared with the smaller \texttt{TF-4} model (Figure~\ref{fig:ablation_fixed_H_vary_M_N_4_C_4_V_1024_d_10_n_ex_40_L_4}).

\section{Interpreting the Attention Maps}
\label{app:sec:attention_maps}

\begin{figure}[t!]
    \centering
    \begin{subfigure}[b]{0.24\textwidth}
        \centering
        \includegraphics[width=\textwidth]{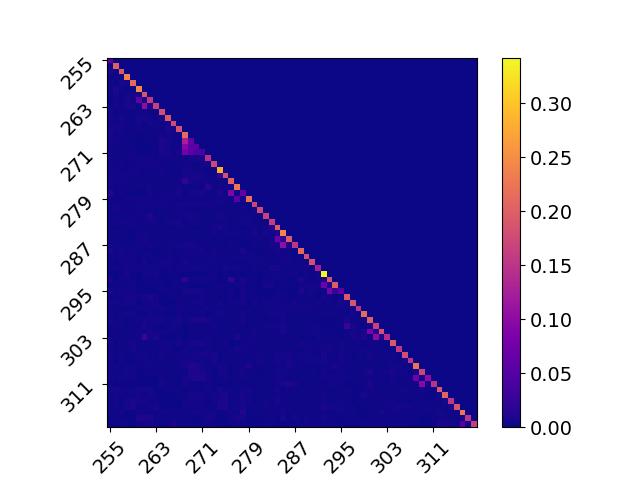}
        \caption{Layer $1$}
    \end{subfigure}
    \begin{subfigure}[b]{0.24\textwidth}
        \centering
        \includegraphics[width=\textwidth]{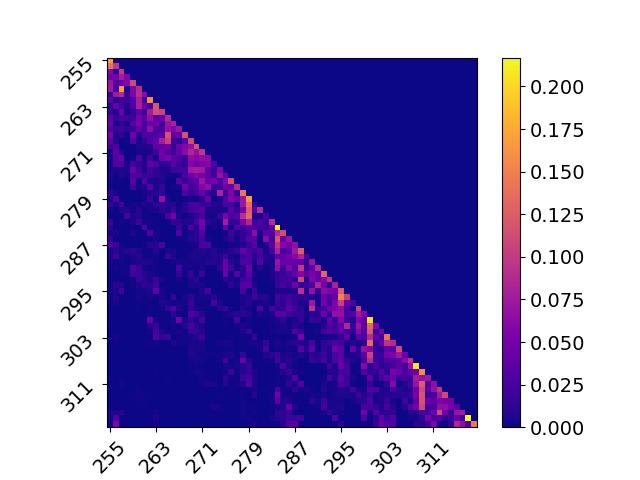}
        \caption{Layer $2$}
        
    \end{subfigure}
    \begin{subfigure}[b]{0.24\textwidth}
        \centering
        \includegraphics[width=\textwidth]{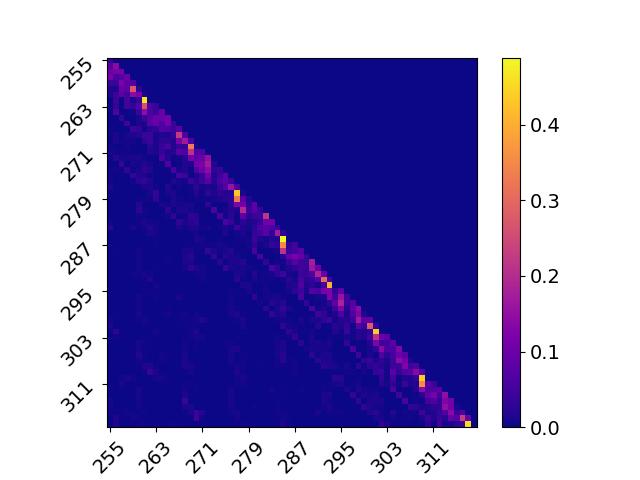}
        \caption{Layer $3$}
        
    \end{subfigure}
    \begin{subfigure}[b]{0.24\textwidth}
        \centering
        \includegraphics[width=\textwidth]{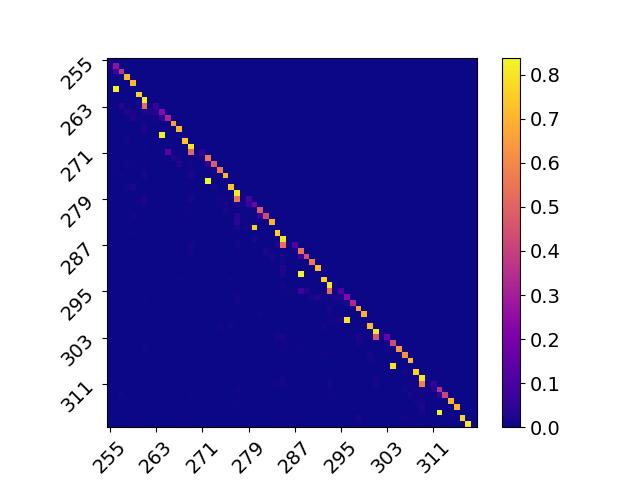}
        \caption{Layer $4$}
        
    \end{subfigure}
    \caption{Mean attention matrices (last $64$ rows and columns) of all $32$ heads across layers of the fully trained \texttt{TF-4} model with setup: $M=1,N=4,C=4, \gG(\texttt{random})$ and fixed token processors sampled from $\gH(1, \texttt{LeakyReLU})$. The DAG structure is $y_1 \leftarrow \{x_4\}, y_2 \leftarrow \{x_1\},y_3 \leftarrow \{y_1\}, y_4 \leftarrow \{y_2\}$. }
\label{fig:fixed_H_all_head_attn_TF_4_M_1_N_4_C_4}
\vspace{-2mm}
\end{figure}

\begin{figure}[t!]
    \centering
    \begin{subfigure}[b]{0.24\textwidth}
        \centering
        \includegraphics[width=\textwidth]{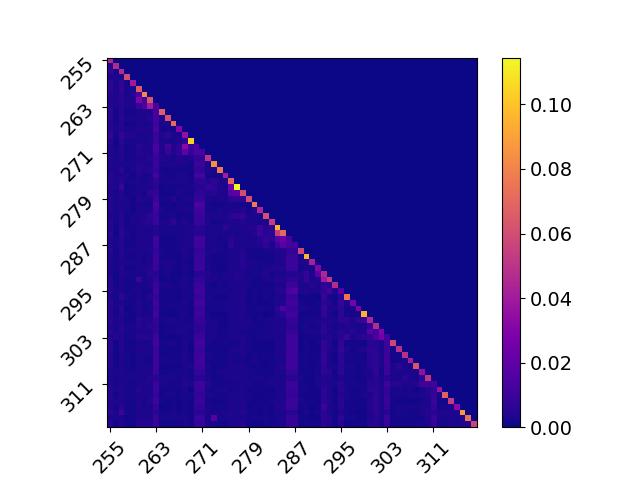}
        \caption{Layer $1$}
    \end{subfigure}
    \begin{subfigure}[b]{0.24\textwidth}
        \centering
        \includegraphics[width=\textwidth]{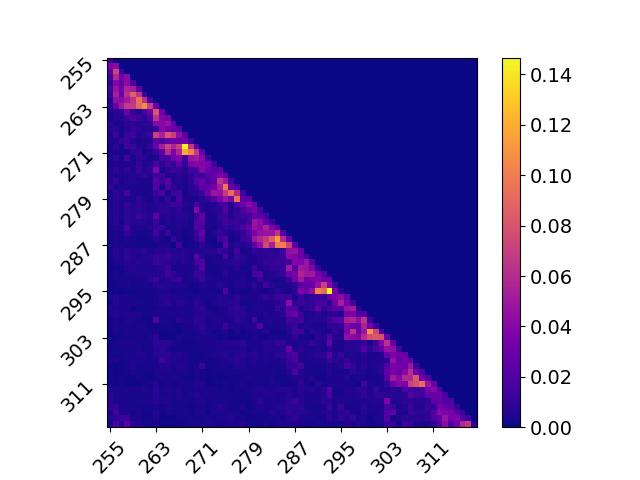}
        \caption{Layer $2$}
        
    \end{subfigure}
    \begin{subfigure}[b]{0.24\textwidth}
        \centering
        \includegraphics[width=\textwidth]{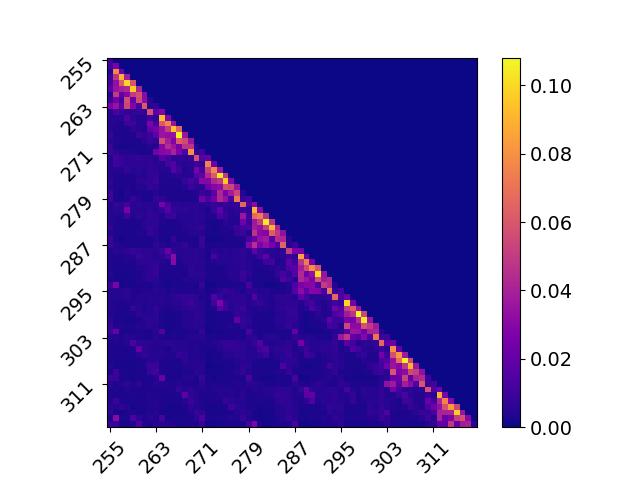}
        \caption{Layer $3$}
        
    \end{subfigure}
    \begin{subfigure}[b]{0.24\textwidth}
        \centering
        \includegraphics[width=\textwidth]{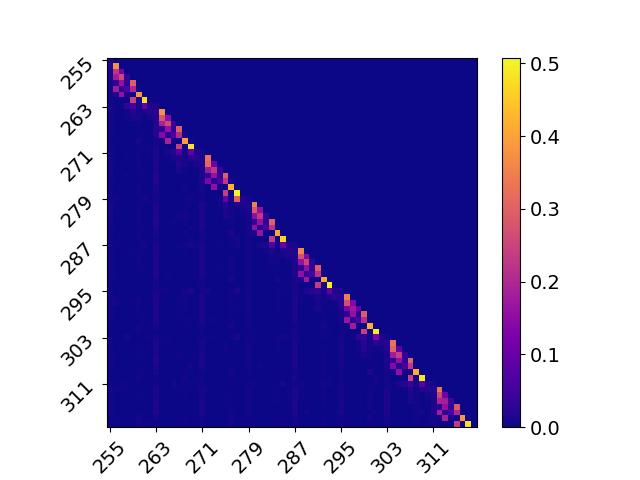}
        \caption{Layer $4$}
        
    \end{subfigure}
    \caption{Mean attention matrices (last $64$ rows and columns) of all $32$ heads across layers of the fully trained \texttt{TF-4} model with setup: $M=2,N=4,C=4, \gG(\texttt{random})$ and fixed token processors sampled from $\gH(1, \texttt{LeakyReLU})$. The DAG structure is $y_1 \leftarrow \{x_4, x_2\}, y_2 \leftarrow \{x_1, x_4\}, y_3 \leftarrow \{y_1, x_2\}, y_4 \leftarrow \{y_2, x_4\}$.}
\label{fig:fixed_H_all_head_attn_TF_4_M_2_N_4_C_4}
\vspace{-2mm}
\end{figure}

\begin{figure}[t!]
    \centering
    \begin{subfigure}[b]{0.24\textwidth}
        \centering
        \includegraphics[width=\textwidth]{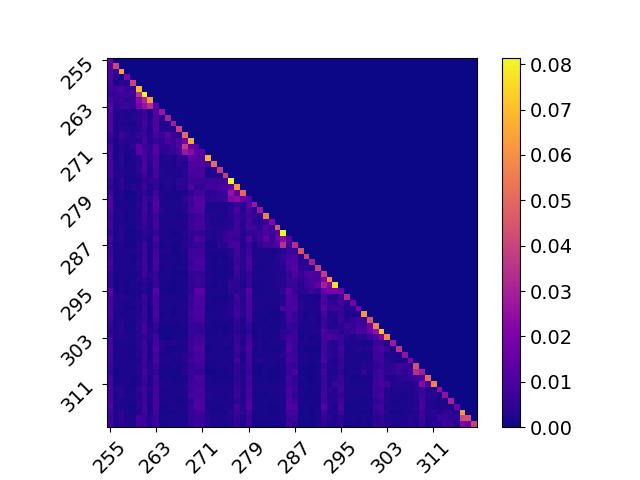}
        \caption{Layer $1$}
        \label{fig:}
    \end{subfigure}
    \begin{subfigure}[b]{0.24\textwidth}
        \centering
        \includegraphics[width=\textwidth]{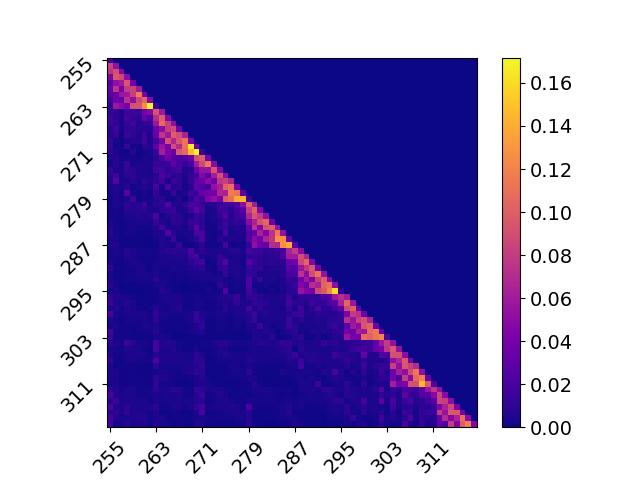}
        \caption{Layer $2$}
        
    \end{subfigure}
    \begin{subfigure}[b]{0.24\textwidth}
        \centering
        \includegraphics[width=\textwidth]{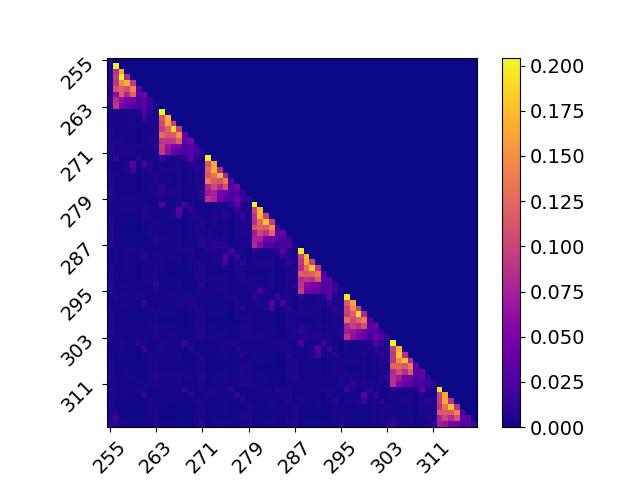}
        \caption{Layer $3$}
        
    \end{subfigure}
    \begin{subfigure}[b]{0.24\textwidth}
        \centering
        \includegraphics[width=\textwidth]{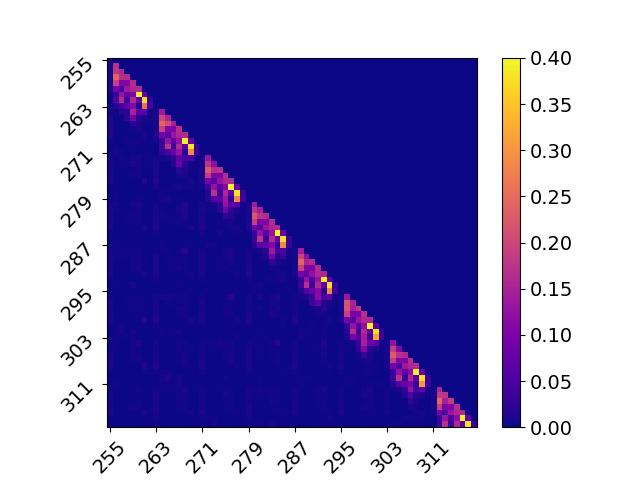}
        \caption{Layer $4$}
        
    \end{subfigure}
    \caption{Mean attention matrices (last $64$ rows and columns) of all $32$ heads across layers of the fully trained \texttt{TF-4} model with setup: $M=3,N=4,C=4, \gG(\texttt{random})$ and fixed token processors from $\gH(1, \texttt{LeakyReLU})$. The DAG structure is $y_1 \leftarrow \{x_4, x_2, x_3\}, y_2 \leftarrow \{x_1, x_4, y_1\}, y_3 \leftarrow \{y_1, x_2, x_4\}, y_4 \leftarrow \{y_2, x_4, x_2\}$.}
\label{fig:fixed_H_all_head_attn_TF_4_M_3_N_4_C_4}
\vspace{-2mm}
\end{figure}

To interpret the attention maps, we consider the same ablation setup as Section~\ref{subsec:ablations_G_H} in the main text with fixed token processing functions and consider $M=1, |\gV|=1024$ and the \texttt{TF-4} model for simplicity. Let the tokens $x_1, x_2, x_3, x_4$ and $y_1, y_2, y_3, y_4$ represent the input and chain tokens of a CoT example respectively. We consider a validation sequence with the DAG structure (i.e the parent tokens) for the $4$ chain tokens given by: $y_1 \leftarrow \{x_4\}, y_2 \leftarrow \{x_1\},y_3 \leftarrow \{y_1\}, y_4 \leftarrow \{y_2\}$ for the analysis. Now, given such a validation sequence prepared with $K=40$, we auto-regressively generate the $4$ chain tokens and consider the attention maps used for generating the last chain token (i.e the answer token). We take the mean of all attention maps across the heads in a layer and plot the last $64$ rows and columns in Figure~\ref{fig:fixed_H_all_head_attn_TF_4_M_1_N_4_C_4}. Furthermore, we also plot such attention maps for the $M=2$ case in Figure~\ref{fig:fixed_H_all_head_attn_TF_4_M_2_N_4_C_4} and the $M=3$ case in Figure~\ref{fig:fixed_H_all_head_attn_TF_4_M_3_N_4_C_4} with the DAGs described in the captions.

\begin{figure}[t!]
    \centering
    \begin{subfigure}[b]{0.24\textwidth}
        \centering
        \includegraphics[width=\textwidth]{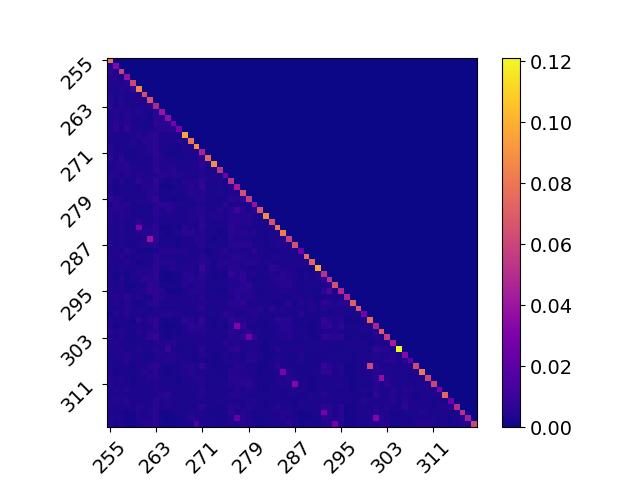}
        \caption{Layer $1$}
    \end{subfigure}
    \begin{subfigure}[b]{0.24\textwidth}
        \centering
        \includegraphics[width=\textwidth]{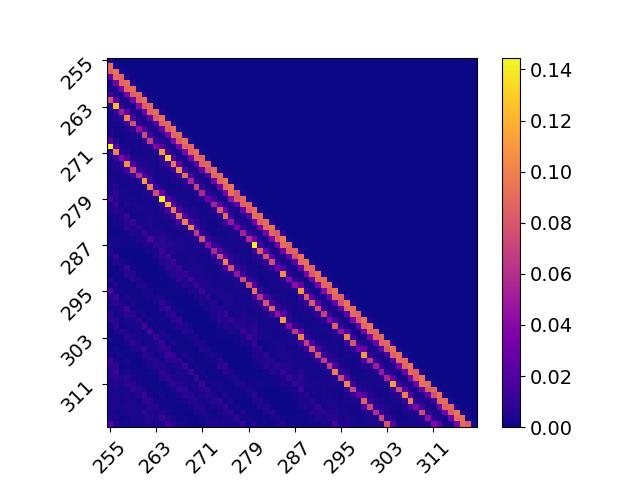}
        \caption{Layer $2$}
        
    \end{subfigure}
    \begin{subfigure}[b]{0.24\textwidth}
        \centering
        \includegraphics[width=\textwidth]{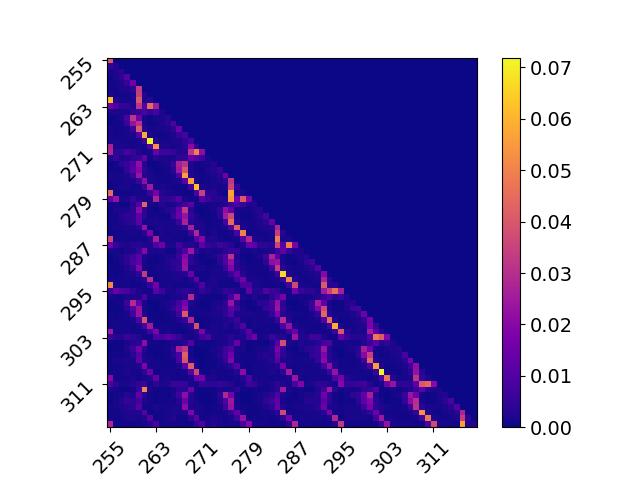}
        \caption{Layer $3$}
        
    \end{subfigure}
    \begin{subfigure}[b]{0.24\textwidth}
        \centering
        \includegraphics[width=\textwidth]{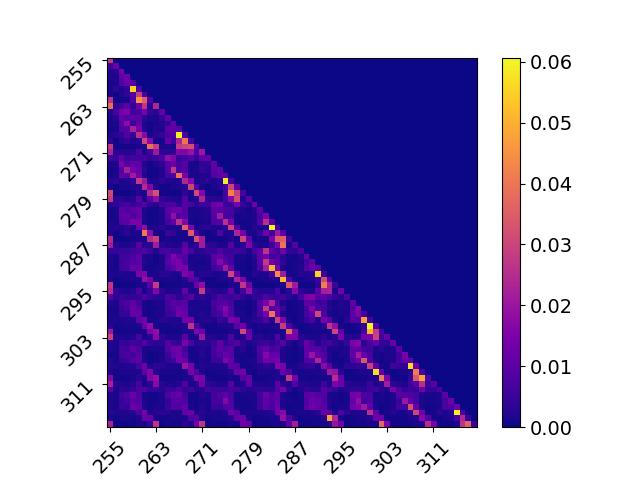}
        \caption{Layer $4$}
        
    \end{subfigure}
    \caption{Mean attention matrices (last $64$ rows and columns) of all $32$ heads across layers of the fully trained \texttt{TF-4} model with setup: $M=1,N=4,C=4, \gG(\texttt{random}), \gH(1, \texttt{LeakyReLU})$. The parent tokens of the chain tokens are $y_1 \leftarrow \{x_4\}, y_2 \leftarrow \{x_1\},y_3 \leftarrow \{y_1\}, y_4 \leftarrow \{y_2\}$. When the cardinality of $\gH$ is not bounded, we notice that the Layer $4$ attention maps indicate a reliance on the answer tokens of previous in-context examples to generate the final answer token. }
\label{fig:all_head_attn_TF_4_M_1_N_4_C_4}
\vspace{-2mm}
\end{figure}

Qualitatively, we observed that even for the relatively difficult setup of $M=\{2,3\}$, larger attention scores for the chain tokens in any example are placed on the parent tokens in that particular example itself (similar to the $M=1$ case in Figure~\ref{fig:fixed_H_all_head_attn_TF_4_M_1_N_4_C_4_layer_3_zoom}). However, in the generic case of randomly sampling the token processors per sequence, we observed that the mean attention scores of the last layer indicate a reliance on the answer tokens of previous examples (see Figure~\ref{fig:all_head_attn_TF_4_M_1_N_4_C_4}). This is an interesting observation which requires further study on the dependence between diversity of token processors and the attention map score distributions.

\section{Reasoning with \texttt{DeepSeek-R1-Distill-Llama-8B}}
\label{app:reasoning_outputs_deepseek_llama}

\begin{figure*}[h]
\begin{tcolorbox}[colback=black!10, colframe=black!80, width=\textwidth, arc=0pt, outer arc=0pt, boxsep=0pt, left=10pt, right=0pt, top=5pt, bottom=6pt]
{\tiny\begin{lstlisting}[basicstyle=\tiny\ttfamily\color{black}\bfseries,
  keywordstyle=\color{codeblue}\bfseries, % Keywords in bold blue
  commentstyle=\color{codegreen}\itshape, % Comments in italic green
  stringstyle=\color{black}            % Strings in gray
]
The following is a multiple choice question about high school mathematics.

The length of a rectangle is twice its width. Given the length of the diagonal is $5\sqrt{5}$,
find the area of the rectangle.
A. 2500
B. 2
C. 50
D. 25
Which of these options is correct? Please provide a concise explanation and either option A,B,C
or D in the format: ### Final Option:
\end{lstlisting}}
\end{tcolorbox}
\caption{The input prompt to the \texttt{DeepSeek-R1-Distill-Llama-8B} model.}
\end{figure*}

\vspace{-3mm}
\begin{figure}[h]
\begin{tcolorbox}[colback=black!10, colframe=black!80, width=\textwidth, arc=0pt, outer arc=0pt, boxsep=0pt, left=10pt, right=0pt, top=5pt, bottom=6pt]
{\tiny\begin{lstlisting}[basicstyle=\tiny\ttfamily\color{black}\bfseries,
  keywordstyle=\color{codeblue}\bfseries, % Keywords in bold blue
  commentstyle=\color{codegreen}\itshape, % Comments in italic green
  stringstyle=\color{black}            % Strings in gray
]
Okay, I'm trying to solve this problem about the rectangle. Let's see. The question says the length is twice its width,
and the diagonal is 5\sqrt{5}. We need to find the area of the rectangle. Hmm, okay, I remember that in a rectangle,
the diagonal forms a right triangle with the length and width. So maybe I can use the Pythagorean theorem here.

Let me denote the width as 'w'. Then the length, since it's twice the width, would be '2w'. Right, so length is 2w,
width is w. Now, the diagonal of the rectangle is the hypotenuse of this right-angled triangle. The formula for the
diagonal (d) would be sqrt( (length)^2 + (width)^2 ). Wait, that's the Pythagorean theorem.

Given that the diagonal is 5\sqrt{5}, I can set up the equation: sqrt( (2w)^2 + w^2 ) = 5\sqrt{5}. Let me square
both sides to eliminate the square root. That gives me (2w)^2 + w^2 = (5\sqrt{5})^2.

Calculating each term: (2w)^2 is 4w^2, and w^2 is just w^2. So adding them together, 4w^2 + w^2 is 5w^2. On the right side,
(5\sqrt{5})^2 is 25 * 5, which is 125.

So now I have 5w^2 = 125. To find w^2, I can divide both sides by 5. That gives w^2 = 25. Taking the square root
of both sides, w = 5. Since width can't be negative, we take the positive value.

So the width is 5 units. Then the length, which is twice that, is 2 * 5 = 10 units. Now, the area of the rectangle is length
multiplied by width, so that's 10 * 5 = 50.

Looking at the options, option C is 50. So I think the correct answer is C. Let me just double-check my steps. I set up
the variables correctly, used the Pythagorean theorem, squared both sides properly, solved for w, and then calculated
the area. Everything seems to add up. Yeah, I'm confident it's C.
</think>

The problem involves finding the area of a rectangle where the length is twice the width and the diagonal is \(5\sqrt{5}\).
By denoting the width as \(w\), the length becomes \(2w\). Using the Pythagorean theorem for the diagonal:

\[
\sqrt{(2w)^2 + w^2} = 5\sqrt{5}
\]

Squaring both sides:

\[
4w^2 + w^2 = 125 \Rightarrow 5w^2 = 125 \Rightarrow w^2 = 25 \Rightarrow w = 5
\]

The width is 5, so the length is \(2 \times 5 = 10\). The area is:

\[
5 \times 10 = 50
\]

Thus, the correct answer is **C. 50**.

### Final Option: C
\end{lstlisting}}
\end{tcolorbox}
\caption{The output of the \texttt{DeepSeek-R1-Distill-Llama-8B} model with \texttt{temperature=0.6}.}
\end{figure}

\begin{figure}[h!]
\begin{tcolorbox}[colback=black!10, colframe=black!80, width=\textwidth, arc=0pt, outer arc=0pt, boxsep=0pt, left=10pt, right=0pt, top=5pt, bottom=6pt]
{\tiny\begin{lstlisting}[basicstyle=\tiny\ttfamily\color{black}\bfseries,
  keywordstyle=\color{codeblue}\bfseries, % Keywords in bold blue
  commentstyle=\color{codegreen}\itshape, % Comments in italic green
  stringstyle=\color{black}            % Strings in gray
]
Okay, so I have this problem here about a rectangle. The problem says that the length is twice its width. Hmm,
I remember that in a rectangle, the sides are pairs of equal lengths. So if the length is twice the width,
I can represent them as variables. Let me think, maybe width is w and length would be 2w.

Then the problem mentions the diagonal is 5 times the square root of 5. That's 5\sqrt{5}. I know that in a rectangle,
the diagonal forms a right triangle with the length and width. So I can use the Pythagorean theorem here. 
The theorem states that in a right-angled triangle, the square of the hypotenuse (which is the diagonal here)
is equal to the sum of the squares of the other two sides.

Okay, so the diagonal is 5\sqrt{5}, so I should square that. Let me calculate that: (5\sqrt{5})^2. Squaring 5 gives 25,
and squaring \sqrt{5} gives 5. So 25 times 5 is 125. So the diagonal squared is 125.

Now, the length is 2w, and the width is w. Plugging into the Pythagorean theorem: (2w)^2 + w^2 = (5\sqrt{5})^2.
Let me compute (2w)^2, which is 4w^2, and w^2 is w^2. Adding them together, it's 5w^2. So 5w^2 = 125.

To find w^2, I can divide both sides by 5. That gives me w^2 = 25. So w is the square root of 25, which is 5.
Therefore, the width is 5 units.

Then the length is twice that, so 2 times 5 is 10. So the length is 10 and the width is 5. To find the area of the rectangle,
I multiply length by width. So 10 times 5 is 50. Hmm, that's 50. Let me check the options. Option C is 50. 
So I think the answer is C.

Wait, let me double-check because sometimes when I do math quickly, I might makes mistakes. So let me recast it.

Given length is twice the width, so l=2w. The diagonal squared equals l squared plus w squared. So substituting,
(2w)^2 + w^2 = (5\sqrt{5})^2.

That's 4w^2 + w^2 = 25 * 5. So 5w^2 = 125. Dividing both sides by 5 gives w^2=25. So w=5, since width can't be negative.
So length is 10. Area is l times w, so 10*5=50. Yep, that seems correct.
So the area should be 50, which is option C.
</think>

The length of the rectangle is twice its width. Using the Pythagorean theorem for the diagonal, we find the width and length,
then calculate the area.

### Step-by-Step Explanation:

1. **Define Variables:**
   - Let the width be \( w \).
   - Then, the length is \( 2w \).

2. **Apply Pythagorean Theorem:**
   - The diagonal is \( 5\sqrt{5} \).
   - \( (length)^2 + (width)^2 = (diagonal)^2 \)
   - Substitute \( length = 2w \) and \( width = w \):
     \[
     (2w)^2 + w^2 = (5\sqrt{5})^2
     \]
   - Simplify:
     \[
     4w^2 + w^2 = 25 \times 5
     \]
     \[
     5w^2 = 125
     \]

3. **Solve for \( w^2 \):**
   - \( w^2 = \frac{125}{5} = 25 \)
   - \( w = \sqrt{25} = 5 \) (since width can't be negative).

4. **Find Length and Area:**
   - Length \( = 2w = 2 \times 5 = 10 \).
   - Area \( = length \times width = 10 \times 5 = 50 \).

### Final Answer:
The area of the rectangle is **50**, which corresponds to option **C**.

### Final Option:
C.
\end{lstlisting}}
\end{tcolorbox}
\caption{The output of the \texttt{DeepSeek-R1-Distill-Llama-8B} model with \texttt{temperature=0.9}.}
\end{figure}

\newpage
\section{Comparison with Related Work}
\label{app:sec:comp_related_work}

\paragraph{ICL with real-valued examples.} Analyzing the ICL capabilities of transformer models with synthetic data has gained massive attention in recent years. In particular, the notion of using these models as ``statisticians" which can learn and approximate arbitrary function classes on real valued inputs has been widely explored~\cite{garg2022can,bai2023transformers,von2023transformers,ahn2023transformers,li2023transformers,oko2024pretrained} (see also \cite{dong2022survey}). Unlike the setup considered in this paper, these works consider single input-output examples $(x_k, y_k)_{k=1}^K$ in context where $x_i \sim \gN(\vzero,\mI_d) \in \sR^d$ and $y_i \in \sR$. The goal here is to learn the linear/non-linear functions $f \in \gF$ such that $y_k = f(x_k), \forall k \in [K]$. Although such a setup has resulted in valuable insights on how transformers learn complex function classes (linear/noisy/sparse regression, single-index models, shallow neural networks etc), it is unclear if it is suitable to explain the ICL capabilities of transformers in NLP settings with tokenized multi input-output examples.

\paragraph{ICL with CoT.} In the real-valued setting, a recent work by \cite{li2023dissecting} explored the role of CoT for learning the MLP function classes. More importantly, they evaluate the CoT outputs of the transformers without teacher-forcing and show how this decomposition can facilitate a hierarchical layer-wise learning of the MLPs. On the other hand, several works have explored simple tokenized settings which employ markov chain like causal structures to study the attention maps and training dynamics of shallow transformers \cite{edelman2024the, nichani2024how,akyurek2024incontext,bietti2023birth}. The main idea here is to consider a small vocabulary $\gV$ and associate a causal dependency between tokens using single/multiple markov chains. In particular, a single input token $x_1 \in \gV$ is used to generate the chain tokens $y_1, \cdots, y_C \in \gV^C$ as per the transition probabilities of the chain.

We highlight that the generic setup of \coticl~can specialize to a markov chain like structures (via DAGs) and also allow researchers to explore token processing functions beyond MLPs (see also Table~\ref{tab:related_work_comparison}) to extend such lines of work. More importantly, the multi input-output ICL setup along with data embeddings allows us to take a step towards modeling NLP like scenarios with interpretable synthetic data.

\begin{table*}[ht!]
\centering
\begin{tabular}{|c|c|c|c|c|}
\hline
\textbf{Previous Work} & \textbf{Tokenized Setup} & \textbf{CoT in ICL} & \textbf{Multi-Input Output ICL} & \textbf{Explicit DAG}   \\ \hline
\citet{von2023transformers}     &  $\times$   &  $\times$  &  $\times$  &  $\times$\\ \hline
\citet{ahn2023transformers}     &  $\times$   &  $\times$  &  $\times$ &  $\times$ \\ \hline
\citet{garg2022can}     &  $\times$   &  $\times$  &  $\times$ &  $\times$ \\ \hline
\citet{bai2023transformers}     &  $\times$   &  $\times$  &  $\times$ &  $\times$ \\ \hline
\citet{li2023transformers}     &  $\times$   &  $\times$  &  $\times$ &  $\times$ \\ \hline
\citet{guo2024how}     &  $\times$   &  $\times$  &  $\times$ &  $\times$ \\ \hline
\citet{panwar2024incontext}     &  $\times$   &  $\times$  &  $\times$ &  $\times$ \\ \hline
\citet{oko2024pretrained}     &  $\times$   &  $\times$  &  $\times$ &  $\times$ \\ \hline
\citet{xie2022an}   &  $\checkmark$   &  $\times$  &  $\times$&  $\times$  \\ \hline
\citet{akyurek2024incontext}   &  $\checkmark$   &  $\times$  &  $\times$ &  $\times$ \\ \hline
\citet{dai2023can}   &  $\checkmark$   &  $\times$  &  $\times$ &  $\times$  \\ \hline
\citet{deutch2024context}   &  $\checkmark$   &  $\times$  &  $\times$ &  $\times$ \\ \hline
\citet{edelman2024the}   &  $\checkmark$   &  $\times$  &  $\times$ &  $\checkmark$ \\ \hline
\citet{nichani2024how}   &  $\checkmark$   &  $\times$  &  $\times$ &  $\checkmark$ \\ \hline
\citet{feng2023towards}  &  $\checkmark$   &  $\checkmark$  &  $\times$  &  $\times$ \\ \hline
\citet{prystawski2023think}  &  $\checkmark$   &  $\checkmark$  &  $\times$ &  $\checkmark$  \\ \hline
\citet{merrill2024the}  &  $\checkmark$   &  $\checkmark$  &  $\times$  &  $\times$ \\ \hline
\citet{hou2023towards}  &  $\checkmark$   &  $\checkmark$  &  $\times$ &  $\checkmark$  \\ \hline
\citet{li2023dissecting}      &  $\times$   &  $\checkmark$  &  $\times$  &  $\times$ \\ \hline
\textbf{\coticl}      &  $\checkmark$   &  $\checkmark$  &  $\checkmark$ &  $\checkmark$ \\ \hline
\end{tabular}
\caption{Comparison with related work on ICL and CoT. \maz{model sizes?}}
\label{tab:related_work_comparison}
\end{table*}

\newpage
\section{Summary of Notations}

\begin{table*}[!htbp]
\centering
\begin{tabular}{|c|c|}
\hline
\textbf{Notation} & \textbf{Description}  \\ \hline
$\gV$ & A custom vocabulary used to generate tokens \\ \hline
$d$ & The data embedding dimension in $\mE_{\texttt{data}} \in \sR^{|\gV| \times d}$. \\ \hline
$\mE_{\texttt{data}}$ & A constant `unknown' data embedding matrix \\ \hline
$\gG$ & A class of functions to filter/select tokens \\ \hline
$\gH$ & A class of functions to process token embeddings (rows in $\mE_{\texttt{data}}$) \\ \hline
$\gF$ & A function class composed of $\gG, \gH$ \\ \hline
$l$ & The depth of the MLP in $\gH$ \\ \hline
$\phi$ & The activation function used in the MLP in $\gH$ \\ \hline
 $N$   & Number of input tokens per example   \\ \hline  
 $M$   & Number of tokens selected by $\gG$   \\ \hline  
 $C$   & Number of chain tokens (Chain length)  \\ \hline 
 $K$   & Number of examples per sequence   \\ \hline  
 \texttt{TF}   & A decoder only transformer model   \\ \hline  
 $\vx = (x_1, \cdots, x_N) \in \gV^N$ & Input tokens in an example \\ \hline  
 $\vy = (y_1, \cdots, y_C) \in \gV^C$ & Chain tokens in an example \\ \hline  
  $\vp^K(f)$ & A sequence composed using $f \in \gF$ with $K$ examples in-context. \\ \hline
  $\vp_{CoT}^K(f)$ & A CoT sequence composed using $f \in \gF$ with $K$ examples in-context. \\ \hline
   $\tTF^{\circ C}(\cdot)$ & The $C$-step auto-regressive greedy token generation by the $\tTF$ model. \\ \hline
\end{tabular}
\caption{A summary of notations used throughout the paper.}
\label{tab:notations}
\end{table*}

\end{document}